\documentclass[twoside]{IEEEtran}
\usepackage[T1]{fontenc}
\usepackage[utf8]{inputenc}
\usepackage{color}
\definecolor{document_fontcolor}{rgb}{0, 0, 0}
\color{document_fontcolor}
\usepackage{array}
\usepackage{verbatim}
\usepackage{float}
\usepackage{mathtools}
\usepackage{endnotes}
\usepackage{multirow}
\usepackage{amsmath}
\usepackage{amsthm}
\usepackage{amssymb}
\usepackage{stackrel}
\usepackage{graphicx}
\usepackage{setspace}
\PassOptionsToPackage{normalem}{ulem}
\usepackage{ulem}
\usepackage{nameref}

\makeatletter

\providecommand{\tabularnewline}{\\}
\floatstyle{ruled}
\newfloat{algorithm}{tbp}{loa}
\providecommand{\algorithmname}{Algorithm}
\floatname{algorithm}{\protect\algorithmname}

\theoremstyle{plain}
\newtheorem{thm}{\protect\theoremname}
\theoremstyle{definition}
\newtheorem{problem}[thm]{\protect\problemname}

\usepackage{microtype}
\usepackage{graphicx}

\usepackage{colortbl}
\usepackage{tabularx}
\usepackage{booktabs, adjustbox}
\usepackage{makecell}
\usepackage[colorlinks,bookmarksopen,bookmarksnumbered,linkcolor=blue,citecolor=blue,urlcolor=blue]{hyperref}
\setkeys{Gin}{width=\ifdim\Gin@nat@width>\linewidth
  \linewidth
\else
  \Gin@nat@width
\fi}

\@ifundefined{showcaptionsetup}{}{%
 \PassOptionsToPackage{caption=false}{subfig}}
\usepackage{subfig}
\makeatother

\usepackage[style=numeric-comp,backend=bibtex,bibstyle=ieee,citestyle=numeric-comp]{biblatex}
\providecommand{\problemname}{Problem}
\providecommand{\theoremname}{Theorem}

\begin{document}
\global\long\def\VECJ#1#2#3#4#5{\prescript{#4}{#5}{\boldsymbol{#1}}_{{#2}}^{{#3}}}%

\global\long\def\fr#1#2#3#4#5{\prescript{#1}{#2}{#3}_{{#4}}^{{#5}}}%

\global\long\def\set#1#2#3#4#5{\prescript{#4}{#5}{\mathcal{#1}}_{{#2}}^{{#3}}}%

\global\long\def\SP#1{\VECJ ps{#1}{}{}}%

\global\long\def\c#1{c_{#1}}%

\global\long\def\CR#1{\VECJ rs{fix}{_{#1}}{}}%

\global\long\def\SPn#1{\VECJ p{s,#1}{}{}{}}%

\global\long\def\CF{\set Is{}{}{}}%

\global\long\def\CFn#1#2{\set I{s_{#1}}{#2}{}{}}%

\global\long\def\CV#1#2{\set V{#1}{#2}{}{}}%

\global\long\def\CC#1#2#3{\set C{#1}{#2}{#3}{}}%

\global\long\def\ACM{\mathcal{I\text{-space}}}%

\global\long\def\ACC{\mathcal{C\text{-space}}}%

\global\long\def\ACV{\mathcal{V\text{-space}}}%

\global\long\def\AVGP{\text{VGP}}%

\global\long\def\AVPP{\text{VPP}}%

\global\long\def\ASCP{\text{SCP}}%

\global\long\def\ARVS{\text{RVS}}%

\global\long\def\ACSP{\text{CSP}}%

\global\long\def\ACSG{\text{CSG}}%

\title{Viewpoint Generation using Feature-Based Constrained Spaces for Robot
Vision Systems}
\author{Alejandro Magaña, Jonas Dirr, Philipp Bauer, and Gunther Reinhart
\thanks{The authors are with the Institute for Machine Tools and
Industrial Management of the Technical University of Munich in Munich
Germany.}}
\maketitle
\begin{abstract}
The efficient computation of viewpoints under consideration of various
system and process constraints is a common challenge that any robot
vision system is confronted with when trying to execute a vision task.
Although fundamental research has provided solid and sound solutions
for tackling this problem, a holistic framework that poses its formal
description, considers the heterogeneity of robot vision systems,
and offers an integrated solution remains unaddressed. Hence, this
publication outlines the generation of viewpoints as a geometrical
problem and introduces a generalized theoretical framework based on
Feature-Based Constrained  Spaces ($\ACC$s) as the backbone for solving
it. \textmd{\normalsize{}A $\ACC$} can be understood as the topological
space that a viewpoint constraint spans, where the sensor can be positioned
for acquiring a feature while fulfilling the regarded constraint.
The present study demonstrates that many viewpoint constraints can
be efficiently formulated as $\ACC$s providing geometric, deterministic,
and closed solutions. The introduced $\ACC$s are characterized based
on generic domain and viewpoint constraints models to ease the transferability
of the present framework to different applications and robot vision
systems. The effectiveness and efficiency of the concepts introduced
are verified on a simulation-based scenario and validated on a real
robot vision system comprising two different sensors.
\end{abstract}

\begin{IEEEkeywords}
viewpoint planning, 3D sensors, vision task automation, constraint
planning,  robot vision system
\end{IEEEkeywords}

\begin{table}[tbh]
\centering{}%
\begin{tabular}{|cl|}
\hline 
\multicolumn{2}{|>{\centering}m{0.4\textwidth}|}{Nomenclature}\tabularnewline
\hline 
$\ACC$ & Feature-Based Constrained Space\tabularnewline
$\ACSG$ & Constructive Solid Geometry\tabularnewline
$\ACM$ & Frustum space\tabularnewline
$\ARVS$ & Robot Vision System\tabularnewline
$\AVGP$ & Viewpoint Generation Problem\tabularnewline
$\AVPP$ & Viewpoint Planning Problem\tabularnewline
\hline 
\end{tabular}
\end{table}

\section{Introduction}

\begin{figure}[tbh]
\centering{}\includegraphics[width=0.5\textwidth]{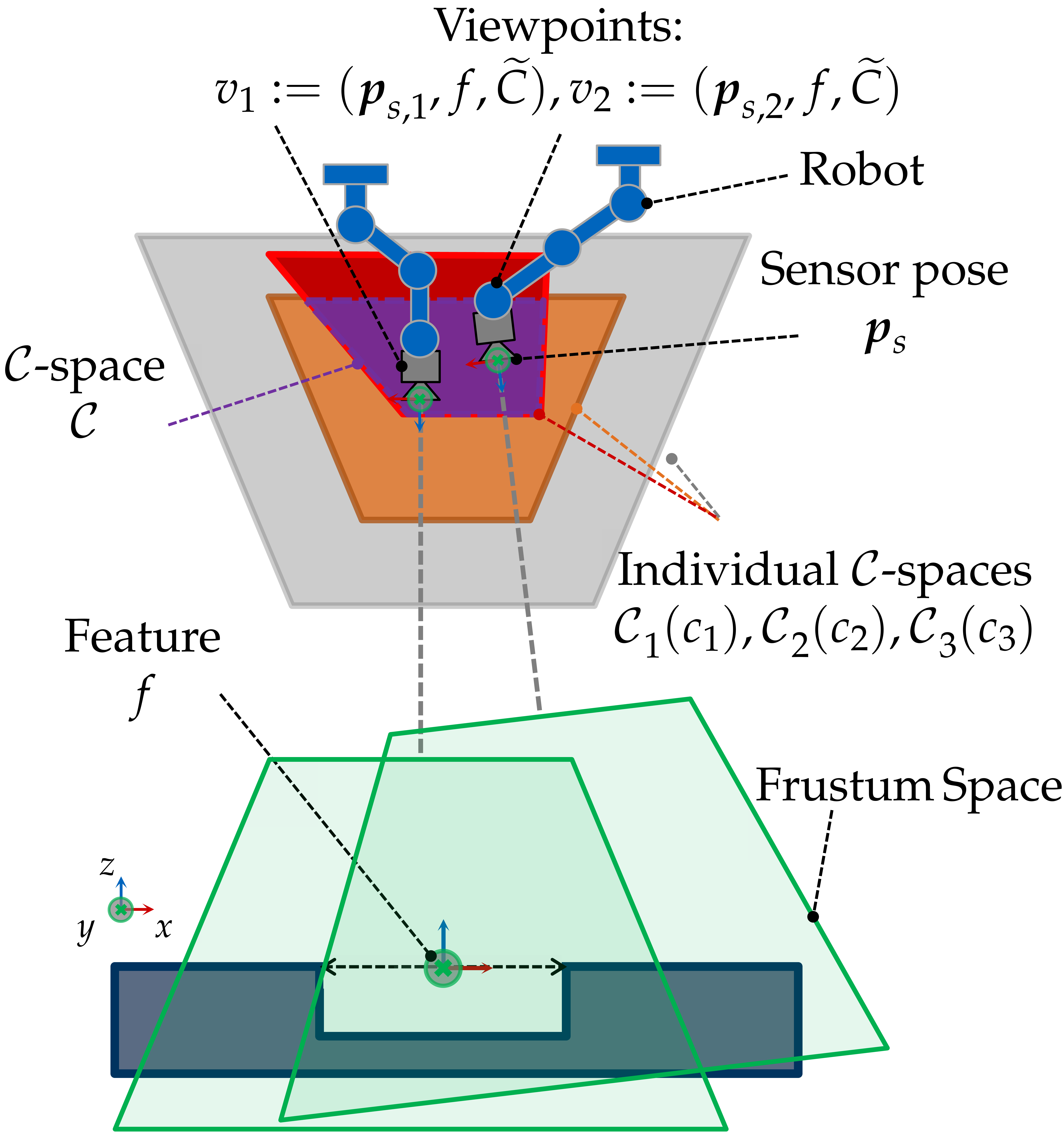}\caption{Simplified, graphical representation of the Viewpoint Generation Problem
($\protect\AVGP$): which are valid sensor poses $\protect\SP{}$
to acquire a feature $f$ considering a set of diverse viewpoint constraints
$\widetilde{C}$? To answer this question, this study proposes the
characterization of Feature-Based Constrained  Spaces ($\protect\ACC$s).
The $\protect\ACC$ denoted as $\protect\CC{}{}{}$ can be regarded
as the geometrical representation of all viewpoint constraints in
the special Euclidean $SE(3)$. Any sensor pose within it $\forall\protect\SP{}\in$$\protect\CC{}{}{}$
can be considered to be valid to acquire a feature satisfying all
viewpoint constraints $\widetilde{C}$. The $\protect\ACC$ is constituted
by individual $\protect\ACC$s $\protect\CC i{}{}(c_{i})$, i.e.,
geometrical representations of each viewpoint constraint $c_{i}\in\widetilde{C}$.
\label{fig:Motivation_VGP}}
\end{figure}

The increasing performance of 2D and 3D image processing algorithms
and the falling prices on electronic components (processors and optical
sensors) over the last two decades, have motivated not only researchers
but also the industry to investigate and automate different machine
vision tasks using robot vision systems ($\ARVS$s) consisting of
a manipulator and a 2D or 3D sensor \parencite{Kragic.2016,PeuzinJubert.2021}.
 Whether programmed offline or online, $\ARVS$s demand multiple
planning modules to execute motion and vision tasks efficiently and
robustly. For instance, the efficient and effective planning of valid
viewpoints to fulfill a vision task considering different constraints
known as—the view(point) planning problem ($\AVPP$)—still represents
an open planning problem within diverse applications \parencite{PeuzinJubert.2021,Gospodnetic.2022},
e.g., camera surveillance, scene exploration, object detection, visual
servoing, object reconstruction, image-based inspection, robot calibration,
and mobile navigation \parencite{Chen.2011,PeuzinJubert.2021}. 

\subsection{Viewpoint Generation Problem solved using $\protect\ACC$s \label{subsec:VGP-Intro}}

To tackle the $\AVPP$, we first re-examine its reformulation and
propose its modularization. Then, this study focuses on the most fundamental
sub-problem of the $\AVPP$, i.e., the Viewpoint Generation Problem
($\AVGP$). The $\AVGP$ addresses the calculation of valid viewpoints
to acquire a single feature considering the fulfillment of different
viewpoint constraints. 

With this in mind, this paper outlines the $\AVGP$ as a purely geometrical
problem that can be solved in the special Euclidean denoted as $SE(3)$
(6D spatial space) based on the concept of Feature-Based Constrained
Spaces ($\ACC$s). $\ACC$s represent the spatial solution space up
to 6D of each viewpoint constraint $c_{i}\in\widetilde{C}$ denoted
as $\CC i{}{}$ that comprises all valid sensor poses $\SP{}$ to
acquire a feature $f$. In other words, it can be assumed that any
sensor pose lying w

ithin an $i$ $\text{\ensuremath{\ACC}}$ fulfills the corresponding
$i$ viewpoint constraint. Hence, this solution space can be interpreted
as an analytical \parencite{Tarabanis.1995b}, geometrical solution
with an infinite set of valid viewpoints to satisfy the regarded viewpoint
constraint. Moreover, the integration of multiple $\ACC$s spans the
jointed $\text{\ensuremath{\ACC}}$ denoted as $\CC{}{}{}$, where
all viewpoint constraints are simultaneously fulfilled. Figure \ref{fig:Motivation_VGP}
depicts a simplified representation of the $\AVGP$, $\ACC$s, and
overview of the most relevant components. 

In this context, the most significant challenge behind the conceptualization
of $\ACC$s lies in the generic and geometric formulation and characterization\footnote{Throughout this paper, we will use the term \emph{formulation} to
refer to the formal, mathematical definition of an individual $\ACC$.
On the contrary, the characterization addresses the concrete implementation
or computation of a $\ACC$ using a specific algorithm or method.} of diverse viewpoint constraints. This publication introduces nine
$\ACC$s corresponding to different viewpoint constraints (i.e., sensor
imaging parameters, feature geometry, kinematic errors, sensor accuracy,
occlusion, multisensors, multi-features, and robot workspace) aligned
to a consistent modeling framework to ensure their consistent integration. 

\subsection{Outline}

After providing an overview of the related work that has addressed
the $\AVPP$ and $\AVGP$ in Section \ref{sec:Related-Work}, our
work comprises four sections that describe the core concepts for formulating
and characterizing $\ACC$s of different viewpoint constraints. 

First, Section \ref{sec:Robot-Based-Vision-System} presents the domain
models of a generic $\ARVS$'s used through the present study. Section
\ref{sec:VS-Formulation} introduces the fundamental formulations
of the $\AVGP$ and $\text{\ensuremath{\ACC}}$s. Exploiting these
formulations, the fundamental $\text{\ensuremath{\ACC}}$ based on
the sensor imaging parameters and the feature position is characterized
in Section \ref{sec:Volumetric-Viewpoint-Space}. Using this core
$\text{\ensuremath{\ACC}}$, the geometrical formulations of the rest
of the viewpoint constraints and a strategy for their integration
are introduced in Section \ref{sec:Constraining-CV}. 

Finally, in Section \ref{sec:Evaluation} we assess the validity of
our formulations and characterization of $\text{\ensuremath{\ACC}}$s
and demonstrate its potential and generalization using a real $\ARVS$. 

\subsection{Contributions }

Our publication presents the fundamental concepts of a generic framework
comprising innovative and efficient formulations to compute valid
viewpoints based on $\ACC$s to solve the fundamental sub-problem
of the $\AVPP$, i.e. the $\AVGP$. The key contributions of this
paper are summarized as follows:
\begin{itemize}
\item Mathematical, model-based, and modular framework to formulate the
$\AVGP$ based on $\ACC$s and generic domain models 
\item Formulation of nine viewpoint constraints using linear algebra, trigonometry,
geometric analysis, and Constructive Solid Geometry ($\ACSG$) Boolean
operations, in particular:
\begin{itemize}
\item Efficient and simple characterization of $\ACC$ based on sensor frustum,
feature position and feature geometry
\item Generic characterization of $\ACC$s to consider bi-static nature
of range sensors extendable to multisensor systems
\end{itemize}
\item Exhaustive supporting material (surface models, manifolds of computed
$\ACC$s, rendering results) to encourage benchmark and further development
(see supplement material).
\end{itemize}
Additionally, we consider the following principal advantages associated
with the formulation of $\ACC$s:
\begin{itemize}
\item \textbf{Determinism, efficiency, and simplicity:} $\ACC$s can be
efficiently characterized using geometrical analysis, linear algebra,
and $\ACSG$ Boolean techniques.
\item \textbf{Generalization, transferability, and modularity:} $\ACC$s
can be seamlessly used and adapted for different vision tasks and
RVSs, including different sensor imaging sensors (e.g., stereo, active
light sensors) or even multiple range sensor systems.
\item \textbf{Robustness against model uncertainties: }Known model uncertainties
(e.g., kinematic model, sensor, or robot inaccuracies) can be explicitly
modeled and integrated while characterizing $\text{\ensuremath{\ACC}}$s.
If unknown model uncertainties affect a chosen viewpoint, alternative
solutions guaranteeing constraint satisfiability can be found seamlessly
within $\ACC$s.
\end{itemize}
In combination with a suitable strategy, $\ACC$s can be straightforwardly
integrated into a holistic approach for entirely solving the $\AVPP$.
The use of $\text{\ensuremath{\ACC}}$s within an adequate strategy
represents the second sub-problem of the $\AVPP$, which falls outside
the scope of this paper and will be handled within a future publication.

\section{Related Work\label{sec:Related-Work}}

Our study treats the $\AVGP$ as a sub-problem of the $\AVPP$. Since
most authors do not explicitly consider such a problem separation,
this section provides an overview of related research that addresses
the $\AVPP$ in general. In a broader sense, the $\AVPP$ can be even
categorized as a sub-problem of a more popular challenge within robotics,
i.e., the coverage path planning problem \parencite{Tan.2021}. 

Over the last three decades, the $\AVPP$ has been investigated within
a wide range of vision tasks that integrate an imaging device, but
not necessarily a robot, and require the computation of generalized
viewpoints. For a vast overview of the overall progress, challenges,
and applications of the $\AVPP$, we refer to the various surveys
\parencite{Tarabanis.1995,Scott.2003,Chen.2011,Mavrinac.2013,Kritter.2019,PeuzinJubert.2021}
that have been published. 

The approaches for viewpoint planning can be classified depending
on the knowledge required a priori about the $\ARVS$ to compute a
valid viewpoint. Thus, a rough distinction can be made between \emph{model-based}
and \emph{non-model-based} approaches \parencite{Scott.2002}. 

\subsection{Model-Based \label{subsec:Viewpoint-Space-Related-Work}}

Most of the model-based viewpoint planning methods can be roughly
differentiated between \emph{synthesis} and \emph{sampling-based}
(related terms: \emph{generate and test}) modeling approaches \parencite{Tarabanis.1995b}.
While synthesis approaches use analytical relationships to first characterize
a continuous or discrete solution space before searching for an optimal
viewpoint, sampling techniques are more optimization-oriented and
compute valid viewpoints using a set of objective functions. 

Since, the present study seeks to characterize a solution space, i.e.,
a $\ACC$, for each individual viewpoint constraint using analytical
and geometrical relationships, the mathematical foundation of our
framework can be classified as a model-based method following a synthesis
approach. Hence, this section focuses mainly on the related literature
following a similar approach.

\subsubsection{Synthesis }

Many of the reviewed publications considering model-based approaches
have built their theoretical framework based on set theory to formulate
either a continuous or discrete search space in a first step. Then,
in a second step, optimization algorithms are used to find valid viewpoints
within these search spaces and assess the satisfiability of the remaining
constraints that were not explicitly considered.

The concept of characterizing such topological search spaces, in our
work addressed as $\ACC$s (\emph{related terms: viewpoint space,
visibility map, visibility matrix, visibility volumes, imaging space,
scannability frustum, configuration space, visual hull}, \emph{search
space}), has been proposed since the first studies addressing the
$\AVPP$. Such a formulation has the advantage of providing a straightforward
comprehension and spatial interpretation of the general problem. 

One of the first seminal studies that considered the characterization
of a continuous solution space in $\mathbb{R}^{3}$ can be attributed
to the publication of \textcite{Cowan.1988}. In their work, they
introduced a model-based method for 2D sensors, which synthesized
analytical relationships to characterize a handful of constraints
geometrically: resolution, focus, field of view, visibility, view
angle, occluding regions, and in later works \parencite{Cowan.1989}
even constraints for placement of a lighting source. 

Based on the analytical findings provided by the previous works of
\textcite{Cowan.1989}, \textcite{Tarabanis.1995} introduced a model-based
sensor planning system called \emph{Machine Vision Planner} (MVP).
On the one hand, the MVP can be seen as a synthesis approach that
characterizes a feature-based occlusion-free region using surface
model decomposition \parencite{Tarabanis.1992,Tarabanis.1996}. On
the other hand, the authors posed the problem in the context of an
optimization setting using objective functions to find valid viewpoints
within the occlusion-free space that meet imaging constraints.

The MVP was extended by \textcite{Abrams.1999} for its use with an
industrial robot and moving objects. Their study addressed the drawbacks
(non-linearity and convergence guarantee) of the optimization algorithms
and opted to characterize 3D search spaces for the sensor's resolution,
field of view, and workspace of the robot. Although the authors could
not synthesize every constraint in the Euclidean space, they confirmed
the benefits of solving the problem in $\mathbb{R}^{3}$ instead of
optimizing equations for finding suitable viewpoints. Furthermore,
in a series of publications \emph{Reed et al.} \parencite{Reed.1998,Reed.2000}
extended some of the models introduced in the MVP and addressed the
characterization of a search space in $\mathbb{R}^{3}$ for range
sensors, which integrates imaging, occlusion, and workspace constraints.
Their study also proposed the synthesis of an imaging space based
on extrusion techniques of the surface models in combination with
the imaging parameters of the sensor. 

Another line of research within the context of model-based approaches
follow the works of \emph{Tarbox and Gottschlich}, which proposed
the synthesis of a discretized search space using \emph{visibility
matrices} to map the visibility between the solution space and the
surface space of the object. In combination with an efficient volumetric
representation of the object of interest using octrees, \textcite{Tarbox.1994}
and \textcite{Tarbox.1995} presented different algorithms based on
the concept of visibility matrices to perform automated inspection
tasks. The visibility matrices consider a discretized view space with
all viewpoints lying on a tessellated sphere with a fixed camera distance.
Analogously, under the consideration of further constraints, \parencite{Scott.2002,Scott.2009}
introduced the measurability matrix extending the visibility matrix
of \emph{Tarbox and Gottschlich} to three dimensions. Within his work,
Scott considered further sensor parameters, e.g., the shadow effect,
measurement precision, and the incident angle, which many others have
neglected. More recent works \parencite{Gronle.2016,Jing.2017,Mosbach.2021,Gospodnetic.2022},
confirmed the benefits of such an approach and used the concept of
visibility matrices for encoding information between a surface point
and a set of valid viewpoints. 

In the context of space discretization and feature-driven approaches,
further publications \parencite{Tarbox.1995,Trucco.1997,Pito.1999}
suggested the characterization of the positioning space for the sensor
using tessellated spheres to reduce the 6D sensor positioning problem
to a 2D orientation optimization problem. Similarly, \textcite{Stoel.2004}
and \textcite{Ellenrieder.2005} introduced the concept of visibility
maps to encode feature visibility mapped to a visibility sphere. Years
later, \textcite{Raffaeli.2013b} and \textcite{Koutecky.2016} considered
variations of this approach for their viewpoint planning systems.
The major shortcomings of techniques considering a problem reduction
is that most of them require a fixed working distance, which limits
their applicability for other sensors and reduces its efficiency for
the computation of multi-feature acquisition. 

In the context of laser scanners, other relevant works, for example,
\parencite{Lee.2000}, \parencite{Derigent.2006}, and \parencite{TekouoMoutchiho.2012}
also considered solutions to first synthesize a search space before
searching for feasible solutions. Also, the publication of \textcite{Park.2006}
needs to be mentioned, since besides \textcite{Scott.2009} it is
one of the few authors who considered viewpoint visibility for multisensor
systems using a lookup table.

\subsubsection{Sampling-Based}

Many other works \parencite{GonzalezBanos.2001,Chen.2004,Erdem.2006,Mavrinac.2015,Glorieux.2020}
do not rely on the explicit characterization of a search space and
assess the satisfiability of each viewpoint constraint individually
by sampling the search space using metaheuristic optimization algorithms,
e.g., simulated annealing or evolutionary algorithms. Such approaches
concentrate on the adequate and efficient formulation of objective
functions to satisfy the viewpoint constraints and find reasonable
solutions.

\subsection{Non-Model Based}

In contrast, \emph{non-model based }approaches require no a priori
knowledge; the object can be utterly unknown to the planning system.
In this case, online exploratory techniques based on the captured
data are used to compute the next best viewpoint \parencite{Pito.1999,Chen.2005,VasquezGomez.2014,Kriegel.2015,Lauri.2020}.
Most of these works focus on reconstruction tasks and address the
problem as the \emph{next-best-view} planning problem. Since our work
is considered a feature-driven approach requiring a priori knowledge
of the system, this line of research will be not further discussed. 

\subsection{Comparison and Need for Action \label{subsec:SA_Formulation-of-Constraints}}

Although over the last three decades, many works presented well-grounded
solutions to tackle the $\AVPP$ for individual applications, a generic
approach for solving the $\AVPP$ has not been established yet for
commercial or industrial applications nor research. Hence, we are
convinced that a well-founded framework, comprising a consistent formulation
of viewpoints constraints combined with a model-based synthesis approach
considering a continuous solution space, has the greatest potential
to search for viewpoints that efficiently satisfy different viewpoints
constraints. 

\textbf{\emph{Synthesis vs. Sampling }}Within the related works, we
have found recent publications following an explicit synthesis and
sampling techniques of solution spaces for the same applications.
Hence, a clear trend towards any of these model-based approaches could
not be identified. On the one hand, sampling methods can be especially
advantageous and computationally efficient within simple scenarios
considering few constraints. On the other hand, model uncertainties
and nonlinear constraints are more difficult to model using objective
functions and within multi-feature scenarios the computation efficiency
can be severely affected. Therefore, we think this problem can be
solved more efficiently based on $\ACC$s composed of explicit models
of all regarded viewpoint constraints within applications comprising
robot systems and partially known environments with modeling uncertainties.

\textbf{\emph{Continuous vs. Discrete Space}} Most of the latest research
have followed a synthesis approach based on visibility matrices or
visibility maps to encode the surface space and viewpoint space together
for a handful of applications and systems \parencite{Gronle.2016,Jing.2017,Koutecky.2016,Mosbach.2021,Gospodnetic.2022}.
Although these works have demonstrated the use of discrete spaces
to be practical and efficient, from our point of view, its major weakness
lies in the intrinsic limited storage capacity and processing times
associated with matrices. This limitation directly affects the synthesis
of the solution space considering just a fixed distance between sensor
and object. Moreover, in the context of $\ARVS$, limiting the robot's
working space seems to be in conflict with the inherent and most appreciated
positioning flexibility of robots. Due to these drawbacks and taking
into account that the fields of view and working distances of sensors
have been and will continue to increase, we consider that the discretization
of the solution space could become inefficient for certain applications
at some point. 

\textbf{\emph{Problem Formulation}} Many of the revised works considering
synthesized model-based approaches posed the $\AVGP$ formulation
on the fundamentals of set theory. However, our research suggests
that a consistent mathematical framework, which promotes a generic
formulation and integration of viewpoint constraints, has not been
appropriately placed. Hence, we consider that an exhaustive domain
modeling and consistent theoretical mathematical framework are key
elements to provide a solid base for a holistic and generic formulation
of the $\AVGP$.

\section{Domain Models of a Robot Vision System \label{sec:Robot-Based-Vision-System} }

This section outlines the generic domain models and minimal necessary
parameters of an $\ARVS$, including assumptions and limitations,
required to characterize the individual $\ACC$s in Sections \ref{sec:Volumetric-Viewpoint-Space}
and \ref{sec:Constraining-CV}. 

\begin{figure*}[t]
\begin{centering}
\includegraphics[width=1\textwidth]{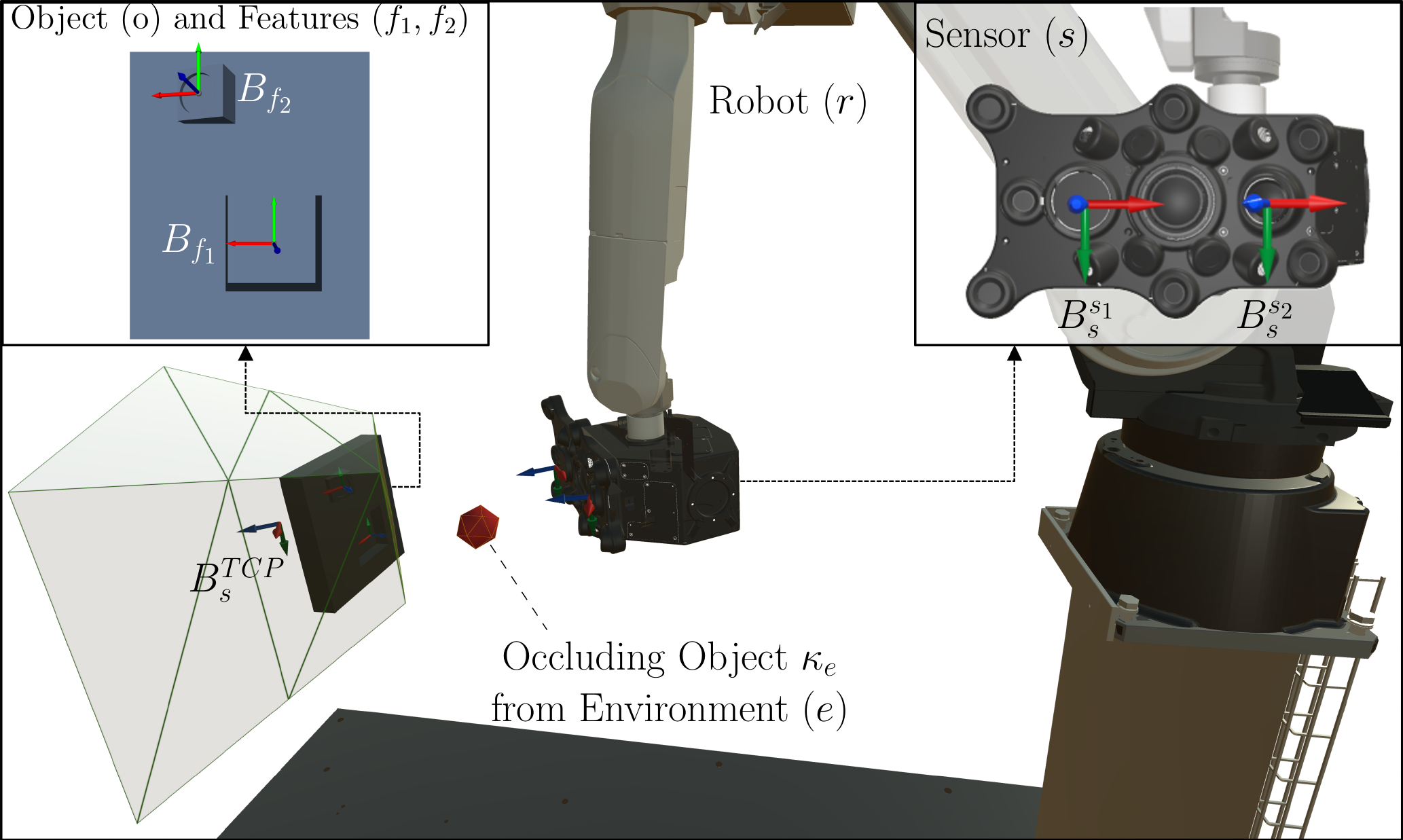}
\par\end{centering}
\caption{Overview of the $\protect\ARVS$ domains and kinematic model. \label{fig:Overview-of-RVS}}
\end{figure*}

We consider an $\ARVS$, a complex mechatronical system that comprises
the following domains: a range \emph{sensor} ($s$) that is positioned
by a \emph{robot} ($r$) to capture a\emph{ feature} ($f$) of an
\emph{object of interest} ($o$) enclosed within an \emph{environment}
($e$). Figure \ref{fig:Overview-of-RVS} provides an overview of
the RVS domains and some parameters described within this section. 

\subsection{General Notes}

\subsubsection{General Requirements\label{subsec:General-Requirements}}

The present paper follows a systematic and exhaustive formulation
of the $\AVGP$, the domains of an $\ARVS$, and the viewpoint constraints
to characterize $\ACC$s in a generic, simple, and scalable way. To
achieve this, and similar to previous studies \parencite{Scott.2002,Chen.2004,TekouoMoutchiho.2012},
throughout our framework following general requirements (GR) are considered:\textbf{
}generalization, computational efficiency, determinism, modularity
and scalability, and limited a priori knowledge. The given order does
not consider any prioritization of the requirements. A more detailed
description of the requirements can be found in Table \ref{tab:General-Requirements}.

\subsubsection{Terminology}

Based on our literature research, we have found that a common terminology
has not been established yet. The employed terms and concepts depend
on the related applications and hardware. To better understand our
terminology's relation to the related work and in an attempt towards
standardization, whenever possible, synonyms or related concepts are
provided. Please note that in some cases, the generality of some terms
is prioritized over their precision. This may lead to some terms not
corresponding entirely to our definition; therefore, we urge the reader
to study these differences before treating them as exact synonyms.

\subsubsection{Notation}

Our publication considers many variables to describe the $\ARVS$
domains comprehensively. \textit{\emph{To ease the identification
and readability of variables, parameters, vectors, frames, and transformations}}
we use \textit{\emph{the index notation given in Table \ref{tab:Notations}.}}
Moreover, all topological spaces are given in calligraphic fonts,
e.g., $\mathcal{V},\mathcal{P},\mathcal{I},\mathcal{C}$, while vectors,
matrices, and rigid transformations are bold. \textit{\emph{Table
\ref{tab:List-Symbols} provides an overview of the most frequently
used symbols.}}

\subsection{General models}

\subsubsection{Kinematic model\label{subsec:Kinematic-model}}

Each domain comprises a \emph{Kinematics} subsection to describe its
kinematic relationships. In particular, all necessary rigid transformations
(given in the right-handed system) are introduced to calculate the
sensor pose\textit{\emph{}}\footnote{\textit{\emph{The pose }}$\VECJ p{}{}{}{}$\textit{\emph{ of any element
is given by its translation }}$\VECJ t{}{}{}{}\in\mathbb{R}^{3}$\textit{\emph{
and a rotation component that can be given as a rotation matrix $\VECJ R{}{}{}{}\in\mathbb{R}^{3x3}$}}\textit{,
Z-Y-X}\textit{\emph{ }}\textit{Euler}\textit{\emph{ angles }}$\VECJ r{}{}{}{}=(\alpha^{z},\beta^{^{y}},\gamma^{x})^{T}$
or a quaternion \textit{\emph{$\VECJ q{}{}{}{}\in\mathbb{H}$. For
readability and simplicity purposes, we use mostly the Euler angles
representation throughout this paper. Considering the special orthogonal
$SO(3)\subset\mathbb{R}^{3x3}$, the pose $p\in SE(3)$ is given in
the special Euclidean ${SE(3)=\mathbb{R}^{3}\times SO(3)}$ }}\parencite{Waldron.2016}.} \textit{\emph{$\VECJ ps{}f{}\in SE(3)$}} in the feature's coordinate
system $B_{f}$:

\begin{equation}
\begin{aligned}\VECJ ps{}f{}= & \VECJ T{}{}fs\\
= & \VECJ Tf{}fo\cdot\VECJ Te{}ow\cdot\VECJ Te{}wr\cdot\VECJ Tr{}rs.
\end{aligned}
\end{equation}

It is assumed that all introduced transformations are roughly known
and given in the world ($w$) coordinate system or any other reference
system. Moreover, we also consider a summed alignment error $\epsilon_{e}$
in the kinematic chain to quantify the sensor's positioning inaccuracy
relative to a feature.

\subsubsection{Surface model}

A set of 3D\emph{ }surface models $\kappa\in K$ characterizes the
volumetric occupancy of all rigid bodies in the environment. The surface
models are not always explicitly mentioned within the domains. Nevertheless,
we assume that the surface model of any rigid body is required if
this collides with the robot or sensor or impedes the sensor's sight
to a feature. 

\subsection{Object}

This domain considers an \emph{object} ($o$) (related terms: object
of interest, workpiece, artifact-, measurement-, inspection- or test-object)
that contains the features to be acquired. 

\subsubsection{Kinematics }

The origin coordinate system of $o$ is located at frame $B_{o}$.
The transformation to the reference coordinate system is given in
the world coordinate system $B_{w}$ by $\VECJ Te{}wo$. 

\subsubsection{Surface Model}

Since our approach does not focus on the object but rather on its
features, the object may have an arbitrary topology. 

\subsection{Feature\label{subsec:Feature}}

A\emph{ feature ($f$) }(related terms: region, point or area of interest,
inspection feature, key point, entity, artifact) can be fully specified
considering its kinematic and geometric parameters, i.e., frame $B_{f}$
and set of surface points $G_{f}(L_{f})$, which depend on a set of
geometric dimensions $L_{f}$:\emph{
\begin{equation}
f:=(B_{f},G_{f}(L_{f})).
\end{equation}
}

\subsubsection{Kinematics }

We assume that the translation $\VECJ tf{}o{}$ and orientation\endnote{In the case that the feature's orientation is given by its minimal
expression, i.e., just the feature's surface normal vector $\VECJ nf{}o{}$.
The full orientation is calculated by letting the feature's normal
to be the basis z-vector $\VECJ efzo{}=\VECJ nf{}o{}$ and considering
the rest basis vectors $\VECJ efxo{}$ and $\VECJ efyo{}$ to be mutually
orthonormal.} $\VECJ rf{}o{}$ of the feature's origin is given in the object's
coordinate system $B_{o}$. Thus, the feature's frame is given as
follows:
\begin{equation}
B_{f}=\VECJ Tf{}o{}(\VECJ tf{}o{},\VECJ rf{}o{}).
\end{equation}

\subsubsection{Geometry\label{subsec:F-Geo}}

While a feature can be sufficiently described by its position and
normal vector, a broader formulation is required within many applications.
For example, dimensional metrology tasks deal with a more comprehensive
catalog of geometries, e.g., edges, pockets, holes, slots, and spheres. 

Thus, the present study considers explicitly the geometrical topology
of a feature and a more extensive model of it \parencite{Tarabanis.1996,Ellenrieder.2005}.
Let the feature topology be described by a set of geometric parameters,
denoted by $L_{f}$, such as the radius of a hole or a sphere or the
lengths of a square.

\subsubsection{Generalization and Simplification of the Feature Geometry \label{subsec:Simplification-Feature} }

Moreover, we consider a discretized geometry model of a feature comprising
a finite set of surface points corresponding to a feature $\VECJ{g_{f}}{}{}{}{}\in G_{f}$
with $\VECJ{g_{f}}{}{}{}{}\in\mathbb{R}^{3}$. Since our work primarily
focuses on 2D features, it is assumed that all surface points lie
on the same plane, which is orthogonal to the feature's normal vector
$\VECJ nf{}o{}$ and collinear to the z-axis of the feature's frame
$B_{f}$.

\begin{figure}[tbh]
\begin{centering}
\begin{minipage}[t]{0.8\columnwidth}%
\begin{center}
\includegraphics{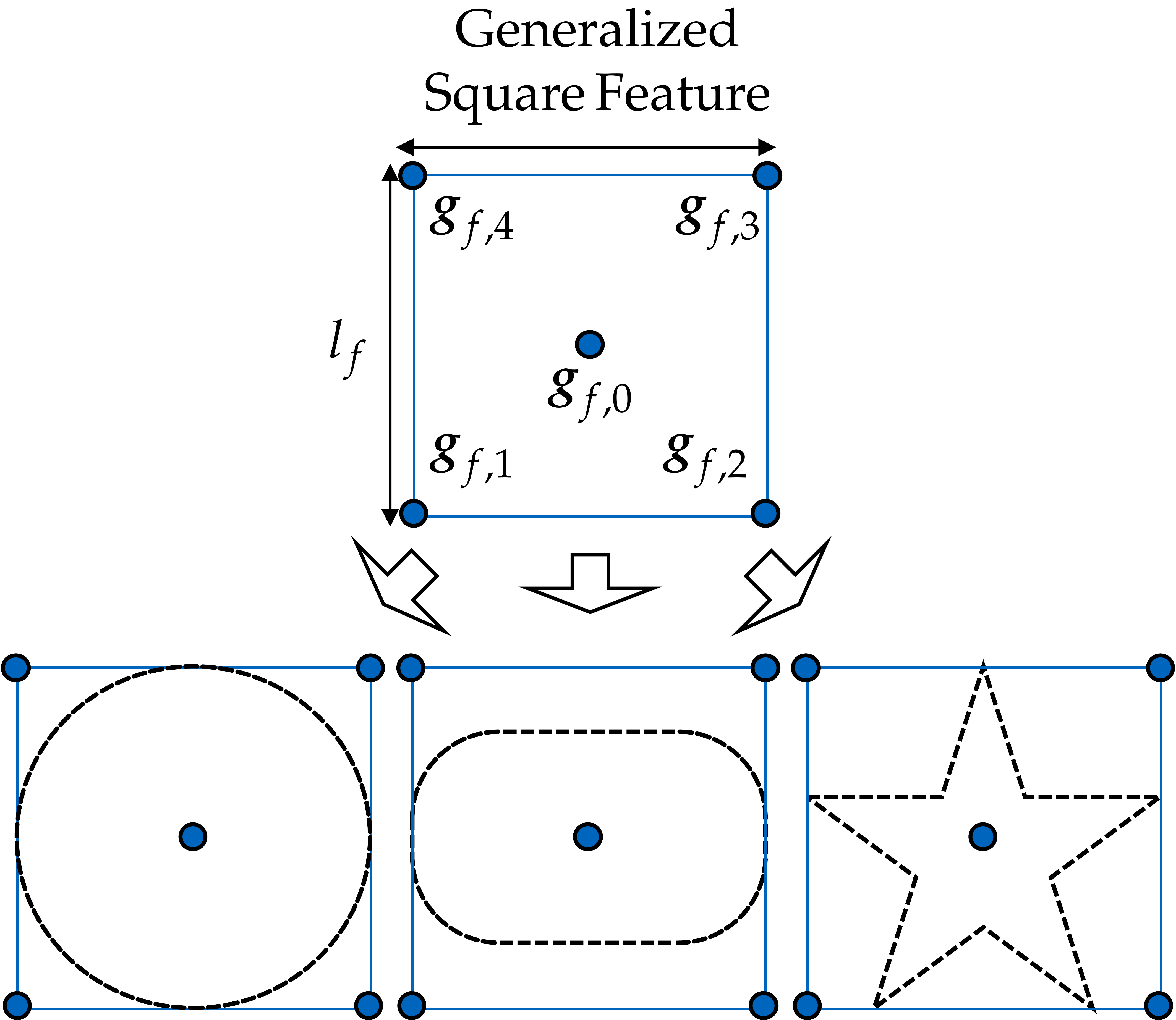}
\par\end{center}%
\end{minipage}
\par\end{centering}
\caption{A square feature with the length $l_{f}$ comprising five surface
points $\protect\VECJ{g_{f}}{,c}{}{}{},$ is used to generalize any
feature topology, e.g, a circle, a slot, or a star (complex geometry)
.\label{fig:Generalized-Features}}
\end{figure}

Towards providing a more generic feature's model, the topology of
all features is approximated using a square feature with a unique
side length of $\{l_{f}\}\in L_{F}$ and five surface points $\VECJ{g_{f}}{,c}{}{}{},\:c=\{0,1,2,3,4\}$
at the center and at the four corners of the square. Figure \ref{fig:Generalized-Features}
visualizes this simplification to generalize diverse feature geometries.

\subsection{Sensor\label{subsec:Sensor}}

We consider a \emph{sensor} ($s$) (related terms: range camera sensor,
3D sensor, imaging system): a self-contained acquisition device comprising
at least two imaging devices $\{s_{1},s_{2}\}\in\widetilde{S}$ (e.g.,
two cameras or a camera and a lighting source) capable of computing
a range image containing depth information. Such sensors can be classified
by the principles used to acquire this type of depth information,
e.g., triangulation, intensity, or time of flight \parencite{Beyerer.2016}.
The present work does not explicitly distinguish between these acquisition
principles. Moreover, this subsection outlines a generic and minimal
sensor model that is in line with our framework. Note that even though
the present report focuses primarily on range sensors, the models
can also be considered for single imaging devices. 

\subsubsection{Kinematics\label{subsec:S-Kinematics}}

The sensor's kinematic model considers the following relevant frames:
$B_{s}^{TCP}$,$B_{s}^{s_{1}}$, and $B_{s}^{s_{2}}$. Taking into
account the established notation for end effectors within the robotics
field, we consider that the frame $B_{s}^{TCP}$ lies at the sensor's
\emph{TCP.} We assume that the frame of the TCP is located at the
geometric center of the frustum space and that the rigid transformation
$\VECJ Ts{}{ref}{TCP}$ to  a reference frame such as the sensor's
mounting point is known. 

Additionally, we consider that frame $B_{s}^{s_{1}}$ lies at the
reference frame of the first imaging device that correspond to the
imaging parameters $I_{s}$. We assume that the rigid transformation
$\VECJ Ts{}{ref}{s_{1}}$ between the sensor lens and a known reference
frame are also known. $\VECJ Ts{}{ref}{s_{2}}$ provides the transformation
of the second imaging device at the frame $B_{s}^{s_{2}}$. The second
imaging device $s_{2}$ might be a second camera considering a stereo
sensor or the light source origin in an active sensor system.

\subsubsection{Frustum space \label{subsec:Frustum-Space}}

\begin{figure}[t]
\begin{centering}
\includegraphics{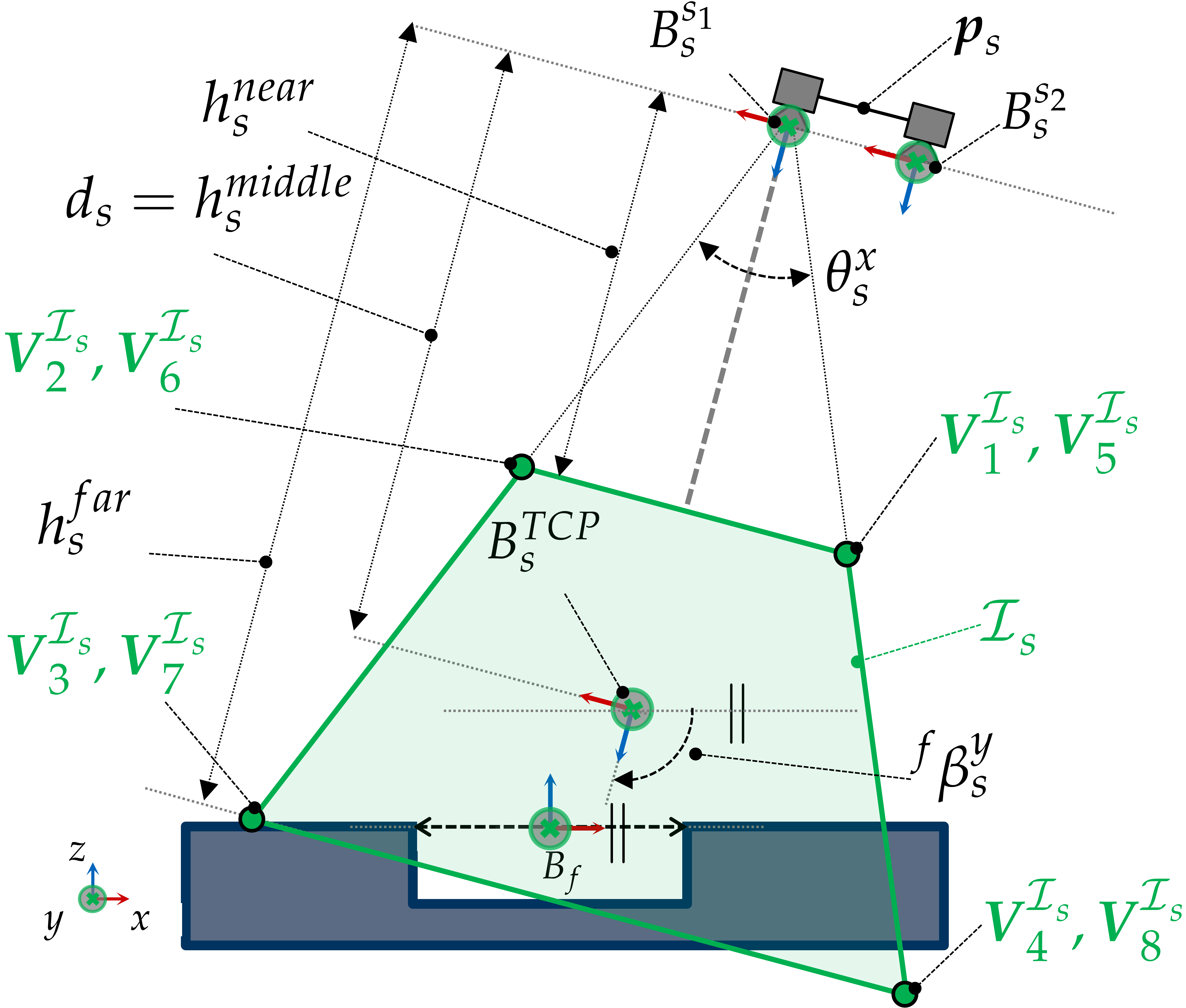}
\par\end{centering}
\caption{Detailed kinematic and imaging model of the sensor in the $x\text{\text{-}}z$
plane. The frustum space $\protect\CF$ is spanned by the imaging
parameters of the sensor $(d_{s},h_{s}^{near},h_{s}^{far},\theta_{s}^{x},\psi_{s}^{y})\in I_{s}$
considering a sensor pose $\protect\SP{}$. The $\protect\ACM$ is
described by a minimum of eight vertices $\protect\VECJ V{1-8}{\protect\CF}{}{}$
(note that in this 2D view the vertices 5–8 lie on the far $x\text{\text{-}}z$
plane and that the FOV angle $\psi_{s}^{y}$ is not illustrated).
\label{fig:Sensor-Model}}
 
\end{figure}

The frustum space $\ACM$ (related terms: visibility frustum, measurement
volume, field-of-view space, sensor workspace) is described by a set
of different sensor imaging parameters $I_{s}$, such as the depth
of field%
{} $d_{s}$ and the horizontal and vertical field of view (FOV) angles
$\theta_{s}^{x}$ and $\psi_{s}^{y}$ %
. Alternatively, some sensor manufacturers may also provide the dimensions
and locations of the near $h_{s}^{near}$, middle $h_{s}^{middle}$,
and far $h_{s}^{far}$ viewing planes of the sensor. The sensor parameters
$I_{s}$ allow to just describe the topology of the $\ACM$. To fully
characterize the topological space in the special Euclidean, the sensor
pose $\VECJ ps{}{}{}$ must be considered:
\begin{equation}
\begin{aligned}\CF:=\CF(\VECJ ps{}{}{},I_{s})= & \{\VECJ ps{}{}{}\in SE(3),\\
 & (d_{s},h_{s}^{near},h_{s}^{far},\theta_{s}^{x},\psi_{s}^{y})\in I_{s}\}
\end{aligned}
.\label{eq:C-I}
\end{equation}
The $\ACM$ can be straightforwardly calculated based on the kinematic
relationships of the sensor and the imaging parameters. The resulting
3D manifold $\CF$ is described by its vertices $\VECJ Vk{\CF}{}{}:=\VECJ Vk{}{}{}(\CF)=(V_{k}^{x},V_{k}^{y},V_{k}^{z})^{T}$
with $k=1,\dots,l$ and corresponding edges and faces. We assume that
the origin of the frustum space is located at the TCP frame, i.e.,
${\fr{}{}Bs{\mathcal{I}_{s}}=\fr{}{}Bs{TCP}}$. The resulting shape
of the $\ACM$ usually has the form of a square frustum. Figure \ref{fig:Sensor-Model}
visualizes the frustum shape and the geometrical relationships of
the $\ACM$. 

\subsubsection{Range Image}

A range image (related terms: 3D measurement, 3D image, depth image,
depth maps, point cloud) refers to the generated output of the sensor
after triggering a measurement action. A range image is described
as a collection of 3D points denoted by $\VECJ gs{}{}{}\in\mathbb{R}^{3}$,
where each point corresponds to a surface point of the measured object. 

\subsubsection{Measurement accuracy\label{subsec:Measurement-accuracy}}

The measurement accuracy depends on various sensor parameters and
external factors and may vary within the frustum space \parencite{Scott.2009}.
If these influences are quantifiable,\endnote{\textcite{Park.2006} proposed a method based on a Look-Up Table to
specify quality disparities within a frustum.} an accuracy model can be considered within the computation of the
$\text{\ensuremath{\ACC}}$. 

\subsubsection{Sensor Orientation\label{subsec:Sensor-Orientation}}

When choosing the sensor pose for measuring an object's surface point
or a feature, additional constraints must be fulfilled regarding its
orientation. One fundamental requirement that must be satisfied to
guarantee the acquisition of a surface point is the consideration
of the incidence angle $\fr f{}{\varphi}s{}$ (related terms: inclination,
acceptance, view, or tilt angle). This angle is expressed as the angle
between the feature's normal $\VECJ nf{}{}{}$ and the sensor's optical
axis ($z$-axis) $\VECJ esz{}{}$ and can be calculated as follows:
\begin{equation}
\fr f{}{\varphi}s{max}>\lvert\fr f{}{\varphi}s{}\rvert,\fr f{}{\varphi}s{}=\arccos\left(\dfrac{\VECJ nf{}{}{}\cdot\VECJ esz{}{}}{\lvert\VECJ nf{}{}{}\rvert\cdot\lvert\VECJ esz{}{}\rvert}\right).\label{eq:IncidenceCondition}
\end{equation}

The maximal incidence angle $\fr f{}{\varphi}s{max}$ is normally
provided by the sensor's manufacturer\endnote{If the maximal angle is not given in the sensor specifications, some
works have suggested empirical values for different systems. \textcite{Biagio.2017}
propose a maximum angle of $60{^\circ}$. \textcite[p. 84]{Bertagnolli.2006}
suggests $45{^\circ}$ as the upper limit for a structured light sensor,
while \textcite{Raffaeli.2013} propose a tilt angle of $30{^\circ}$
to $50{^\circ}$. It is important to note that for image-based sensors,
such as stereo or structured light sensors, the sensor's optical axis
is constant all over the frustum. In contrast, laser scanners must
consider a variable incidence angle while sweeping over an object.}. The incidence angle can also be expressed on the basis of the Euler
angles (pan, tilt) around the x- and y-axes: $\fr f{}{\varphi}s{}(\fr f{}{\beta}sy,\fr f{}{\gamma}sx)$.

Furthermore, the rotation of the sensor around the optical axis is
given by the Euler angle $\alpha_{s}^{z}$ (related terms: swing,
twist). Normally, this angle does not directly influence the acquisition
quality of the range image and can be chosen arbitrarily. Nevertheless,
depending on the lighting conditions or the position of the light
source considering active systems, this angle might be more relevant
and influence the acquisition parameters of the sensor, e.g., the
exposure time. Additionally, if the shape of the frustum is asymmetrical,
the optimization of $\alpha_{s}^{z}$ should be considered. 

\subsection{Robot }

The robot (related terms: manipulator, industrial robot, positioning
device) has the main task to position the sensor to acquire a range
image.

\subsubsection{Kinematics and Workspace}

The robot base coordinate frame is placed at $\fr{}{}Br{}$. We assume
that the rigid transformations between the robot basis and the robot
flange, $\VECJ Tr{}r{f_{r}}$, and between the robot flange and the
sensor, $\VECJ Tr{}{f_{r}}s$, are known. We also assume that the
Denavit-Hartenberg (DH) parameters are known and that the rigid transformation
$\VECJ Tr{}r{f_{r}}(DH)$ can be calculated using an inverse kinematic
model. The sensor pose in the robot's coordinate system is given by
\begin{equation}
\VECJ ps{}r{}=\VECJ Tr{}rs=\VECJ Tr{}r{f_{r}}\cdot\VECJ Tr{}{f_{r}}s.
\end{equation}

The robot workspace is considered to be a subset in the special Euclidean,
thus $\mathcal{W}_{r}\subseteq SE(3)$. This topological space comprises
all reachable robot poses to position the sensor $\VECJ ps{}r{}\in\mathcal{W}_{r}$. 

\subsubsection{Robot Absolute Position Accuracy \label{subsec:Robot-Accuracy}}

It is assumed that the robot has a maximal absolute pose accuracy
error of $\epsilon_{r}$ in its workspace and that the robot repeatability
is much smaller than the absolute accuracy, hence, it is not further
considered.

\subsection{Environment}

The environment domain comprises models of remaining components that
were not explicitly included by other domains. Particularly, we consider
all other rigid bodies that may collide with the robot or affect the
visibility of the sensor, e.g., fixtures, external axes, robot cell
components. Thus, if the environment domain comprises rigid bodies,
these must be included in the set of surface models $\fr{}{}{\kappa}e{}\in K$.

\subsection{General assumptions and limitations\label{subsec:Assumptions}}

The previous subsections have introduced the formal models and parameters
to characterize an RVS. Hereby, we present some general assumptions
and limitations considered within our work. %

\begin{itemize}
\item \textbf{Sensor compatibility with feature geometry:} Our approach
assumes that a feature and its entire geometry can be captured with
a single range image. 
\item \textbf{Range Image Quality:} The sensor can acquire a range image
of sufficient quality. Effects that may compromise the range image
quality and have not been previously regarded are neglected, e.g.,
measurement repeatability, lighting conditions, reflection effects,
and random sensor noise.
\item \textbf{Sensor Acquisition Parameters:} Our work does not consider
the optimization of acquisition sensor parameters such as exposure
time, gain, and image resolution, among others.
\item \textbf{Robot Model:} Since we assumed that a range image can just
be statically acquired, a robot dynamics model is not contemplated.
Hence, constraints regarding velocity, acceleration, jerk, or torque
limits are not considered within the scope of our work. 
\end{itemize}

\section{Problem Formulation \label{sec:VS-Formulation}}

This section first introduces the concept of generalized viewpoints
and briefly describes the viewpoint constraints considered within
the scope of our work. Then the modularization of the $\AVPP$ and
formulation of the $\AVGP$ as a geometric problem are introduced
to understand the placement of the present study. In Subsection \ref{subsec:VS-Formulation}
$\ACC$s are introduced within the context of \emph{Configuration
Spaces} as a practical and simple approach for solving the $\AVGP$.
Moreover, considering that various viewpoint constraints must be satisfied
to calculate a valid viewpoint, we outline the reformulation of the
$\AVGP$ based on $\ACC$s within the framework of \emph{Constraint
Satisfaction Problems. }

\subsection{Viewpoint and $\protect\ACV$ \label{subsec:Viewpoint}}

Although, the concept of generalized viewpoints has been introduced
by some of the related works\footnote{The concept of a generalized viewpoint was first introduced by \textcite{Tarabanis.1995}
and was later used by \textcite{Scott.2002}  and \textcite{Chen.2004}.}, there seems to be no clear definition of a viewpoint\emph{ }$v$.
Hence, in this study, considering a feature-centered formulation we
define a viewpoint being a triple of following elements: a sensor
pose $\VECJ ps{}{}{}\in SE(3)$ to acquire a feature $f\in F$ considering
a set of viewpoint constraints $\widetilde{C}$ from any domain of
the $\ARVS$:
\[
v:=(f,\VECJ ps{}{}{},\widetilde{C}).
\]

Additionally, we consider that any viewpoint that satisfies all constraints
is an element of the viewpoint space ($\ACV$):
\begin{equation}
v\in\CV{}{}.\label{eq:Viewpoint-Base}
\end{equation}
Hence, the $\ACV$ can be formally defined as a tuple comprising a
feature space denoted by a feature set $F$ and the $\ACC$ $\CC{}{}F(\widetilde{C})$,
which satisfies all spatial viewpoint constraints to position the
sensor:
\[
\text{\ensuremath{\CV{}{}}}:=(F,\CC{}{}F(\widetilde{C})).
\]

Note that within this publication, we just consider spatially viewpoint
constraints affecting the placement of the sensor. As given by the
limitations of our work in Subsection \ref{subsec:Assumptions}, additional
sensor setting parameters are not explicitly addressed. Nevertheless,
for purposes of correctness and completeness, let these constraints
be denoted by $\widetilde{C}_{s}$, then Equation \ref{eq:Viewpoint-Base}
can be extended as follows:
\[
(f,\VECJ ps{}{}{},\widetilde{C},\widetilde{C}_{s})\in\CV{}{}.
\]

\subsection{Viewpoint Constraints\label{subsec:Viewpoint-Constraints} }

To provide a comprehensive model of a generalized viewpoint and assess
its validity, it is necessary to formulate a series of viewpoint constraints.
Hence, we propose an abstract formulation of the viewpoint constraints
needed to acquire a feature successfully. The set of viewpoint constraints
$c_{i}\in\widetilde{C},i=1,\dots,j$ comprises all constraints $c_{i}$
affecting the positioning of the sensor; hence, the validity of a
viewpoint candidate. Every constraint $c_{i}$ can be regarded as
a collection of domain variables of the regarded $\ARVS$, e.g., the
imaging parameters $I_{s}$, the feature geometry length $l_{f}$,
the maximal incidence angle $\fr f{}{\varphi}s{max}$. 

This subsection provides a general description of the constraints;
a more comprehensive formulation and characterization are handled
individually within Sections \ref{sec:Volumetric-Viewpoint-Space}
and \ref{sec:Constraining-CV}. An overview of the viewpoint constraints
considered in our work is given in Table \ref{tab:VCs}.

Although some related studies \parencite{Cowan.1988,Tarbox.1995,Scott.2002,Chen.2004}
also considered similar constraints, the main differences to our formulations
are found in their explicit characterization and integration with
other constraints. While some of these works assess a viewpoint's
validity in a reduced 2D space or sampled space, our work focuses
on characterizing each constraint explicitly in a higher dimensional
and continuous space.

\begin{table*}[tbh]
\caption{Overview and description of the viewpoint constraints considered in
our work.\label{tab:VCs}}

\centering{}\begin{center}
\resizebox{0.95\textwidth}{!}{
\begin{tabular}{>{\raggedright}p{0.25\textwidth}>{\raggedright}p{0.7\textwidth}}
\toprule\textbf{ Viewpoint Constraint} & \textbf{Description}\tabularnewline
\midrule 1. Frustum space\textbf{ } & The most restrictive and fundamental constraint is given by the imaging
capabilities of the sensor. This constraint is fulfilled if at least
the feature's origin lies within the frustum space (cf. Subsection
\ref{subsec:Frustum-Space}).\tabularnewline
\midrule 2. Sensor Orientation  & Due to specific sensor limitations, it is necessary to ensure that
the maximal permitted incidence angle between the optical axis and
the feature normal lies within an specified range, see Equation \eqref{eq:IncidenceCondition}.\tabularnewline
\midrule 3. Feature Geometry  & This constraint can be considered an extension of the first viewpoint
constraint and is fulfilled if all surface points of a feature can
be acquired by a single viewpoint, hence lying within the image space.\tabularnewline
\midrule 4. Kinematic Error  & Within the context of real applications, model uncertainties affecting
the nominal sensor pose compromise a viewpoint's validity. Hence,
any factor, e.g., kinematic alignment (Subsection \ref{subsec:Kinematic-model}),
robot's pose accuracy (Subsection \ref{subsec:Robot-Accuracy}), affecting
the overall kinematic chain of the $\ARVS$ must be considered.\tabularnewline
\midrule 5. Sensor Accuracy  & Acknowledging that the sensor accuracy may vary within the sensor
image space (see Subsection \ref{subsec:Measurement-accuracy}), we
consider that a valid viewpoint must ensure that a feature must be
acquired within a sufficient quality.\tabularnewline
\midrule 6. Feature Occlusion  & A viewpoint can be considered valid if a free line of sight exists
from the sensor to the feature. More specifically, it must be assured
that no rigid bodies are blocking the view between the sensor and
the feature.\tabularnewline
\midrule 7. Bistatic Sensor and Multisensor  & Recalling the bistatic nature of range sensors, we consider that all
viewpoint constraints must be valid for all lenses or active sources.
Furthermore, we also extend this constraint for considering a multisensor
$\ARVS$ comprising more than one range sensor.\tabularnewline
\midrule 8. Robot Workspace  & The workspace of the whole $\ARVS$ is limited primarily by the robot's
workspace. Thus, we assume that a valid viewpoint exists if the sensor
pose lies within the robot workspace.\tabularnewline
\midrule 9. Multi-Feature & Considering a multi-feature scenario, where more than one feature
can be acquired from the same sensor pose, we assume that all viewpoint
constraints for each feature must be satisfied within the same viewpoint
\tabularnewline\bottomrule
\end{tabular}}
\end{center}
\end{table*}

\subsection{Modularization of the Viewpoint Planning Problem}

The necessity to break down the $\AVPP$ into two sub-problems can
be better understood by considering the following minimal problem
formulation based on \textcite{Tarbox.1995}:
\begin{problem}
\label{PF:VPP-F1}\textit{ }How many viewpoints are necessary to acquire
a given set of features\textit{}\footnote{\textit{\emph{Although our work focuses on a feature-based approach,
the concept of features can be also be extended to surface points
or areas of interest.}}}? 
\end{problem}
We believe that considering a multi-stage solution to tackle the $\AVPP$
can reduce its complexity and contribute to a more efficient solution.
Thus, in the first step, we consider the modularization of the $\AVPP$
and address its two fundamental problems separately: the $\AVGP$
and the Set Covering Problem ($\ASCP$). 

First, we attribute to the $\AVGP$ the computation of valid viewpoints
to acquire a single feature considering a set of various viewpoint
constraints. Moreover, in the context of multi-feature scenarios and
presuming that all features cannot be acquired using a single viewpoint,
the efficient selection of more viewpoints becomes necessary to complete
the vision task, with which arises a new problem, i.e., the $\ASCP$.

This paper concentrates on the comprehensively formulation of the
$\AVGP$, so this problem is discussed more extensively in the following
sections.

\subsection{The Viewpoint Generation Problem}

The $\AVGP$ (related terms: optical camera placement, camera planning)
and concept of viewpoints can be better understood considering a proper
formulation:
\begin{problem}
\label{PF:VGP-F1} Which is a valid viewpoint $v$ to acquire a feature
$f$ considering a set of viewpoint constraints $\widetilde{C}$?
\end{problem}
A viewpoint $v$ exists, if there is at least one sensor pose $\SP{}$
that can capture a feature $f$ and only if all $j$ viewpoint constraints
$\widetilde{C}$ are fulfilled. The most straightforward way to find
a valid viewpoint for Problem \ref{PF:VGP-F1} is to assume an ideal
sensor pose $\VECJ ps0{}{}$ and assess its satisfiability against
each constraint using a binary function $h_{i}:(\VECJ ps0{}{},\c i)\rightarrow true$.
If the sensor pose fulfills all $j$ constraints, the viewpoint is
valid. Otherwise, another sensor pose must be chosen and the process
must be repeated until a valid viewpoint is found. The mathematical
formulation of such conditions are expressed as follows:
\begin{equation}
\begin{aligned}v= & \{(f,\VECJ ps0{}{},\widetilde{C})\in\CV{}{}\\
 & \mid f\in F,\VECJ ps0{}{}\in SE(3),\\
 & \stackrel[\c i\in\widetilde{C}]{}{\bigcap}h_{i}:(\VECJ ps0{}{},\c i)\rightarrow true\}.
\end{aligned}
\label{eq:Viewpoint}
\end{equation}

The formulation of a generalized viewpoint as given by Equation \eqref{eq:Viewpoint}
can be considered one of the most straightforward formulations to
solve the $\AVGP$, if for each viewpoint constraint, a Boolean condition
can be expressed. For instance, by introducing such cost functions
for different viewpoint constraints, several works \parencite{Beasley.1996,Chen.2004,Erdem.2006,Mittal.2007,Scott.2009,Mavrinac.2013,Kaba.2017}
demonstrated that optimization algorithms (e.g., greedy, genetic,
or even reinforcement learning algorithms) can be used to find local
and global optimal solutions within polynomial times. 

\subsection{$\protect\AVGP$ as a geometrical Problem in the context of Configuration
Spaces \label{subsec:VS-Formulation}}

Although the generalized viewpoint model as given by Equation \eqref{eq:Viewpoint}
yields a sufficient and generic formulation to solve the $\AVGP$,
this formulation is inefficient considering real applications with
model uncertainties. System modeling inevitably involves discrepancies
between virtual and real-world models, particularly within dynamically
changing environments.\textbf{ }Due to such model inconsistencies,
considering optimal viewpoints to acquire a feature could be regarded
as ineffective and inefficient in some applications. Hence, in our
opinion, it is more reasonable to treat the $\AVGP$ as a multi-dimensional
problem and to consider multiple valid solutions throughout its formulation. 

A sound solution for the $\AVGP$ will require characterizing a continuous
topological space comprising multiple solutions that allow deviating
from an optimal solution and efficiently choosing an alternative viewpoint.
This challenge embodies the core motivation of our work to formulate
and characterize $\ACC$s.

If the $\AVGP$ can be handled as a spatial problem that can be solved
geometrically, we refer to the use of configuration spaces, as introduced
by \textcite{LozanoPerez.1990} and \textcite{Latombe.1991} and exhaustively
studied by \textcite{LaValle.2006}, in the well-studied motion planning
field for solving geometrical\emph{ path planning problems.} 
\begin{quotation}
\emph{``Once the configuration space is clearly understood, many
motion planning problems that appear different in terms of geometry
and kinematics can be solved by the same planning algorithms. This
level of abstraction is therefore very important.''}\parencite{LaValle.2006}
\end{quotation}
In our work, we use the general concepts of configuration spaces based
on the formulation of topological spaces to characterize the manifold
spanned by a viewpoint constraint—the $\text{\ensuremath{\ACC}}$.

\subsection{$\protect\AVGP$ with ideal $\protect\ACC$s \label{subsec:VSC-Formulation-1}}

\begin{figure}[t]
\begin{centering}
\includegraphics{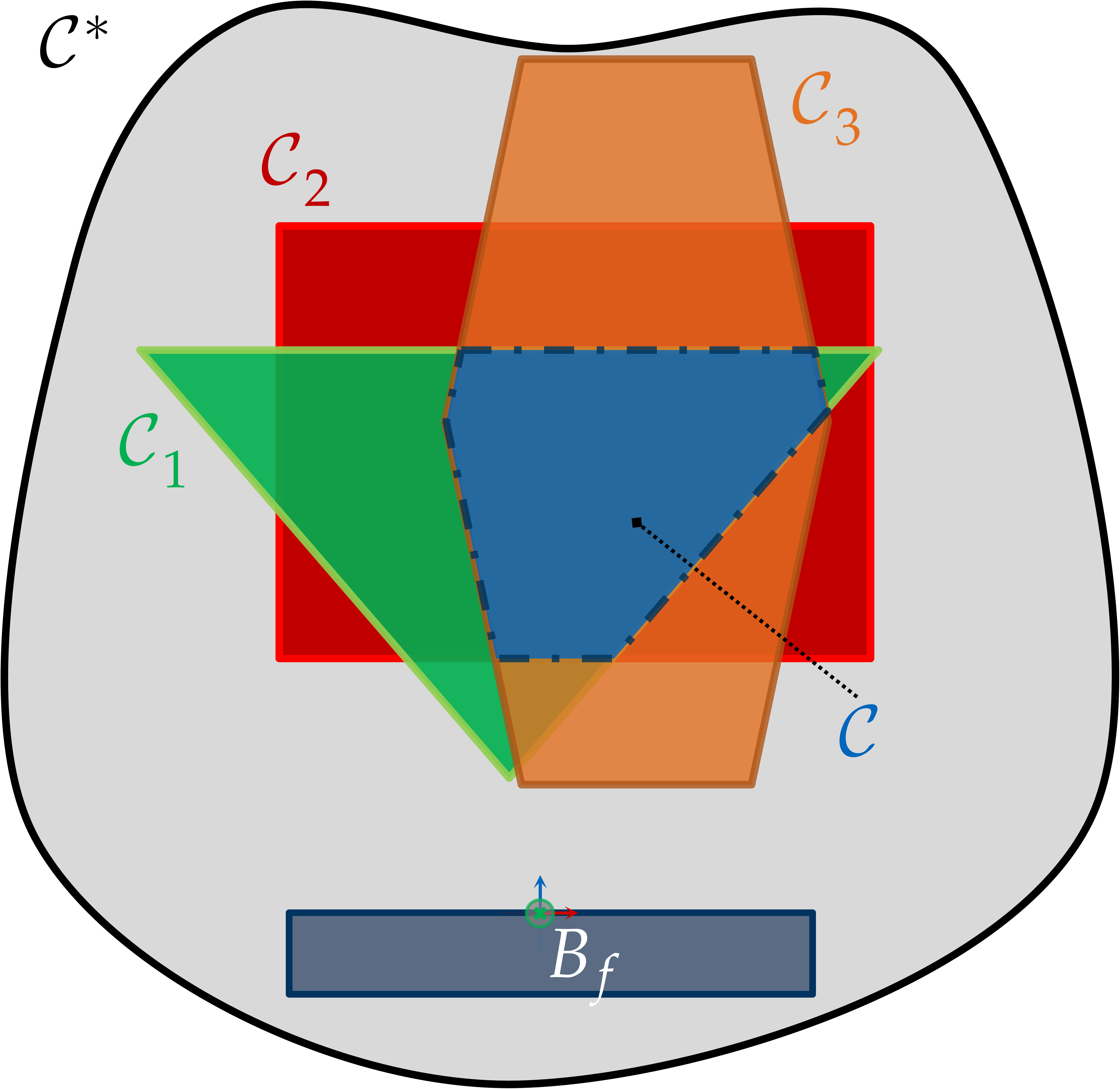}\caption{Abstract and simplified 2D representation of the ideal $\protect\ACC$
$\protect\CC{}*{}$ without viewpoint constraints, if viewpoint constraints
are considered, the intersection of the corresponding $\protect\ACC$s,
e.g., $\protect\CC 1{}{}$, $\protect\CC 2{}{}$, $\protect\CC 3{}{}$,
forms the $\protect\ACC$ $\protect\CC{}{}{}$. \label{fig:Constrained-Space}}
\par\end{centering}
\end{figure}

Following the notation and concepts behind the modeling of configuration
spaces, we first consider a modified formulation of Problem \ref{PF:VGP-F1}
and assume an ideal system (i.e., sensor with an infinite field of
view, without occlusions and neglecting any other constraint) for
introducing some general concepts:
\begin{problem}
\label{PF:VGP-F2} Which is the ideal $\text{\ensuremath{\ACC}}$
$\CC{}*{}$ to acquire a feature $f$? 
\end{problem}
Sticking to the notation established within the motion planning research
field, let us first consider an ideal, unconstrained space denoted
as $\CC{}*{}\subseteq SE(3)$ 
\begin{align}
\CC{}*{}= & \mathbb{R}^{3}\times SO(3)\nonumber \\
= & \{\SP{}\in\CC{}*{},\SP{}\in SE(3)\}\label{eq:C-P}\\
= & \{\SP{}(\VECJ ts{}{}{},\text{\ensuremath{\VECJ rs{}{}{}}})\in\CC{}*{}\nonumber \\
 & \mid\VECJ ts{}{}{}\in\mathbb{R}^{3},\text{\ensuremath{\VECJ rs{}{}{}}}\in SO(3)\},\nonumber 
\end{align}

which is spanned by the Euclidean Space $\mathbb{R}^{3}$ and the
special orthogonal group $SO(3)$, and holds all valid sensor poses
$\SP{}$ to acquire a feature $f$.  An abstract representation of
the unconstrained $\ACC$ $\CC{}*{}$ is visualized in Figure \ref{fig:Constrained-Space}.

\subsection{$\protect\AVGP$ with $\protect\ACC$s \label{subsec:VSC-Formulation}}

The ideal $\ACC$ as given by Equation \ref{eq:C-P} considers a sufficient
generic model that spans an ideal solution space to solve the $\AVGP$.
Assuming a non-ideal $\ARVS$ where a viewpoint must satisfy a handful
of requirements, an extended formulation of the $\text{\ensuremath{\ACC}}$
admitting viewpoint constraints is introduced within this subsection.

\subsubsection{Motivation}

The $\AVGP$ recalls the formulation of decision problems, a class
of computational problems, which has been widely researched within
different applications. Inspired by other research fields dealing
with artificial intelligence and optimization of multi-domain applications,
we observed that decision problems, including multiple constraints,
can be well formulated under the framework of Constraint Satisfaction
Problems ($\ACSP$) \parencite{Ghallab.2004}. This category of problems
does not consider an explicit technique to formulate the regarded
constraints. Moreover, a consistent, declarative, and simple representation
of the domain's constraints can be decisive for their efficient resolution
\parencite{Fruhwirth.2011}. 

\subsubsection{Formulation \label{subsec:CV-Formulation}}

Formulating the $\AVGP$ as a $\ACSP$ requires a proper formulation
to consider viewpoint constraints in the first step; hence, let \nameref{PF:VGP-F2}
be extended by the following:
\begin{problem}
\label{PF:VGP-F3} Which is the $\text{\ensuremath{\ACC}}$ $\CC{}{}{}$
spanned by a set of viewpoint constraints $\widetilde{C}$ to acquire
a feature $f$?
\end{problem}
The $\ACC$ denoted as
\begin{equation}
\CC{}{}{}:=\CC{}{}{}(f,\widetilde{C})\label{eq:CC}
\end{equation}
 can be understood as the topological space that all viewpoint constraints
from the set $\widetilde{C}$ span in the special Euclidean so that
the sensor can capture a feature $f$ fulfilling all of these constraints.
The $\ACC$ that a single viewpoint constraint $\c i\in\widetilde{C}$
spans, is analogously given:
\[
\CC i{}{}:=\CC i{}{}(f,\c i).
\]

To guarantee a consistent formulation and integration of various viewpoint
constraints, within our framework we consider the following characteristics
for the formulation of $\ACC$s: 
\begin{enumerate}
\item If an $i$ constraint, $\c i$, can be spatially modeled, there exists
a topological space denoted as $\CC i{}{}$, which can be ideally
formulated as a proper subset of the special Euclidean: 
\[
\CC i{}{}\subseteq SE(3).
\]
In a broader definition, we consider that the topological space for
each constraint is spanned by a subset of the Euclidean Space denoted
as $\set Ts{}{}{}\subseteq\text{\ensuremath{\mathbb{R}}}^{3}$ and
a special orthogonal group subset given by $\set Rs{}{}{}\subseteq SO(3)$.
Hence, the topological space of a viewpoint constraint is given as
follows:
\begin{align}
\CC i{}{}= & \set Ts{}{}{}\times\set Rs{}{}{}\nonumber \\
= & \{\SP{}\in\CC i{}{},\SP{}\in SE(3)\}\label{eq:C-C-i}\\
= & \{\SP{}(\VECJ ts{}{}{},\text{\ensuremath{\VECJ rs{}{}{}}})\in\CC i{}{}\nonumber \\
 & \mid\VECJ ts{}{}{}\in\set Ts{}{}{},\text{\ensuremath{\VECJ rs{}{}{}}}\in\set Rs{}{}{}\}.\nonumber 
\end{align}
\item If there exists at least one sensor pose in the $i$ $\ACC$ $\exists\SP{}\in\CC i{}{}$,
then this sensor pose fulfills the viewpoint constraint $\c i$ to
acquire feature $f$; hence, a valid viewpoint exists $(f,\SP{},c_{i})\in\CV{}{}$.
\item If there exists a topological space, $\CC i{}{}$, for each constraint
$\forall\c i\in\widetilde{C}$ then the intersection of all individual
constrained spaces constitutes the joint $\ACC$ $\CC{}{}{}\subseteq SE(3)$:
\begin{equation}
\CC{}{}{}(\widetilde{C})=\stackrel[\begin{aligned}\c i\in\widetilde{C}\end{aligned}
]{}{\bigcap}\CC i{}{}(c_{i}).\label{eq:C-C}
\end{equation}
\item If the joint constrained space is a non-empty set, i.e., $\CC{}{}{}\neq\emptyset$,
then there exists at least one sensor pose $\exists\SP{}\in\CC{}{}{}$
and consequently a viewpoint $(f,\SP{},\widetilde{C})\in\CV{}{}$
that fulfills all viewpoint constraints.
\end{enumerate}
An abstract representation of the $\ACC$s and the resulting topological
space $\CC{}{}{}$ intersected by various viewpoint constraints is
depicted in Figure \ref{fig:Constrained-Space}. It is worth mentioning
that although the framework considers an independent formulation of
each viewpoint constraint, the real challenge consists on characterizing
each constraint individually to maintain a high generalization and
flexibility of the framework (cf. Table \ref{tab:General-Requirements}). 

\section{Technical Setup}

This section provides an overview of the hardware and software used
for the characterization of the $\ACC$s within the following sections.
We briefly introduce the considered domains, parameters, and specifications
that were employed to verify the individual formulations of $\ACC$s
presented in Sections \ref{sec:Volumetric-Viewpoint-Space} and \ref{sec:Constraining-CV}. 

\subsection{Domain Models}

\begin{table*}[tbh]
\caption{Imaging parameters of the sensors $s_{1}$ and $s_{2}$\label{tab:Image Paremeters s1 and s2}}

\centering{}\begin{center}
\resizebox{0.98\textwidth}{!}{
\begin{tabular}{>{\raggedright}p{0.22\textwidth}|>{\raggedright}p{0.23\textwidth}|>{\raggedright}p{0.23\textwidth}|>{\raggedright}p{0.23\textwidth}}
\toprule Range Sensor & \multicolumn{2}{>{\centering}p{0.46\textwidth}|}{1} & \multicolumn{1}{>{\centering}p{0.25\textwidth}}{2}\tabularnewline
\midrule Manufacturer & \multicolumn{2}{l|}{Carl Zeiss Optotechnik GmbH} & Roboception\tabularnewline
\midrule Model & \multicolumn{2}{l|}{COMET Pro AE} & rc\_visard 65\tabularnewline
\midrule 3D Acquisition Method & \multicolumn{2}{l|}{Digital Fringe Projection} & Stereo Vision\tabularnewline
\midrule Imaging Device $s_{t}$ & Monochrome Camera: $s_{\ensuremath{1}}$ & Blue Light LED-Fringe Projector: $s_{2}$ & Two monochrome cameras: $s_{3},$$s_{4}$\tabularnewline
\midrule Field of view & $\theta_{s}^{x}\!=\!51.5\,{^\circ}$, $\psi_{s}^{y}\!=\!35.5\,{^\circ}$ & $\theta_{s}^{x}\!=\!70.8\,{^\circ}$, $\psi_{s}^{y}\!=\!43.6\,{^\circ}$ & $\theta_{s}^{x}\!=\!62.0\,{^\circ}$, $\psi_{s}^{y}\!=\!48.0\,{^\circ}$\tabularnewline
\midrule Working distances and near, middle, and far planes relative
to imaging devices lens. & {\small{}$\begin{aligned}[t]@400\,\text{mm}: & (396\times266)\,\text{mm}^{2}\\
@600\,\text{mm}: & (588\times392)\,\text{mm}^{2}\\
@800\,\text{mm}: & (780\times520)\,\text{mm}^{2}
\end{aligned}
$} & {\small{}$\begin{aligned}[t]@ & 200\,\text{mm}:(284\times160)\,\text{mm}^{2}\\
@ & 600\,\text{mm}:(853\times480)\,\text{mm}^{2}\\
@ & 1000\,\text{mm}:(1422\times800)\,\text{mm}^{2}
\end{aligned}
$} & {\small{}$\begin{aligned}[t]@ & 200\,\text{mm}:(118\times178)\,\text{mm}^{2}\\
@ & 600\,\text{mm}:(706\times534)\,\text{mm}^{2}\\
@ & 1000\,\text{mm}:(1178\times890)\,\text{mm}^{2}
\end{aligned}
$}\tabularnewline
\midrule Transformation between sensor lens and TCP $\VECJ T{}{}{TCP}{s_{t}}$ & {\small{}$\VECJ t{}{}{TCP}{s_{1}}:(0,0,-602)\,\text{mm}$}{\small\par}

{\small{}$\VECJ r{}{}{TCP}{s_{1}}:(0,0,0)\,{^\circ}$} & $\VECJ t{}{}{TCP}{s_{2}}:(0,0,-600)\,\text{mm}$

$\VECJ r{}{}{TCP}{s_{2}}:(0,0,0)\,{^\circ}$ & $\VECJ t{}{}{TCP}{s_{3,4}}:(0,0,-600)\,\text{mm}$

$\VECJ r{}{}{TCP}{s_{3,4}}:(0,0,0)\,{^\circ}$\tabularnewline
\midrule Transformation between imaging devices of each sensor $\VECJ T{}{}{s_{1}}{s_{2}}$,$\VECJ T{}{}{s_{3}}{s_{4}}$ & \multicolumn{2}{>{\raggedright}p{0.26\textwidth}|}{$\VECJ t{}{}{s_{1}}{s_{2}}:(217.0,0,8.0)\,\text{mm}$

$\VECJ r{}{}{s_{1}}{s_{2}}:(0,-20.0,0)\,{^\circ}$} & $\VECJ t{}{}{s_{3}}{s_{4}}:(65.0,0,0)\,\text{mm}$

$\VECJ r{}{}{s_{3}}{s_{4}}:(0,0,0)\,{^\circ}$\tabularnewline
\midrule Transformation between both sensors $\VECJ T{}{}{s_{1}}{s_{3}}$ & \multicolumn{3}{l}{$\VECJ t{}{}{s_{1}}{s_{3}}:(348.0,-81.0,42.0)\,\text{mm}$, $\VECJ r{}{}{s_{1}}{s_{3}}:(0.52,0.56,0.34)\,{^\circ}$ }\tabularnewline
 & \multicolumn{1}{>{\raggedright}p{0.23\textwidth}}{} & \multicolumn{1}{>{\raggedright}p{0.23\textwidth}}{} & \tabularnewline\bottomrule
\end{tabular}}
\end{center}
\end{table*}

\begin{itemize}
\item Sensors: We used two different range sensors for the individual verification
of the $\ACC$s and the simulation-based and experimental analyses.
The imaging parameters (cf. Subsection \ref{subsec:Frustum-Space})
and kinematic relations of both sensors are given in Table \ref{tab:Image Paremeters s1 and s2}.
The parameters of the lighting source of the \emph{ZEISS Comet PRO
AE} sensor are conservatively estimated values, which guarantee that
the frustum of the sensor lies completely within the field of view
of the fringe projector. A more comprehensive description of the hardware
is provided in Section \ref{sec:Evaluation}.
\item Object, features, and occlusion bodies: For verification purposes,
we designed an academic object comprising three features and two occlusion
objects with the characteristics given in Table \ref{tab:Verification-Features}
in the Appendix.
\item Robot: We used a \emph{Fanuc M-20ia} six-axis industrial robot and
respective kinematic model to compute the final viewpoints to position
the sensor.
\end{itemize}

\subsection{Software}

The backbone of our framework was developed based on the \emph{Robot
Operating System} (ROS) (Distribution: Noetic Ninjemys) \parencite{Quigley.2009}.
The framework was built upon a knowledge-based service-oriented architecture.
A more detailed overview of the general conceptualization of the architecture
and knowledge-base is provided in our previous works \parencite{Magana.2019,Magana.2020}. 

Most of our algorithms consists on generating and manipulating 3D
manifolds. Hence, based on empirical studies, we evaluated different
open-source Python 3 libraries and used them according to their best
performance for diverse computation tasks. For example, the PyMesh
Library from  \textcite{Zhou.2016} demonstrated the best computational
performance for Boolean operations. On the contrary, we used the trimesh
Library \parencite{DawsonHaggertyetal..2022} for verification purposes
and for performing ray-casting operations, due to its integration
of the performance-oriented Embree Library \parencite{Wald.2014}.
Additionally, for further verification, visualization, and user interaction
purposes, we coupled the ROS kinematic simulation to Unity \parencite{Unity.2021}
using the ROS\# library \parencite{SiemensAG.2021}.

All operations were performed on a portable workstation Lenovo W530
running Ubuntu 20.04 with the following specifications: Processor
Intel Core i7-4810MQ @2.80GHz, GPU Nvidia 3000KM, and 32 GB Ram.

\section{Core $\protect\ACC$ using Frustum Space, Feature Position, and Sensor
Orientation\label{sec:Volumetric-Viewpoint-Space}}

This section outlines the formulation and characterization of the
$\ACC$ spanned by two fundamental viewpoint constraints, the imaging
parameters, characterized by the frustum space $\ACM$, the feature's
position, and the sensor orientation. We systematically analyze how
the sensor frustum space can be used to describe a $\ACC$ in $SE(3)$
and introduce simple and generic formulations for its computation. 

This section begins by introducing an extended formulation of the
$\ACM$ aligned to the concept of configuration spaces. Then, by simplifying
the $\AVGP$ considering just a fixed sensor orientation, we show
how the frustum space and the feature position are used to characterize
the core $\ACC$ of our work. The second subsection shows how $\ACC$s
can be combined to span a topological space in the special Euclidean
using different sensor orientations. 

For the benefit of the comprehension of the concepts introduced within
this section, we consider a feature as just a single surface point
with a normal vector. The manifolds of the computed $\ACC$s and additional
supporting material from this and the following section are found
in the digital appendix of this publication.

\subsection{Frustum Space, Feature Position, and fixed Sensor Orientation \label{subsec:Frustum-Constraint-R3}}

This section shows how the feature position and the frustum space
$\ACM$ can be directly employed to characterize a $\ACC$, which
fulfills the first viewpoint constraint for a fixed sensor orientation. 

\subsubsection{Base Constraint Formulation}

In the first step, we introduce a simple condition for the first constraint,
$\c 1:=\c 1(\VECJ{g_{f}}{,0}{}{}{},\SP{},\CF)$, which considers the
feature (minimally represented by a surface point), and the frustum
space, which is characterized by all imaging parameters and the sensor
pose. Let $\text{\ensuremath{\c 1}}$ be fulfilled for all sensor
poses $\forall\SP{}\in SE(3)$, if and only if the feature surface
point lies within the corresponding frustum space at the regarded
sensor pose: 
\begin{equation}
\c 1\iff\VECJ{g_{f}}{,0}{}{}{}\in\CF(\SP{}).\label{eq:cV1}
\end{equation}

\subsubsection{Problem Simplification considering a fixed Sensor Orientation \label{subsec:Problem-Simplification-One-Orientation}}

Due to the limitations of some sensors regarding their orientation,
it is a common practice within many applications to defined and optimize
the sensor orientation beforehand. Then, the $\AVGP$ can be reduced
to an optimization of the sensor position $\VECJ ts{}{}{}$. Hence,
let condition \eqref{eq:cV1} be reformulated to consider a fixed
sensor orientation ${\VECJ rs{fix}{}{}\in SO(3)}$ and to be true
for all sensor positions ${\forall\VECJ ts{}{}{}\in\mathbb{R}^{3}}$
that fulfill following condition:
\begin{equation}
\c 1\iff\VECJ{g_{f}}{,0}{}{}{}\in\CF(\SP{}(\VECJ ts{}{}{},\VECJ rs{}{}{}=\VECJ rs{fix}{}{})).\label{eq:cV1-2}
\end{equation}

\subsubsection{Constraint Reformulation based on Constrained Spaces}

Recalling the idea to characterize geometrically any viewpoint constraint
(see Subsection \ref{subsec:VSC-Formulation}), we find that the viewpoint
constraint formulation of Equation \eqref{eq:cV1-2} to be unsatisfactory.
We believe that this problem can be solved efficiently using geometric
analysis and assume there exists a topological space denoted by $\CC 1{}{}:=\CC 1{}{}(\c 1)$,
which can be characterized based on the $\ACM$ considering a fixed
sensor's orientation. If such space exists, then all sensor positions
within it fulfill the viewpoint constraint given by Equation \eqref{eq:cV1-2}.

Combining the formulation for $\ACC$s given by Equation \ref{eq:C-C-i}
and the viewpoint constraint condition from Equation \eqref{eq:cV1-2},
the formal definition of the topological space $\CC 1{}{}$ is given:
\begin{equation}
\begin{aligned}\CC 1{}{} & =\{\forall\SP{}(\VECJ ts{}{}{},\VECJ rs{}{}{}=\VECJ rs{fix}{}{})\in\CC 1{}{}\\
 & \mid\forall\VECJ ts{}{}{}\ensuremath{\in}\set Ts{}{}{},\VECJ rs{fix}{}{}\in\set Rs{}{}{},\VECJ{g_{f}}{,0}{}{}{}\in\CF\}.
\end{aligned}
\label{eq:C-C-1}
\end{equation}

\subsubsection{Characterization}

Within the framework of our research, we found out that the manifold
of $\CC 1{}{}$ can be characterized in different ways. This subsection
presents two possible solutions to characterize the $\ACC$ as given
by Equation \ref{eq:C-C-1} using analytic geometry.
\begin{description}
\item [{1.}] \textbf{Extreme Viewpoints Interpretation\label{subsec:Extreme-V-Interpretation}} 
\end{description}
The simplest way to understand and visualize the topological space
$\CC 1{}{}$ is to consider all possible extreme viewpoints to acquire
a feature $f$. These viewpoints can be easily found by positioning
the sensor so that each vertex (corner) of the $\ACM$, $\VECJ Vk{\set Is{}{}{}}{}{}$,
lies at the feature's origin $B_{f}$, which corresponds to the position
of the surface point $\VECJ{g_{f}}{,0}{}{}{}$. The position of such
an extreme viewpoint corresponds to the $k$ vertex $\VECJ Vk{\CC 1{}{}}{}{}\in\fr{}{}V{}{\CC 1{}{}}$
of the manifold $\CC 1{}{}$. Depending on the positioning frame of
the sensor $\fr{}{}Bs{TCP}$ or $B_{s}^{s_{1}}$, the space can be
computed for the TCP($\set C1{}{}{TCP}$) or the sensor lens($\set C1{}{}{s_{1}}$).
The vertices can be straightforwardly computed following the steps
given in Algorithm \eqref{alg:CV1-Extreme-Viewpoint}.

\begin{algorithm}[tbh]
\caption{Extreme Viewpoint Characterization of the Constrained Space $\protect\CC 1{}{}$\label{alg:CV1-Extreme-Viewpoint}}

\begin{enumerate}
\item Consider a constant sensor orientation $\VECJ rs{fix}{}{ref}$ to
acquire a feature $f$.
\item Position the sensor so that the $k$ vertex of the frustum space lies
at the feature's origin, 
\[
\VECJ p{s,k}{}{}{ref}(\VECJ ts{}{}{ref}(\VECJ Vk{\CF}{}{}=B_{f}),\VECJ rs{fix}{}{ref}).
\]
\item Let the coordinates of the $k$ vertex of the constrained space $\set C1{}{}{ref}$
be equal to the translation vector of the sensor frame $\fr{}{}Bs{ref}$
:
\[
\VECJ Vk{\set C1{}{}{ref}}{}{ref}=\VECJ ts{}{}{ref}(\fr{}{}Bs{ref}).
\]
\item Repeat Steps 2 and 3 for all ${l}$ vertices of the frustum space.
\item Connect all vertices from $\fr{}{}V{}{\set C1{}{}{ref}}$ analogously
to the vertices of the frustum space $\fr{}{}V{}{\CF}$ to obtain
the $\set C1{}{}{ref}$ manifold.
\end{enumerate}
\end{algorithm}

\begin{figure*}[t]
\centering{}%
\begin{minipage}[t]{0.46\textwidth}%
\begin{center}
\subfloat[Extreme Viewpoint Formulation: Characterization of $\protect\ACC$
$\protect\CC 1{}{}$ by positioning all vertices of $\protect\CF$
at the feature's frame $B_{f}$ \label{fig:CV1-Extreme}]{\centering{}\includegraphics{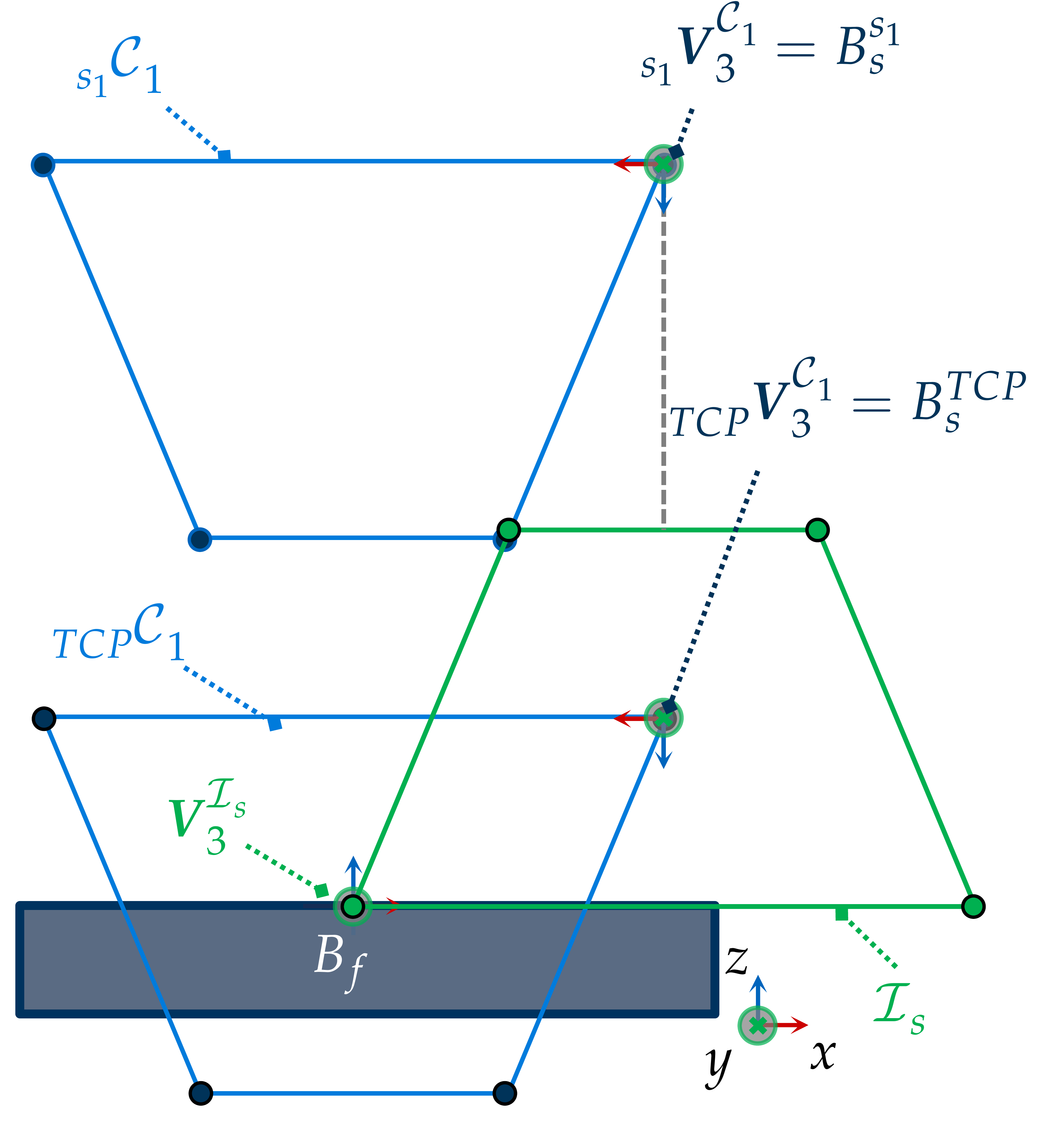}}
\par\end{center}%
\end{minipage}\hfill{}%
\begin{minipage}[t]{0.46\textwidth}%
\begin{center}
\subfloat[Homeomorphism Formulation: Characterization of $\protect\ACC$ $\protect\CC 1{}{TCP}$
by reflecting all vertices of $\protect\CF$ around the feature's
frame $B_{f}$. \label{fig: CV1-Homeo}]{\centering{}\includegraphics{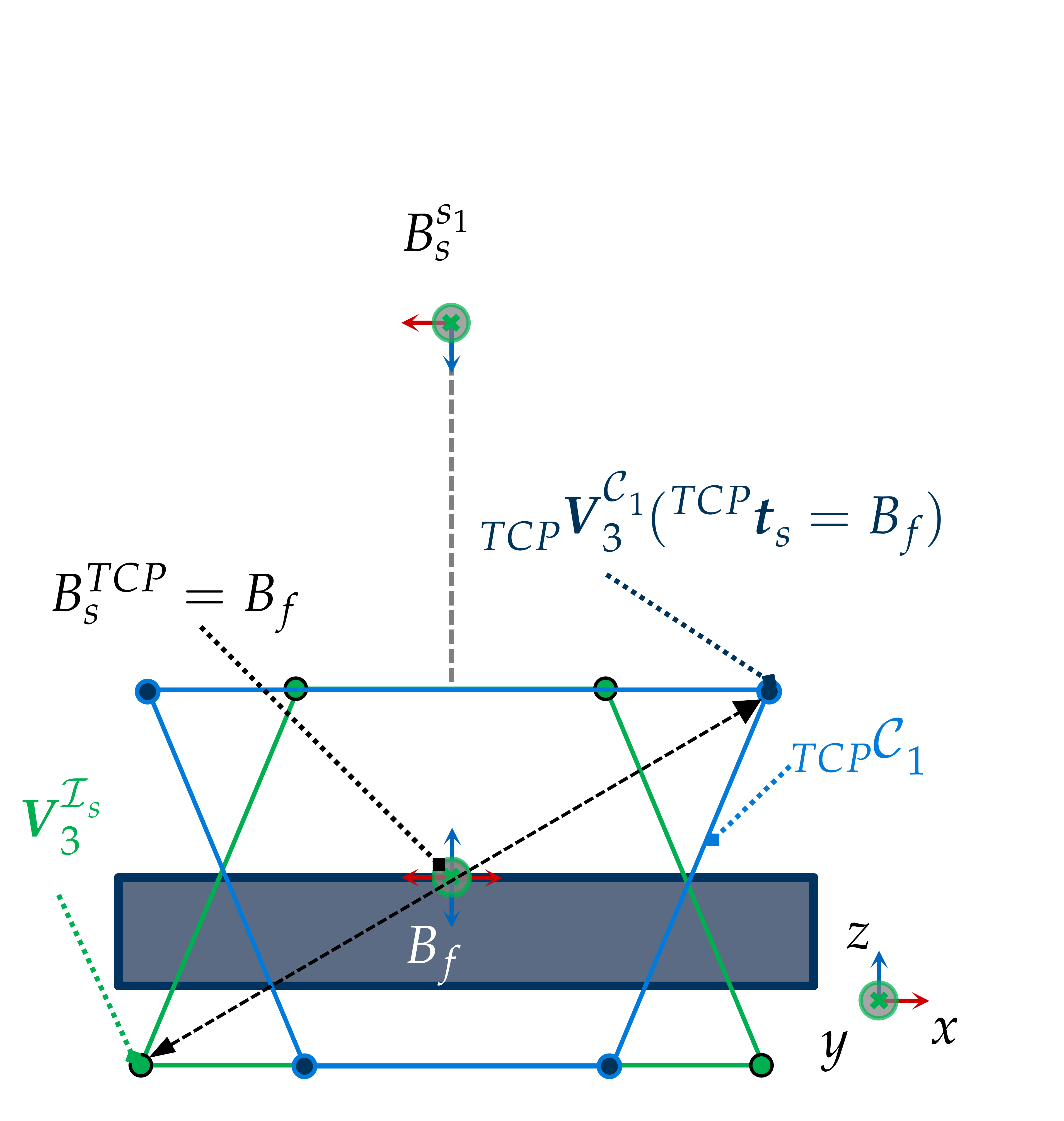}}
\par\end{center}%
\end{minipage}\caption{Geometrical characterization of the $\protect\ACC$ $\protect\CC 1{}{}$
using the frustum space with two different approaches.\label{fig:CV1}}
\end{figure*}

The left Figure \ref{fig:CV1} illustrates the geometric relations
for computing $\CC 1{}{}$ and a simplified representation of the
resulting manifolds in $\mathbb{R}^{2}$ for the sensor TCP $\fr{}{}Bs{TCP}$
and lens $B_{s}^{s_{1}}$.
\begin{description}
\item [{2.}] \textbf{Homeomorphism Formulation\label{subsec:Homeomorphism-Formulation}}
\end{description}
Note that the manifold $\set C1{}{}{ref}$ illustrated in Figure \ref{fig:CV1-Extreme}
has the same topology as the $\ACM$. Thus, it can be assumed there
exists a homeomorphism between both spaces such that $h:\CF\rightarrow\CC 1{}{}$.
Letting the function $h$ correspond to a point reflection over the
geometric center of the frustum space, the vertices of the manifold
$\fr{}{}V{}{\set C1{}{}{ref}}$ can be straightforwardly estimated
following the steps described in the Algorithm \ref{alg:homeomorphism P-Space}.
The resulting manifold for the TCP frame is shown in Figure \ref{fig: CV1-Homeo}.

\begin{algorithm}[tbh]
\caption{Homeomorphism Characterization of the Constrained Space $\protect\CC 1{}{}$
\label{alg:homeomorphism P-Space}}

\begin{enumerate}
\item Consider a constant sensor orientation $\VECJ rs{fix}{}{ref}$ to
acquire a feature $f$.
\item Position the sensor reference frame at the feature's surface point
origin $\VECJ psf{}{ref}(\VECJ ts{}{}{ref}=B_{f})$.
\item For each ${k}$ vertex of the frustum space, compute its reflection
transformation $h$ across the reference pivot frame $B_{s}^{ref}$
\[
\VECJ Vk{\set C1{}{}{ref}}{}{ref}=h(\VECJ Vk{\CF}{}{}(\VECJ ts{}{}{ref}=B_{f}),B_{s}^{ref}).
\]
\item Connect all vertices from $\fr{}{}V{}{\set C1{}{}{ref}}$ analogously
to the vertices of the frustum space $\fr{}{}V{}{\CF}$ to obtain
the $\set C1{}{}{ref}$ manifold.
\end{enumerate}
\end{algorithm}

\begin{itemize}
\item \textbf{General Notes }
\end{itemize}
We considered the steps described in Algorithm \ref{alg:homeomorphism P-Space}
to be the most traceable strategy using a homeomorphism to compute
the constrained space $\set C1{}{}{ref}$. Nevertheless, we do not
refuse any alternative approach for its characterization. For instance,
the same manifold of $\set C1{}{}{ref}$ for any reference frame can
also be obtained by first computing the reflection model of the frustum
space $\set Is*{}{}$ over its geometric center at $\fr{}{}Bs{\set Is{}{}{}}$:
\begin{equation}
h:(\fr{}{}Bs{\set Is{}{}{}},\set Is{}{}{})\rightarrow\set Is*{}{}.\label{eq: CV1-Reflection-Origin}
\end{equation}
The manifold of $\set Is*{}{}$ can then be just simply translated
to the desired reference frame so that $\fr{}{}Bs{\set Is*{}{}}=B_{s}^{ref}$,
considering that the TCP must be positioned at the origin of the feature
$\VECJ psf{}{TCP}(\VECJ ts{}{}{TCP}=B_{f},\VECJ rs{fix}{}{ref})$. 

Moreover, our approach considers that the topological space spanned
by $\CC 1{}{}$ exists if the following conditions hold: 
\begin{itemize}
\item the frames of all vertices of the frustum space $\VECJ V{s,i}{}{}{}(\CF),i=1..j$
are known,
\item the frustum space is a watertight manifold, 
\item and the space between connected vertices of the frustum space is linear;
hence, adjacent vertices are connected only by straight edges. 
\end{itemize}
Throughout this paper, we characterize most of the $\ACC$s considering
just the reference frame for the sensor lens $s_{1}$; hence, if not
stated otherwise consider $\CC 1{}{}:=\set C1{}{}{s_{1}}$. 

\subsubsection{Verification}

Any of the two formulations presented in this subsection can be straightforwardly
extended to characterize the $\ACC$ $\CC 1{}{}$ in $SE(3)$. However,
we found the homeomorphism formulation to be the most practical way
to compute the $\CC 1{}{}$ manifold. Hence, to verify the characterization
of $\CC 1{}{}$ using this approach, we first defined a sensor orientation
in $SE(3)$ denoted as $\VECJ rs{fix}{f_{0}}{}$ to acquire the feature
$f_{0}$. We computed then the $\ACM$ manifold using a total of $j=8$
vertices\endnote{Some sensors might consider a variation of the FOV between the near-middle
planes and the middle-far planes. Hence, in such cases 12 vertices
would be necessary to characterize the frustum space correctly. If
the slope between the near and far planes is constant, eight vertices
are sufficient.} with the imaging parameters of $s_{1}$ from Table \ref{tab:Image Paremeters s1 and s2}
and computed the reflected manifold of the $\ACM$ $\set Is*{}{}$
as proposed by Equation \ref{eq: CV1-Reflection-Origin}. In the next
step, we transformed $\set Is*{}{}$ using the rigid transformation
\[
\VECJ T{}{}f{s_{1}}=\VECJ psf{}{TCP}(\VECJ ts{}{}{TCP}=B_{f_{o}},\VECJ rs{}{}{TCP}=\VECJ rs{fix}{f_{0}}{})
\]
 to obtain the $\ACC$ manifold $\set C1{}{}{TCP}$ for the sensor
TCP frame and the transformation $\VECJ T{}{}f{s_{1}}=\VECJ T{}{}f{TCP}\cdot\VECJ T{}{}{TCP}{s_{1}}$
to characterize the manifold $\set C1{}{}{s_{1}}$ for the sensor
lens frame.

\begin{figure*}[t]
\begin{centering}
\begin{minipage}[t]{0.48\textwidth}%
\begin{center}
\subfloat[$\protect\ACC$s manifolds $\protect\set C1{}{}{s_{1}}(\protect\VECJ rs{fix}{f_{0}}{})$
(blue filled) and $\protect\set C1{}{}{TCP}(\protect\VECJ rs{fix}{f_{0}}{})$
(transparent with blue edges) in $SE(3)$ considering a fixed orientation
$\protect\VECJ rs{fix}{f_{0}}{}(\alpha_{s}^{z}=170{^\circ},\beta_{s}^{y}=5{^\circ},\gamma_{s}^{x}=45{^\circ})$.
\label{fig:CV1-Veri}]{\centering{}\includegraphics{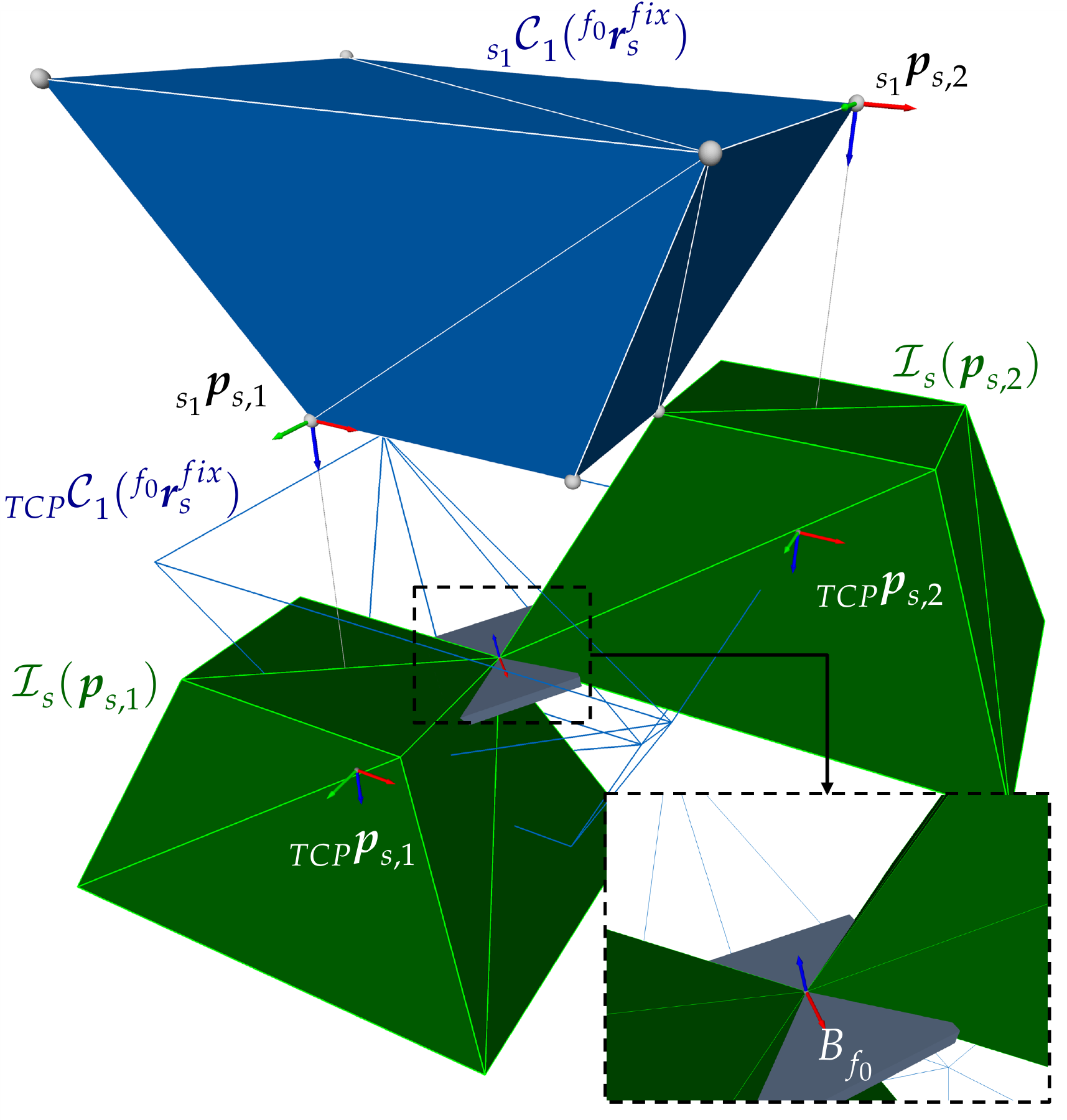}}
\par\end{center}%
\end{minipage}\hfill{}%
\begin{minipage}[t]{0.48\textwidth}%
\begin{center}
\subfloat[$\protect\ACC$s manifolds considering two different sensor orientations
$\protect\CC 1{}{}(\protect\VECJ r{s,1}{}{f_{0}}{}(\alpha_{s}^{z}=180{^\circ},\gamma_{s}^{x}=0{^\circ},\beta_{s}^{y}=-25{^\circ}))$
and $\protect\CC 1{}{}(\protect\VECJ r{s,2}{}{f_{0}}{}(\alpha_{s}^{z}=180{^\circ},\gamma_{s}^{x}=0{^\circ},\beta_{s}^{y}=30{^\circ}))$.\label{fig:CV1-3D-Multipe}]{\begin{centering}
\includegraphics{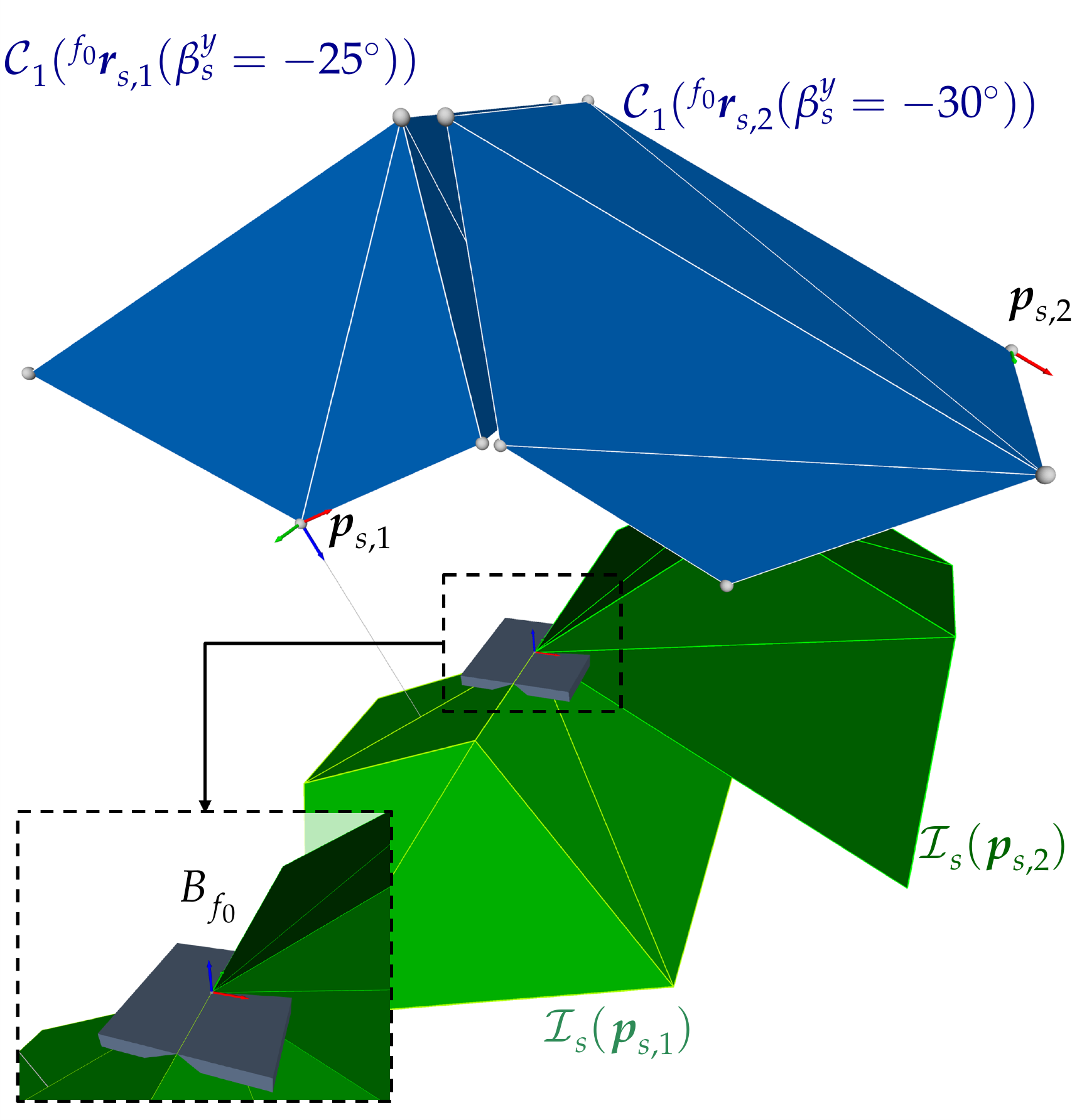}
\par\end{centering}
}
\par\end{center}%
\end{minipage}
\par\end{centering}
\centering{}\caption{Characterization of different $\text{\ensuremath{\protect\ACC}s}$
$\protect\CC 1{}{}(f_{0},\protect\CF,\protect\VECJ rs{}{f_{0}}{})$
(blue manifolds) in $SE(3)$ considering different sensor orientations
using the homeomorphism formulation. The $\protect\ACM$s (green manifolds)
corresponding to different evaluated extreme viewpoints demonstrate
that the feature $f_{0}$ can be captured even from a sensor pose
lying at the vertices of the $\protect\ACC$; hence, any sensor pose
within the $\protect\ACC$ $\protect\SP{}\in\protect\CC 1{}{}$ can
also be considered valid. \label{fig:CV1-3D-Verification}}
\end{figure*}

Figure \ref{fig:CV1-3D-Verification} shows the resulting $\CC 1{}{}$
manifolds considering different sensor orientations. The left Figure
\ref{fig:CV1-Veri} visualizes the $\set C1{}{}{s_{1}}$ and $\set C1{}{}{TCP}$
manifolds considering following orientation in $SE(3)$: $\VECJ rs{fix}{f_{0}}{}(\alpha_{s}^{z}=170{}^{\circ},\beta_{s}^{y}=5{}^{\circ},\gamma_{s}^{x}=45{}^{\circ})$.
On the other hand, Figure \ref{fig:CV1-3D-Multipe} visualizes the
$\ACC$s just for the sensor lens considering two different sensor
orientations. 

To assess the validity of the characterized $\ACC$s, we selected
eight extreme sensor poses lying at the vertices of each $\ACC$ manifold
\[
\begin{aligned}\{\SPn 1(\VECJ ts{}{}{}=\VECJ V1{\CC 1{}{}}{}{},\VECJ rs{}{}{}=\VECJ rs{fix}{f_{0}}{}),\dots,\\
\SPn 8(\VECJ ts{}{}{}=\VECJ V8{\CC 1{}{}}{}{},\VECJ rs{}{}{}=\VECJ rs{fix}{f_{0}}{})\}\in\CC 1{}{}
\end{aligned}
\]
and computed their corresponding frustum spaces ${\CF(\SPn 1),\dots,\CF(\SPn 8)}$.
Our simulations confirmed that the feature $f_{0}$ lay within the
frustum space for all extreme sensor poses, hence satisfying the core
viewpoint condition \ref{eq:cV1-2}. Some exemplary extreme sensor
poses and their corresponding $\ACM$ are shown in the Figures \ref{fig:CV1-3D-Verification}.
The rest of the renders, manifolds of the $\ACC$s, frames, and object
meshes for this example and following examples can be found in the
digital supplement material of this paper. 

As expected, the computational efficiency for characterizing one $\ACC$
showed a good performance with an average computation time (30 repetitions)
of $4.1\,ms$ and a standard deviation of $\sigma=2.4\,ms$. The computation
steps comprise a read operation of the vertices (hard-coded) of the
frustum space as well as the required reflection and transformation
operations of a manifold with eight vertices. 

\subsubsection{Summary}

This subsection outlined the formulation and characterization of the
fundamental $\ACC$ $\CC 1{}{}$, which is characterized based on
the sensor imaging parameters, the feature position, and a fixed sensor
orientation. Using an academic example, we demonstrated that any sensor
pose (fix orientation) within $\CC 1{}{}$ was valid to acquire the
regarded feature satisfying the imaging sensor constraints. Moreover,
two different strategies were proposed to efficiently characterize
such a topological space based on fundamental geometric analysis.

The formulations and characterization strategies introduced in this
subsection are considered the backbone of our framework. The potential
and benefits of the core $\ACC$ $\CC 1{}{}$ are exploited within
the following subsection and Section \ref{sec:Constraining-CV} to
regard the integration of further viewpoint constraints. 

\subsection{Range of Orientations\label{subsec:C-P-SE3}}

In the previous subsection, the formulation of $\ACC$s for a fixed
sensor orientation was introduced. Based on this formulation, this
subsection outlines the formulation of a topological space in the
special Euclidean $SE(3)$, which allows a variation of the sensor
orientation.

\subsubsection{Motivation}

Within the scope of our work, taking into account applications that
comprise an a priori model of the object and its features and the
problem simplification addressed in Subsection \ref{subsec:Problem-Simplification-One-Orientation},
we consider it unreasonable and inefficient to span a configuration
space that comprises all orientations in $\set Rs{}{}{}\subseteq SO(3)$.
This assumption can be confirmed by observing Figure \ref{fig:CV1-3D-Multipe},
which demonstrates that the topological space, which will allow sensor
rotations with an incidence angle of $-25{^\circ}$ and $30{^\circ}$,
does not exist.

For this reason, we consider it more practical to span a configuration
space, which comprises a minimal and maximal sensor orientation range
$\VECJ rs{min}{}{}\leq\VECJ rs{}{}{}\leq\VECJ rs{max}{}{}\in\set Rsf{}{}$
instead of an unlimited space with all possible sensor orientations.
The minimal and maximal orientation values can be defined considering
the sensor limitations given by the second viewpoint constraint.

\subsubsection{Formulation}

First, consider the range of sensor orientations with
\[
\VECJ r{s,m}{}{}{}\in R_{s},\VECJ rs{min}{}{}\leq\VECJ r{s,m}{}{}{}\leq\VECJ rs{max}{}{},m=1,\dots,n,
\]
and let the $\ACC$ for a single orientation as given by Equation
\ref{eq:C-C-1} be extended as follows:
\begin{align}
\begin{aligned}\CC 2{}{}(R_{s})= & \{\SP{}(\VECJ ts{}{}{},\VECJ r{s,m}{}{}{})\in\CC 2{}{}(R_{s})\\
 & \mid\VECJ ts{}{}{}\ensuremath{\in}\set Ts{}{}{},\VECJ{g_{f}}{,0}{}{}{}\in\CF,\\
 & \VECJ r{s,m}{}{}{}\in R_{s},R_{s}\subseteq\set Rsf{}{}\}.
\end{aligned}
\label{eq:C-C-2}
\end{align}

The topological space, which considers a range of sensor orientations,
denoted by $\CC 2{}{}(R_{s})$ can be seamlessly computed by intersecting
the individual configuration spaces for each $m$ orientation:
\begin{equation}
\CC 2{}{}(R_{s})=\stackrel[r_{s,m}\in R_{s},]{n}{\bigcap}\CC 1{}{}(\VECJ r{s,m}{}{}{},\CF,B_{f}).\label{eq:CV2}
\end{equation}

\subsubsection{Characterization}

The $\ACC$ $\CC 2{}{}(R_{s})$ as given by Equation \ref{eq:CV2}
can be seamlessly computed using $\ACSG$ Boolean Intersection operations.
Considering that each intersection operation yields a new manifold
with more vertices and edges, it is well known that the computation
complexity of $\ACSG$ operations increases with the number of vertices
and edges. Thus, to compute $\CC 2{}{}(R_{s})$ in a feasible time,
a discretization of the orientation range must be first considered.

\begin{figure*}[t]
\centering{}\includegraphics{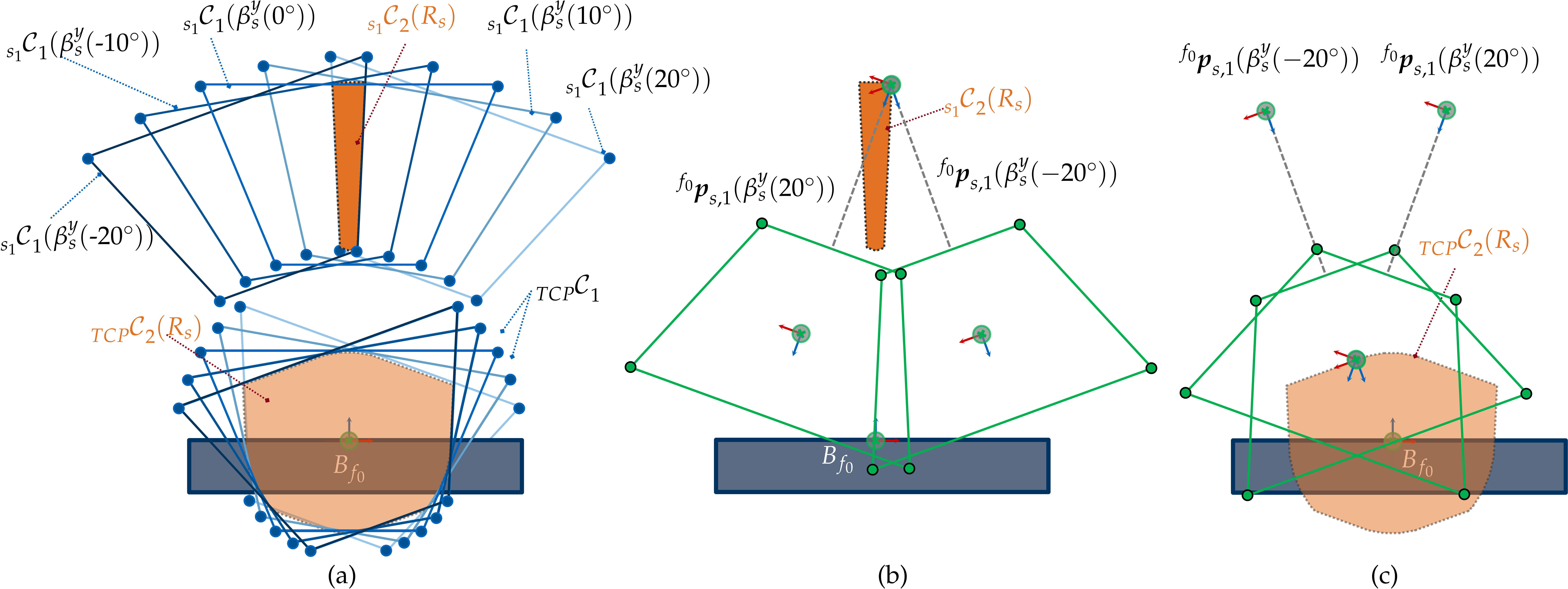}\caption{a) Characterization of $\protect\ACC$s in $\mathbb{R}^{2}$ $\protect\set C2{}{}{s_{1}}(R_{s})$
and $\protect\set C2{}{}{TCP}(R_{s})$ for different positioning frames
with following range of sensor orientations $\{\ensuremath{-20^{{^\circ}},-10^{{^\circ}},0^{{^\circ}},10^{{^\circ}},20^{{^\circ}}\}\in\beta_{s}^{y}}$.
b) Verification of two different viewpoints using sensor lens at positioning
frame $\{\protect\VECJ p{s,1}{}{f_{0}}{s_{1}}(\beta_{s}^{y}=20^{{^\circ}}),\protect\VECJ p{s,2}{}{f_{0}}{s_{1}}(\beta_{s}^{y}=-20^{{^\circ}})\}\in\protect\set C2{}{}{s_{1}}(R_{s})$.
b) Verification of two different viewpoints using sensor TCP as positioning
frame $\{\protect\VECJ p{s,1}{}{f_{0}}{TCP}(\beta_{s}^{y}=20^{{^\circ}}),\protect\VECJ p{s,2}{}{f_{0}}{TCP}(\beta_{s}^{y}=-20^{{^\circ}})\}\in\protect\set C2{}{}{TCP}(R_{s})$
\label{fig:CV2-Char}}
\end{figure*}

One simple and pragmatic solution is to consider different sensor
orientations comprising the maximal and minimal allowed sensor orientations,
e.g., $(\VECJ rs{min}{}{},\VECJ r{s,}{ideal}{}{},\VECJ rs{max}{}{})\in R_{s}$.
Figure \ref{fig:CV2-Char}a illustrates the 2D $\CC 1{}{}$ manifolds
of five different sensor orientations $\{\ensuremath{-20^{{^\circ}},-10^{{^\circ}},0^{{^\circ}},10^{{^\circ}},20^{{^\circ}}\}\in\beta_{s}^{y}}$
using a discretization step of $r_{s}^{d}=10{^\circ}$ for the positioning
frames TCP $\set C2{}{}{TCP}(R_{s})$ and sensor lens frame $\set C2{}{}{s_{1}}(R_{s})$.
The $\CC 2{}{}$ manifolds are characterized by intersecting all individual
spaces $\CC 1{}{}$ as given by Equation \ref{eq:CV2}.

As it can be observed from Fig. \ref{fig:CV2-Char}b and Fig. \ref{fig:CV2-Char}c,
it should be noted that the manifolds $\set C2{}{}{s_{1}}$ and $\set C2{}{}{TCP}$
span different topological spaces (here in $SE(1)$) depending on
the selected positioning frame. Contrary to the $\ACC$s $\set C1{}{}{s_{1}}$
and $\set C1{}{}{TCP}$ considering a fixed sensor position, the selection
of the reference positioning frame should be considered before computing
$\CC 2{}{}$ and taking into account the explicit requirements and
constraints of the vision task. 

\textbf{Discretization without Interpolation} Note that\textbf{ }$\CC 2{}{}(R_{s})$
spans a topological space that is just valid for the sensor orientations
in $R_{s}$ and that the sensor orientation $\VECJ rs{}{}{}$ cannot
be arbitrary chosen within the range $\VECJ rs{min}{}{}<\VECJ rs{}{}{}<\VECJ rs{max}{}{}$.
This characteristic can be more easily understood by comparing the
volume form, $Vol$, of the $\ACC$s $\CC 2{}{}(\VECJ rs{min}{}{},\VECJ rs{max}{}{})$
and $\CC 2{}{}(\VECJ rs{min}{}{},\VECJ r{s,}{ideal}{}{},\VECJ rs{max}{}{})$,
which would show that the $\ACC$ $\CC 2{}{}(\VECJ rs{min}{}{},\VECJ rs{max}{}{})$
is less restrictive:
\begin{equation}
Vol(\CC 2{}{}(\VECJ rs{min}{}{},\VECJ rs{max}{}{}))>Vol(\CC 2{}{}(\VECJ rs{min}{}{},\VECJ r{s,}{ideal}{}{},\VECJ rs{max}{}{})).\label{eq:CV2_inequality}
\end{equation}
This characteristic can particularly be appreciated in the top of
the $\set C2{}{}{s_{1}}(\VECJ rs{min}{}{},\VECJ rs{max}{}{})$ manifold
in Figure \ref{fig:CV2-Char}a. Thus, it should be kept in mind that
the constrained space $\CC 2{}{}(R_{s})$ does not allow an explicit
interpolation within the orientations of $R_{s}$. 

\textbf{Approximation of $\CC 2{}{}$ }However, as it can be observed
from Figure \ref{fig:CV2-Char} the topological space spanned considering
a step size of $10{^\circ}$, $\CC 2{}{}(R_{s}(r_{s}^{d}=10{^{\circ}}))$,
is almost identical to the space if we would relax the step size to
$20{^\circ}$, $\CC 2{}{}(R_{s}(r_{s}^{d}=20{^{\circ}}))$. Hence,
it can be assumed for this case that the $\ACC$s are almost identical
and the following condition will hold:
\begin{equation}
\CC 2{}{}(R_{s}(r_{s}^{d}=10{^{\circ}}))\approx\CC 2{}{}(R_{s}(r_{s}^{d}=20{^{\circ}})).\label{eq:Cv2-Approx}
\end{equation}
\begin{figure}[t]
\centering{}\includegraphics{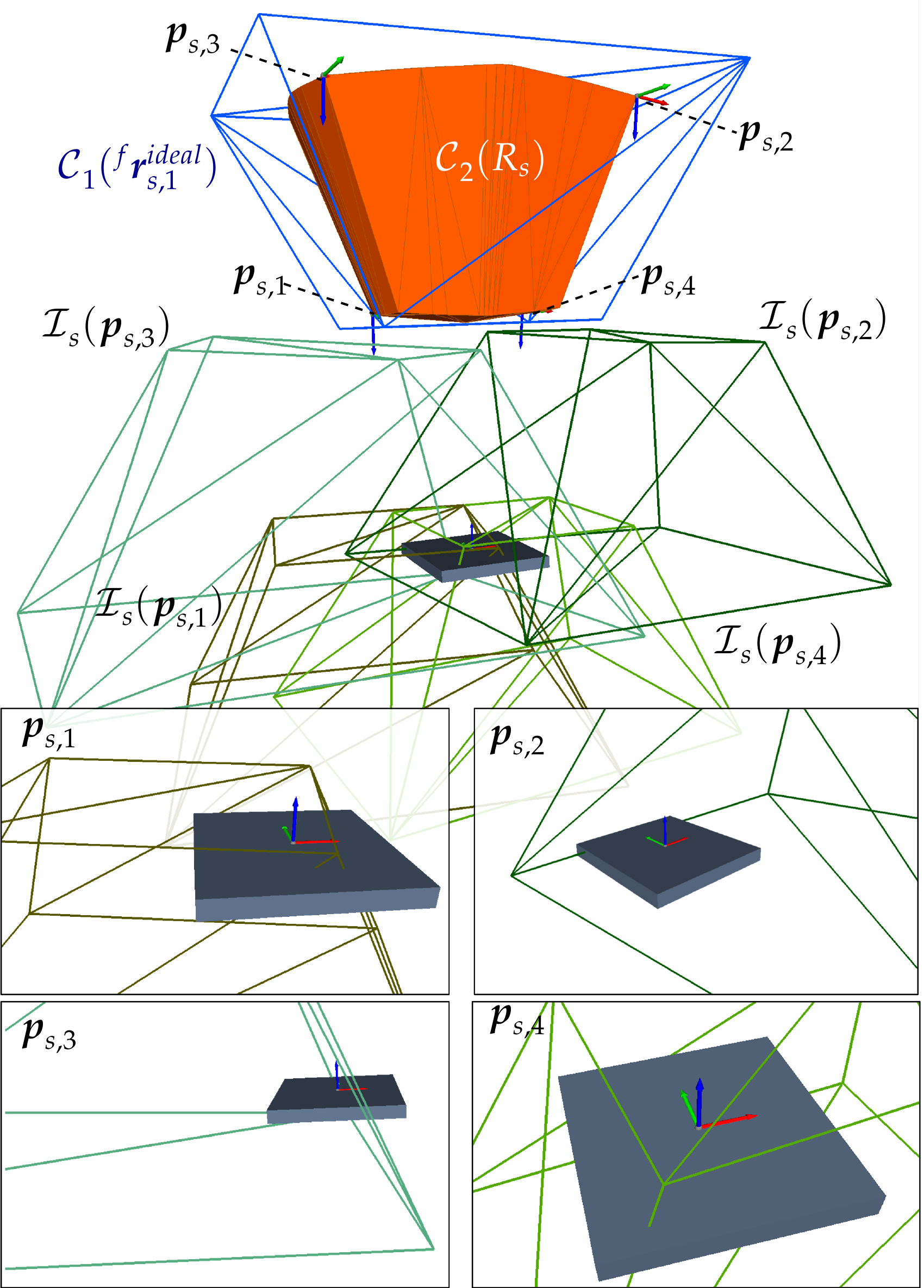}\caption{Characterization of the the $\text{\ensuremath{\protect\ACC}}$, $\protect\CC 2{}{}(R_{s})$,
in $SE(3)$ comprising multiple sensor orientations $R_{s}.$ \label{fig:CV2-Veri}}
\end{figure}

\subsubsection{Verification}

For verification purposes, we consider an academic example with the
following sensor orientation ranges: $\gamma_{s}^{x}=\{-5,0,5\}{}^{\circ}$,
$\beta_{s}^{y}=\{-5,0,5\}{}^{\circ}$, and $\alpha_{s}^{z}=\{-10,0,10\}{}^{\circ}$.
The resulting $\ACC$, $\CC 2{}{}(R_{s})$, were computed by intersecting
the constrained space $\CC 1{}{}(f_{0},\CF,\VECJ rs{}f{})$ for each
possible sensor orientation combination, i.e., $3^{3}=27$ sensor
orientations (cf. Subsection \ref{subsec:Frustum-Constraint-R3}).
The computation time correspond to $t(\CC 2{}{}(R_{s}))=15\,s$. Figure
\ref{fig:CV2-Veri} visualizes the 6D manifold of the $\ACC$ obtained
through Boolean intersection operations. Additionally, the ideal constrained
space considering a null rotation is also displayed to show a qualitative
comparison of the reduction of the $\ACC$ considering a rotation
space in $SE(3)$. 

For verifying the validity of the computed manifold, four extreme
viewpoints and their corresponding frustum spaces are displayed in
Figure \ref{fig:CV2-Veri} considering the following random orientations:
\[
\begin{aligned}\VECJ r{s,1}{}f{} & (\gamma_{s}^{x}=5{^\circ},\beta_{s}^{y}=0{^\circ},\alpha_{s}^{z}=10{^\circ}),\\
\VECJ r{s,2}{}f{} & (\gamma_{s}^{x}=0{^\circ},\beta_{s}^{y}=-5{^\circ},\alpha_{s}^{z}=-10{^\circ}),\\
\VECJ r{s,3}{}f{} & (\gamma_{s}^{x}=-4{^\circ},\beta_{s}^{y}=-5{^\circ},\alpha_{s}^{z}=-8{^\circ}),\\
\VECJ r{s,4}{}f{} & (\gamma_{s}^{x}=3{^\circ},\beta_{s}^{y}=-3{^\circ},\alpha_{s}^{z}=-9{^\circ}).
\end{aligned}
\]

Note that while the first two viewpoints consider an explicit orientation
within the given orientation range $\{\VECJ r{s,1}{}f{},\VECJ r{s,2}{}f{}\}\in R_{s}$,
the sensor orientation of the third and fourth viewpoints are not
elements of $\{\VECJ r{s,3}{}f{},\VECJ r{s,4}{}f{}\}\notin R_{s}$,
however lie within the interpolation range. The frustum spaces prove
that all viewpoints can capture $f$ satisfactorily. Although the
sensor poses $\SPn 3$ and $\SPn 4$ show to be valid in this case,
this assumption cannot be guaranteed for any other arbitrary orientation.
Nevertheless, this confirms  that the approximation condition as given
by Equation \ref{eq:Cv2-Approx} holds to some extent. 

To provide a more quantifiable evaluation of this approximation, the
constrained space, $\CC 2{}{}$, considering finer discretization
steps $r_{s}^{d}$ of $2.5{^\circ}$ and $1{^\circ}$ for $\gamma_{s}^{x}$
and $\beta_{s}^{y}$ was computed. The total number of computed manifolds
corresponds to $5\times5\times3=75$ $\ACC$s with $t(\CC 2{}{}(2.5{^{\circ}}))=40\,s$
for $r_{s}^{d}=2.5{^{\circ}}$ and $11\times11\times3=363$ $\ACC$s
with $t(\CC 2{}{}(1{^{\circ}}))=720\,s$ for $r_{s}^{d}=2.5{^{\circ}}$.
The relative volumetric ratio between the computed spaces is given
as follows: $Vol(\CC 2{}{}(1{^\circ}))/Vol(\CC 2{}{}(2.5{^\circ}))=0.9999$
and $Vol(\CC 2{}{}(1{^\circ}))/Vol(\CC 2{}{}(5{^\circ}))=0.9995$.
These experiments show that the differences between the manifold volume
ratios for the selected steps are insignificant and that the approximation
with a step of $r_{s}^{d}=2.5{^{\circ}}$ holds for this case.

\subsubsection{General Notes}

It is important to note that the validity of the approximation introduced
by Equation \ref{eq:Cv2-Approx} must be individually assessed for
each individual application, imaging parameters, and other constraints.
Some preliminary experiments showed that when considering further
viewpoint constraints that depend on the sensor orientation, e.g.,
the feature geometry, see Subsection \ref{subsec:CV-Feature-Geometry},
the differences between the spaces using different discretization
steps may be more considerable. A more comprehensive analysis falls
outside the scope of this study and remains to be further investigated.
We urge the reader to perform some empirical experiments for choosing
an adequate discretization step and a good trade-off between accuracy
and efficiency. 

\subsubsection{Summary}

Contrary to the previously introduced $\ACC$ $\CC 1{}{}$ which is
limited to a fixed sensor orientation, this subsection outlined the
formulation and characterization of the $\ACC$ $\CC 2{}{}$ in $SE(3)$
that satisfies the sensor imaging parameters for different sensor
orientations. We demonstrated that the manifold $\CC 2{}{}$ is straightforwardly
characterized by intersecting multiple $\ACC$s with different sensor
orientations. 

\section{$\protect\ACC$s of Remaining Viewpoint Constraints \label{sec:Constraining-CV}}

Finding a simple constraint formulation might sometimes be considerably
more challenging than solving the overall problem \parencite{Fruhwirth.2011}.
Hence, posing the $\AVGP$ as a Constraint Satisfaction Problem requires
a comprehensive, individual, and compatible formulation of each viewpoint
constraint. While some of the related works have proposed binary coverage
functions to assess the validity of each constraint for each viewpoint,
we opt to exploit the concept of $\text{\ensuremath{\ACC}}$s to solve
this problem geometrically using linear algebra, trigonometry, and
geometric analysis. 

While Section \ref{sec:Volumetric-Viewpoint-Space} introduced the
core constraints based on the frustum space, sensor orientation, and
feature position, this section outlines an individual formulation
and characterization of the remaining viewpoint constraints (see Table
\ref{tab:VCs}). Moreover, Subsection \ref{subsec:Constraints-Integration-Strategy}
presents one possible strategy to integrate all $\ACC$s demonstrating
the advantages of a consistent and modular characterization.

The formulations presented in this section are motivated by the general
requirements (cf. Table \ref{tab:General-Requirements}) that aim
to deliver a high generalization of the models to facilitate their
use with different $\ARVS$s and vision tasks. Hence, whenever possible,
the characterization of some constraints using simple scalar arithmetic
is prioritized over more complex techniques, and simplifications are
introduced for the benefit of pragmatism, efficiency, and generalization
of the approaches considered. 

\subsection{Feature Geometry\label{subsec:CV-Feature-Geometry}}

In many applications, the feature geometry is a fundamental viewpoint
constraint that may considerably limit the space in $SE(3)$ for positioning
the sensor. This subsection shows that the required $\ACC$ affected
by the feature geometry can be efficiently and explicitly characterized
using trigonometric relationships that depend on the feature geometry,
the sensor's FOV angles, and the sensor orientation.

\subsubsection{Formulation}

Taking into account the third viewpoint constraint (see Table \ref{subsec:Assumptions}),
it can be assumed that all feature surface points $G_{f}(L_{f})$
must be acquired simultaneously. The $\ACC$ that fulfills this requirement
can be easily formulated by extending the base constraint of $\CC 1{}{}$
as given by Equation \ref{eq:C-C-1}, considering that all surface
points must lie within the frustum space:
\begin{equation}
\begin{aligned}\CC 3{}{}= & \{\forall\SP{}(\VECJ ts{}{}{},\CR{})\in\CC 3{}{}(\CC 1{}{},G_{f}(L_{f}),\CR{},\CF)\\
 & \mid\forall\VECJ ts{}{}{}\ensuremath{\in}\set Ts{}{}{},\CR{}\in\set Rs{}{}{},G_{f}(L_{f})\subseteq\CF\}.
\end{aligned}
\label{eq:CV3}
\end{equation}

\subsubsection{Generic Characterization}

Taking into account the generic formulation of the $\ACC$ $\CC 3{}{}$
from Equation \ref{eq:CV3}, in the simplest case, it can be assumed
that $\CC 3{}{}$ could be obtained by scaling $\CC 1{}{}$. Let the
required scaling vector be denoted by $\VECJ{\Delta}{}{}{}{}(\CR{},L_{f},\theta_{s}^{x},\psi_{s}^{y})$
and depend on the feature geometry, the sensor rotation and the FOV
angles of the sensor. The generic characterization of the $\CC 3{}{}$
manifold can then be expressed as follows 
\[
\CC 3{}{}=\CC 1{}{}(\VECJ{\Delta}{}{}{}{}(\CR{},L_{f},\theta_{s}^{x},\psi_{s}^{y})).
\]

Recalling that the $\CC 1{}{}$ manifold is not symmetrical in all
planes, hence, rotation variant, assume that the $\CC 3{}{}$ manifold
cannot be correctly scaled regarding the same scaling vector. Thus,
a more generic and flexible approach can then be considered by letting
each $k$ vertex of $\forall\VECJ V{v,k}{\CC 1{}{}}{}{}\in\fr{}{}Vv{\CC 1{}{}}$
be individually scaled. Hence, each vertex of $\CC 3{}{}$ can be
computed with
\begin{equation}
\VECJ Vk{\CC 3{}{}}{}{}=\VECJ Vk{\CC 1{}{}}{}{}-\VECJ{\Delta}k{}{}{}(\CR{},L_{f},\theta_{s}^{x},\psi_{s}^{y}),\label{eq:V-CV3}
\end{equation}
considering the following generalized vector:
\begin{equation}
\VECJ{\Delta}k{}{}{}(\CR{},L_{f},\theta_{s}^{x},\psi_{s}^{y})=\begin{pmatrix}\Delta_{k}^{x}(\CR{},L_{f},\theta_{s}^{x},\psi_{s}^{y})\\
\Delta_{k}^{y}(\CR{},L_{f},\theta_{s}^{x},\psi_{s}^{y})\\
\Delta_{k}^{z}(\CR{},L_{f},\theta_{s}^{x},\psi_{s}^{y})
\end{pmatrix}.\label{eq:deltaVertex}
\end{equation}

The explicit characterization of the scaling vector from Equation
\ref{eq:deltaVertex} requires an individual and comprehensive trigonometric
analysis of each $k$ vertex of $\CC 3{}{}$, which depends on the
chosen sensor orientation. Moreover, since the scaling vector $\VECJ{\Delta}k{}{}{}$
depends also on the feature geometrical properties, from now on we
will assume the generalization of the feature geometry as introduced
in Subsection \ref{subsec:Simplification-Feature} to characterize
any feature by a square of the length $l_{f}$. This simplification
contributes to a higher generalization of our models for different
topologies and facilitates the comprehension of the trigonometric
relationships introduced in the following subsections.

\subsubsection{Characterization of the $\protect\ACC$ with null rotation\label{subsec:CV3-null-rot}}

The most straightforward scenario to quantify the influence of the
feature geometry on the constrained space is first to consider a null
rotation, $\VECJ rs0f{}$, of the sensor relative to the feature,
i.e., the feature's plane is parallel to the $xy$-plane of the TCP
and the rotation around the optical axis equals zero, ${\VECJ rs0f{}(\gamma_{s}^{x}=\beta_{s}^{y}=\alpha_{s}^{z}=0)}$. 

\begin{figure}[t]
\centering{}\includegraphics{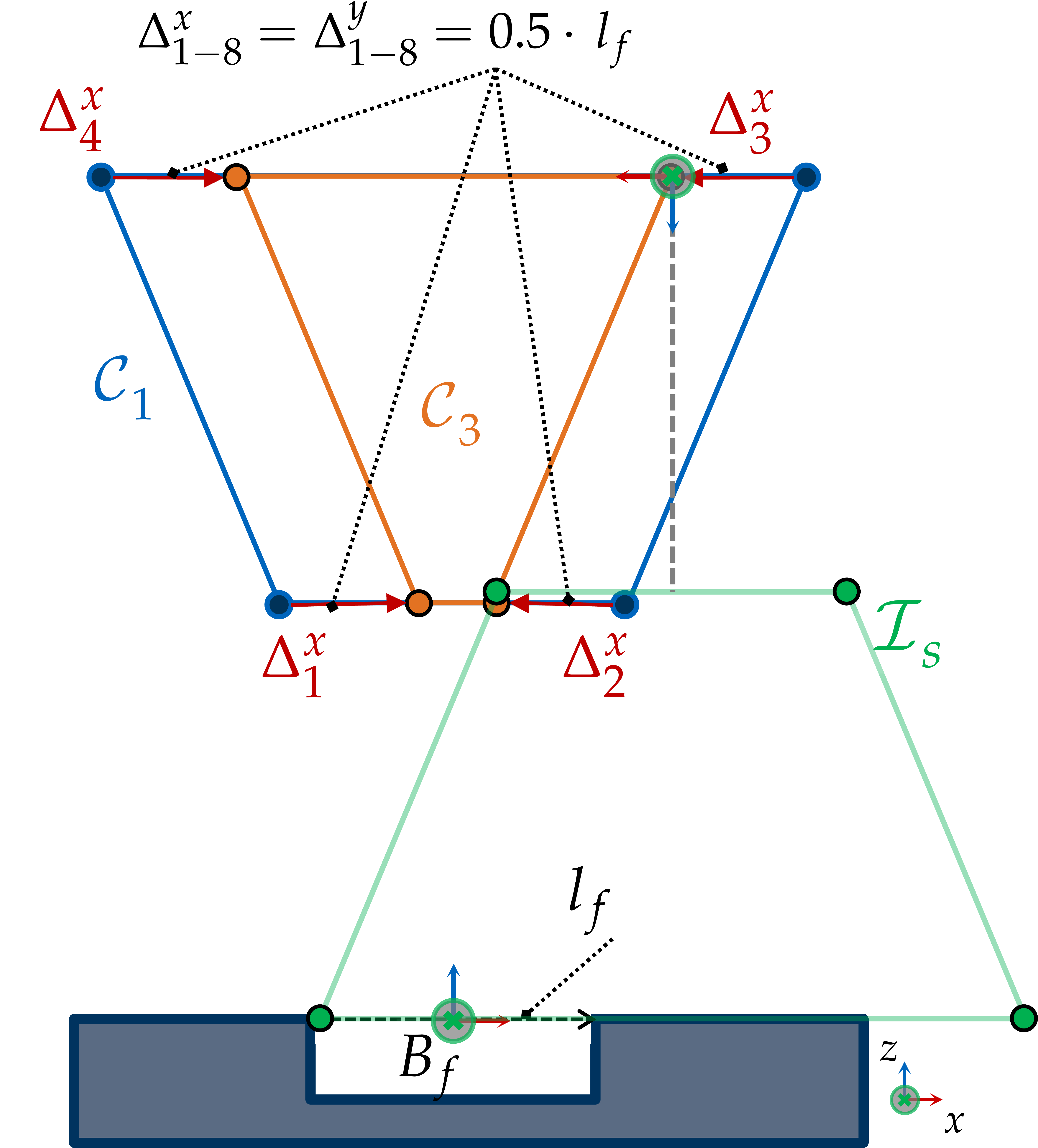}\caption{Characterization of the $\protect\ACC$ $\protect\CC 3{}{}$ considering
a null rotation $\protect\VECJ rs0f{}$ over the feature $f$: scale
all vertices of $\protect\CC 1{}{}$ in the $x$ and $y$ axes considering
the feature geometric length of $0.5\cdot l_{f}$. \label{fig:CV3-R3}}
\end{figure}

First, span the core constrained space, $\CC 1{}{}$, considering
the feature position and the null rotation of the sensor. Then, parting
from one vertex of $\CC 1{}{}$, let the $\ACM$ be translated in
one direction until $\CF$ entirely encloses the whole feature. This
step is exemplary shown in the $x\text{-}z$ plane in Figure \ref{fig:CV3-R3}
at the third vertex of $\CC 1{}{}$ and can be interpreted as an analogy
to the \emph{Extreme Viewpoints Interpretation} (cf. Subsection \ref{subsec:Extreme-V-Interpretation})
used to span $\CC 1{}{}$. Then, it is easily understood that to characterize
$\CC 3{}{}$ all vertices $\forall\VECJ V{v,k}{\CC 1{}{}}{}{}\in\fr{}{}Vv{\CC 1{}{}}$
of $\CC 1{}{}$ must be shifted in the x and y directions by a factor
of $0.5\cdot l_{f}$:
\begin{equation}
\begin{array}{cc}
\Delta_{k}^{x}(\VECJ rs0{}{},l_{f})= & 0.5\cdot l_{f}\\
\Delta_{k}^{y}(\VECJ rs0{}{},l_{f})= & 0.5\cdot l_{f}\\
\Delta_{k}^{z}(\VECJ rs0{}{},l_{f})= & 0.
\end{array}\label{eq:V-CV3-null-rotation-scaling}
\end{equation}

\subsubsection{Rotation around one axis\label{subsec:CV3-SE1}}

Any other sensor orientation different from the null orientation requires
an individual analysis of the exact trigonometric relationships for
each vertex. To break down the complexity of this problem, within
this subsection, we first provide the geometrical relationships needed
to characterize the constrained space $\CC 3{}{}$ considering an
individual rotation around each axis. The characterization of the
constrained spaces follows the same approach described in the previous
subsection, which requires first the characterization of the base
constraint, $\CC 1{}{}$, and then the derivation of the scaling vectors.
\begin{itemize}
\item \textbf{Rotation around $z$-axis} $\alpha_{s}^{z}\neq0$:
\end{itemize}
Assuming a sensor rotation around the optical axis, ${\VECJ rszf{}(\alpha_{s}^{z}\neq0,\varphi_{s}(\beta_{s}^{y},\gamma_{s}^{x})=0)}$
(see Figure \ref{fig:CV3-zAxis}), the $\ACC$ is scaled just along
the vertical and horizontal axes, using the following scaling factors:
\begin{equation}
\begin{array}{ccc}
\Delta_{k}^{x}(\VECJ rsz{}{},l_{f}) & = & \frac{l_{f}}{2}\cdot(\cos(|\alpha_{s}^{z}|)+\sin(|\alpha_{s}^{z}|))\\
\Delta_{k}^{y}(\VECJ rsz{}{},l_{f}) & = & \frac{l_{f}}{2}\cdot(\cos(|\alpha_{s}^{z}|)+\sin(|\alpha_{s}^{z}|))\\
\Delta_{k}^{z}(\VECJ rsz{}{},l_{f}) & = & 0.
\end{array}\label{eq:CV3-zRot}
\end{equation}
The derivation of the trigonometric relationships from Equation \ref{eq:CV3-zRot}
can be better understood by looking at Figure \ref{fig:CV3-SE1-zRot}.
\begin{center}
\begin{figure}[t]
\centering{}\includegraphics{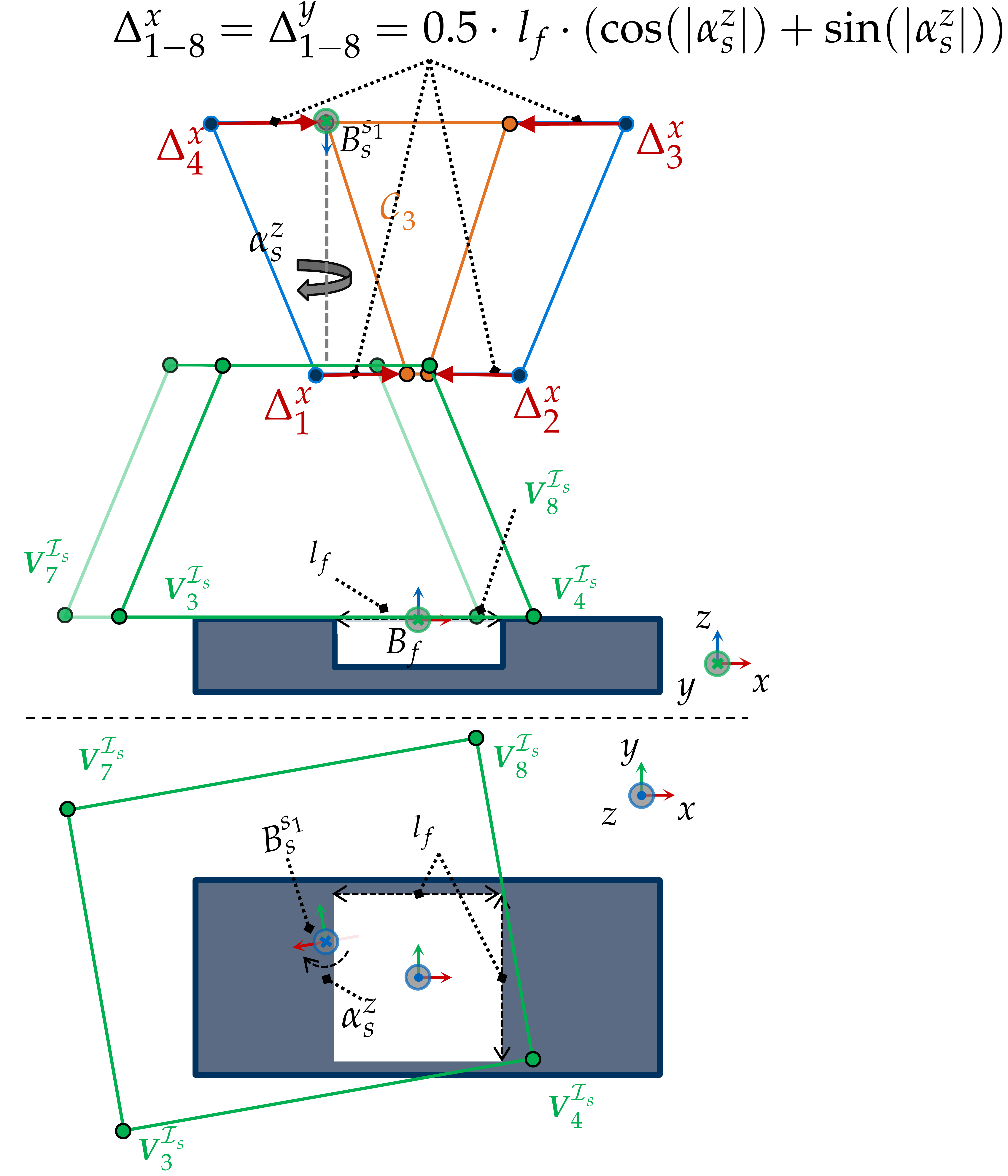}\caption{Characterization of the vertices of the $\protect\ACC$, $\protect\CC 3{}{}$,
considering a rotation around the optical axis $\protect\VECJ rszf{}(\alpha_{s}^{z}\protect\neq0,\varphi_{s}(\beta_{s}^{y},\gamma_{s}^{x})=0)$.
\label{fig:CV3-zAxis}}
\end{figure}
\par\end{center}
\begin{itemize}
\item \textbf{Rotation around $x$-axis or $y$-axis} ($\gamma_{s}^{x}\neq0\text{\ensuremath{\veebar}}\beta_{s}^{y}\neq0$):
\end{itemize}
A rotation of the sensor around the $x$-axis, ${\VECJ rs{}{}{}(\gamma_{s}^{x}\neq0,\alpha_{s}^{z}=\beta_{s}^{y}=0)}$,
or $y$-axis, ${\VECJ rs{}{}{}(\beta_{s}^{y}\neq0,\alpha_{s}^{z}=\gamma_{s}^{x}=0)}$,
requires deriving individual trigonometric relationships for each
vertex of $\CC 3{}{}$. Besides the feature length, other parameters
such as the FOV angles ($\theta_{s}^{x},\psi_{s}^{y}$) and the direction
of the rotation must be considered. 

The scaling factors for the eight vertices of the $\ACC$ considering
a rotation around the $x$-Axis or $y$-Axis can be found in Table
\ref{tab:Vertices-Deltas-xyRot} regarding the following general auxiliary
lengths for $\VECJ rs{}f{}(\alpha_{s}^{z}=\gamma_{s}^{x}=0,\beta_{s}^{y}\neq0)$:
\begin{align*}
\rho^{z,y}= & \frac{l_{f}}{2}\cdot\sin(|\beta_{s}^{y}|)\\
\rho^{x}= & \frac{l_{f}}{2}\cdot\cos(|\beta_{s}^{y}|)\\
\varsigma^{x}= & 2\cdot\rho^{z,y}\cdot\tan(0.5\cdot\theta_{s}^{x})\\
\varsigma^{x,y}= & 2\cdot\rho^{z,y}\cdot\tan(0.5\cdot\psi_{s}^{y})\\
\lambda^{x}= & \rho^{x}-\varsigma^{x}\\
\sigma^{x}= & \rho^{x}+\varsigma^{x},
\end{align*}
and for $\VECJ rs{}f{}(\alpha_{s}^{z}=\beta_{s}^{y}=0,\gamma_{s}^{x}\neq0)$:

\begin{align*}
\rho^{z,x}= & \frac{l_{f}}{2}\cdot\sin(|\gamma_{s}^{x}|)\\
\rho^{y}= & \frac{l_{f}}{2}\cdot\cos(|\gamma_{s}^{x}|)\\
\varsigma^{y}= & 2\cdot\rho^{z,x}\cdot\tan(0.5\cdot\psi_{s}^{y})\\
\varsigma^{y,x}= & 2\cdot\rho^{z,x}\cdot\tan(0.5\cdot\theta_{s}^{x})\\
\lambda^{y}= & \rho^{y}-\varsigma^{y}\\
\sigma^{y}= & \rho^{y}+\varsigma^{y}.
\end{align*}

\begin{figure}[t]
\centering{}\includegraphics{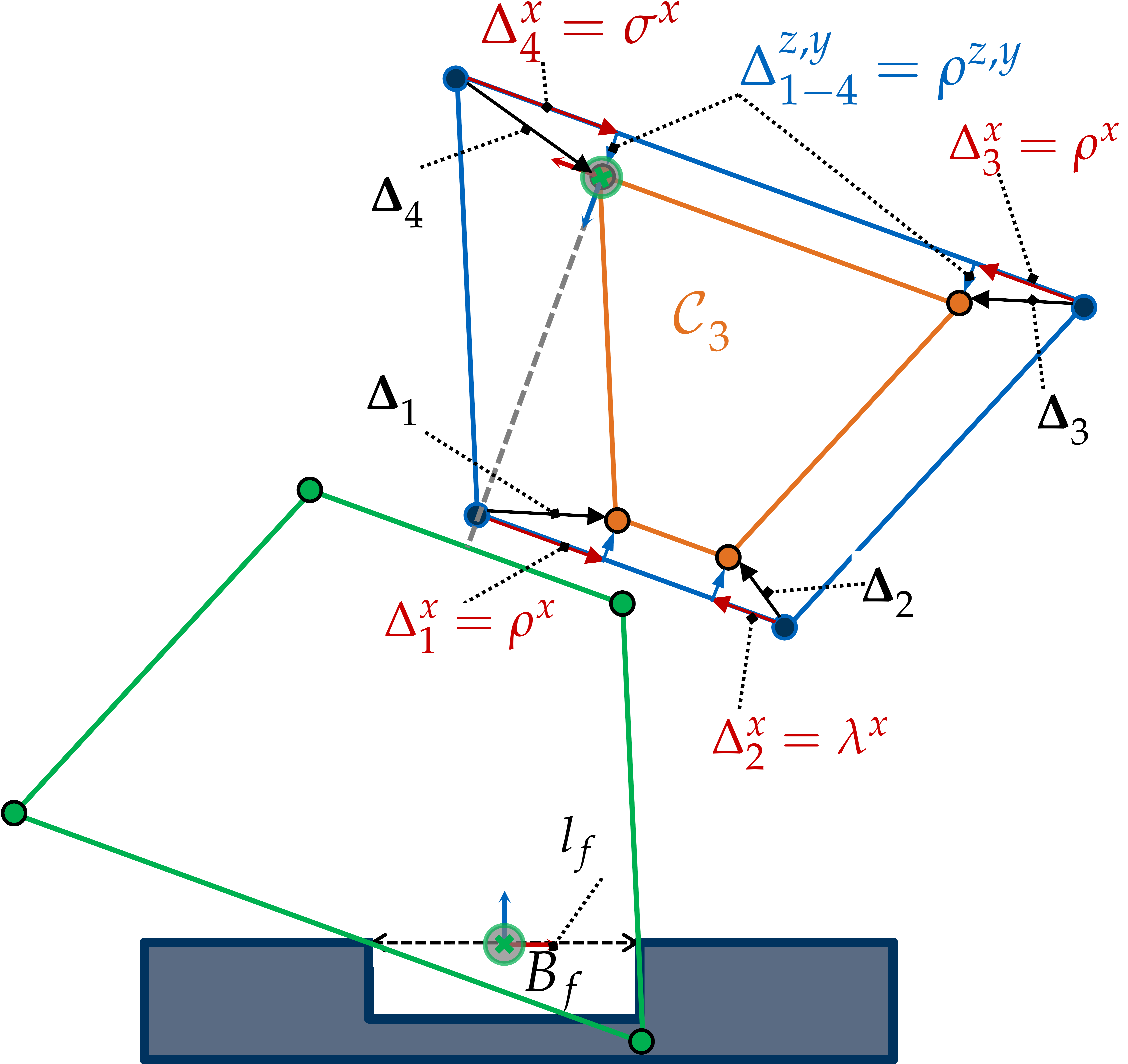}\caption{Characterization of the vertices of the $\protect\ACC$, $\protect\CC 3{}{}$,
considering a rotation around the y-axis $\protect\VECJ rsyf{}(\gamma_{s}^{x}=\alpha_{s}^{z}=0,\beta_{s}^{y}<0)$.\label{fig:-CV3-yAxis}}
\end{figure}

The derivation of the trigonometric relationships can be better understood
using an exemplary case. Thus, first assume a rotation of the sensor
around the y-axis of $\beta_{s}^{y}>0\wedge\gamma_{s}^{x}=0$, as
illustrated in Figure \ref{fig:-CV3-yAxis}. The trigonometric relationships
can then be derived for each vertex following the \emph{Extreme Viewpoints
Interpretation} as exemplary depicted in Figure \ref{fig:CV3-SE1-yRot}).

\subsubsection{Generalization to 3D Features}

Although our approach contemplates primary 2D features, the constrained
space $\CC 3{}{}$ can be seamlessly extended to acquire 3D features
considering a feature height $h_{f}\in L_{f}$.

\begin{figure}[tbh]
\begin{centering}
\includegraphics{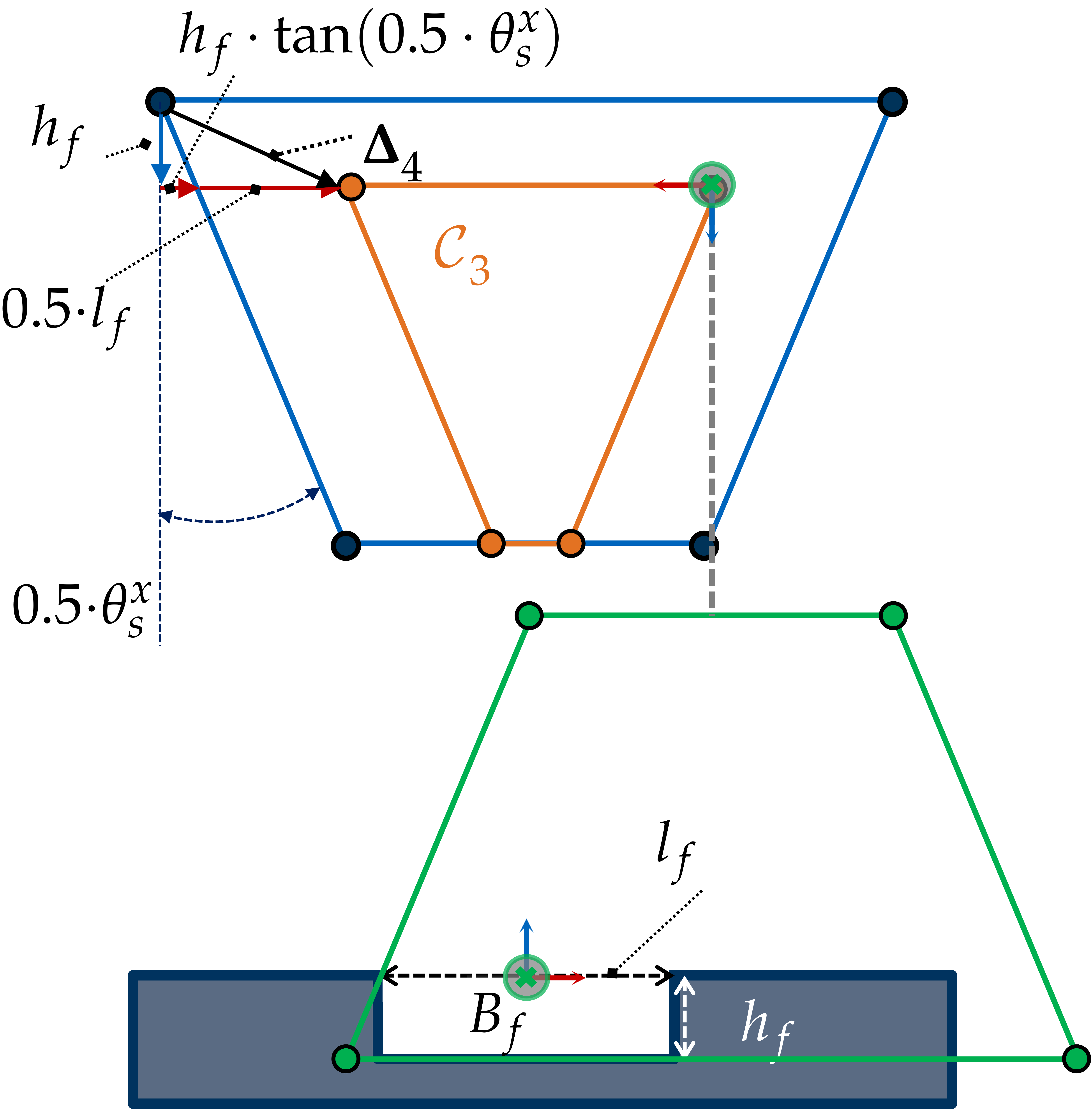}
\par\end{centering}
\caption{Characterization of the vertices of the $\protect\ACC$, $\protect\CC 3{}{}$,
considering a 3D feature and null rotation $\protect\VECJ rs0f{}$.\label{fig:CV3-3D-Feature}}
\end{figure}

This paper just considers the characterization of the scaling vectors
for concave (e.g., pocket, slot) and convex (e.g., cube, half-sphere)
features with a null rotation of the sensor, $\VECJ rs0f{}$. For
instance, the back vertices $(k=1,2,5,6)$ of $\CC 3{}{}$ to capture
a concave feature , as shown in Figure \ref{fig:CV3-3D-Feature},
must be scaled using the following factors:
\begin{align}
\Delta_{k}^{x}= & h_{f}\cdot\tan(0.5{\cdot}\theta_{s}^{x})+0.5{\cdot}l_{f}\nonumber \\
\Delta_{k}^{y}= & h_{f}\cdot\tan(0.5{\cdot}\psi_{s}^{y})+0.5{\cdot}l_{f}\label{eq:V-CV3-3DFeature}\\
\Delta_{k}^{z}= & h_{f}.\nonumber 
\end{align}

The front vertices are scaled using the same factors as for a 2D feature
as given by Equation \ref{eq:V-CV3-null-rotation-scaling}. For convex
features, let all vertices be scaled with the factors given by Equation
\ref{eq:V-CV3-3DFeature}, except for the depth delta factor of the
back vertices, which follows $\Delta_{k}^{z}=0$.

Note that the characterization of the $\ACC$, $\CC 3{}{}$, for considering
3D features just guarantees that the entire feature surface lies within
the frustum space. We neglect any further visibility constraints that
may influence the viewpoint's validity, such as the maximal angles
for the interiors of a concave feature. Moreover, it should be noted
that the scaling factors given within this subsection hold for just
a null rotation. The characterization of the scaling factors for other
sensors orientations can also be derived by extending the previously
introduced relationships for 2D features in Subsection \ref{subsec:CV3-SE1}
.

\subsubsection{Verification}

The verification of the geometrical relationships introduced within
this subsection was performed based on an academic example to acquire
a square feature $f_{1}$ and a 3D pocket feature $f_{1}^{*}$. The
$\ACC$s for $f_{1}$, i.e., $\CC{3,1}{}{}(\VECJ r{s,1}{}{f_{1}}{})$
and $\CC{3,2}{}{}(\VECJ r{s,2}{}{f_{1}}{})$, consider a sensor orientation
of $\VECJ r{s,1}{}{f_{1}}{}(\alpha_{s}^{z}=\beta_{s}^{y}=0,\gamma_{s}^{x}=30{^{\circ}})$
and $\VECJ r{s,2}{}{f_{1}}{}(\gamma_{s}^{x}=\beta_{s}^{y}=0,\alpha_{s}^{z}=15{^{\circ}})$,
while the $\ACC$ $\CC{3,3}{}{}(\VECJ r{s,3}{}{f_{1}^{*}}{})$ for
$f_{1}^{*}$ considers a null sensor orientation $\VECJ r{s,3}{}{f_{1}^{*}}{}(\alpha_{s}^{z}=\beta_{s}^{y}=,\gamma_{s}^{x}=0{^{\circ}})$.
All constrained spaces were computed using the imaging parameters
of sensor $s_{1}$ (cf. Table \ref{tab:Image Paremeters s1 and s2})
and geometric parameters of the features from Table \ref{tab:Verification-Features}.

\begin{figure}[t]
\begin{centering}
\includegraphics{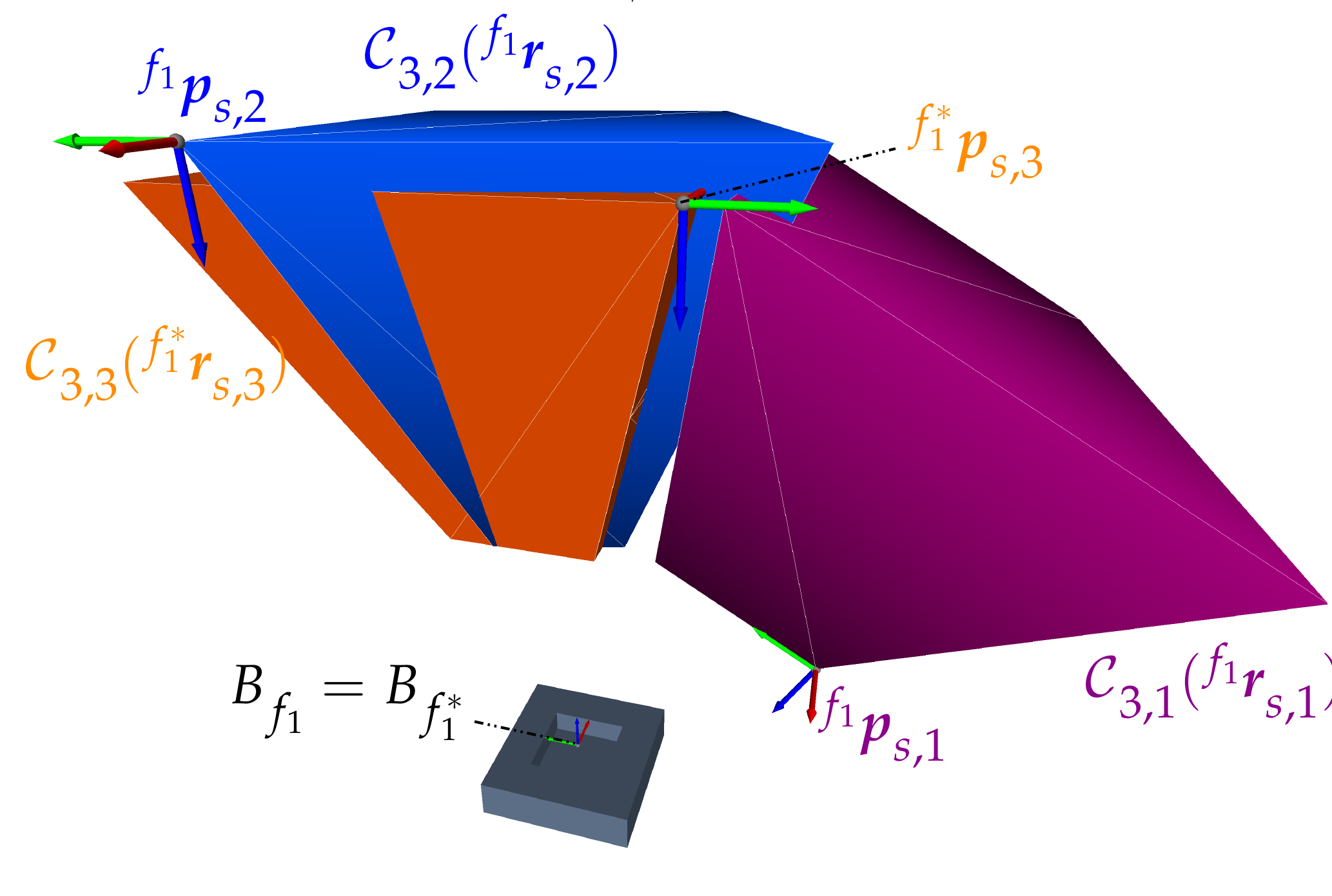}
\par\end{centering}
\caption{Characterization of diverse $\protect\ACC$s in $SE(3)$ considering
the feature geometry to capture a 2D square feature $f_{1}$ and a
3D pocket feature $f_{1}^{*}$. The exemplary scene displays two $\protect\ACC$s
for acquiring feature $f_{1}$ with two different sensor orientations,
$\protect\CC{3_{1}}{}{}(\protect\VECJ r{s,1}{}{f_{1}}{})$ and $\protect\CC{3_{2}}{}{}(\protect\VECJ r{s,2}{}{f_{1}}{})$,
one $\protect\ACC$ $\protect\CC{3_{3}}{}{}(\protect\VECJ r{s,3}{}{f_{1}^{*}}{})$
for capturing $f_{1}^{*}$, and the frames of one extreme viewpoint
at each constrained space. \label{fig:CV3-Veri-Scene}}
\end{figure}

Figure \ref{fig:CV3-Veri-Scene} visualizes the scene comprising the
$\text{\ensuremath{\ACC}s}$ for acquiring $f_{1}$ and $f_{1}^{*}$.
All $\CC{3,1-3}{}{}$ manifolds were computed by scaling first the
manifold of the frustum space considering the scaling factors addressed
within the past subsections and then by reflecting and transforming
the manifold with the corresponding sensor orientation (for $\CC{3,1}{}{}$
see Table \ref{tab:Vertices-Deltas-xyRot}, for $\CC{3,2}{}{}$ see
Equations \ref{eq:CV3-zRot}, and for $\CC{3,3}{}{}$ see Equations
\ref{eq:V-CV3-3DFeature}).

\begin{figure*}[tbh]
\begin{centering}
\begin{minipage}[t]{0.32\textwidth}%
\begin{center}
\subfloat[Rendered scene at extreme viewpoint:  $\protect\VECJ p{s,1}{}{f_{1}}{}\in\text{\ensuremath{\protect\CC{3,1}{}{}(\protect\VECJ r{s,1}{}{f_{1}}{})}}$
\label{fig:CV3-v1} ]{\centering{}\includegraphics{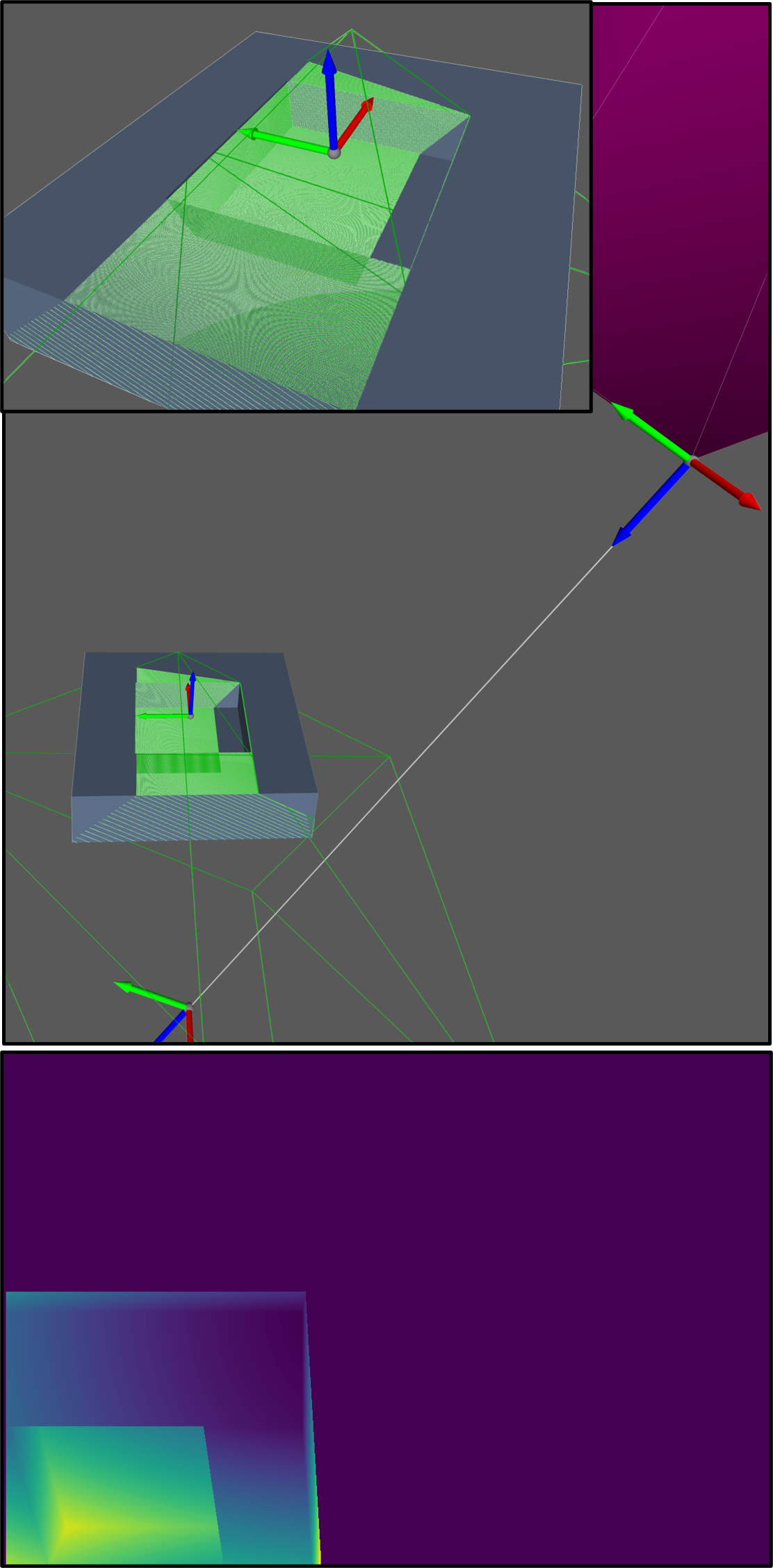}}
\par\end{center}%
\end{minipage}~%
\begin{minipage}[t]{0.32\textwidth}%
\begin{center}
\subfloat[Rendered scene at extreme viewpoint: $\protect\VECJ p{s,2}{}{f_{1}}{}\in\protect\CC{3,2}{}{}(\protect\VECJ r{s,2}{}{f_{1}}{})$
\label{fig:CV3-v2}]{\centering{}\includegraphics{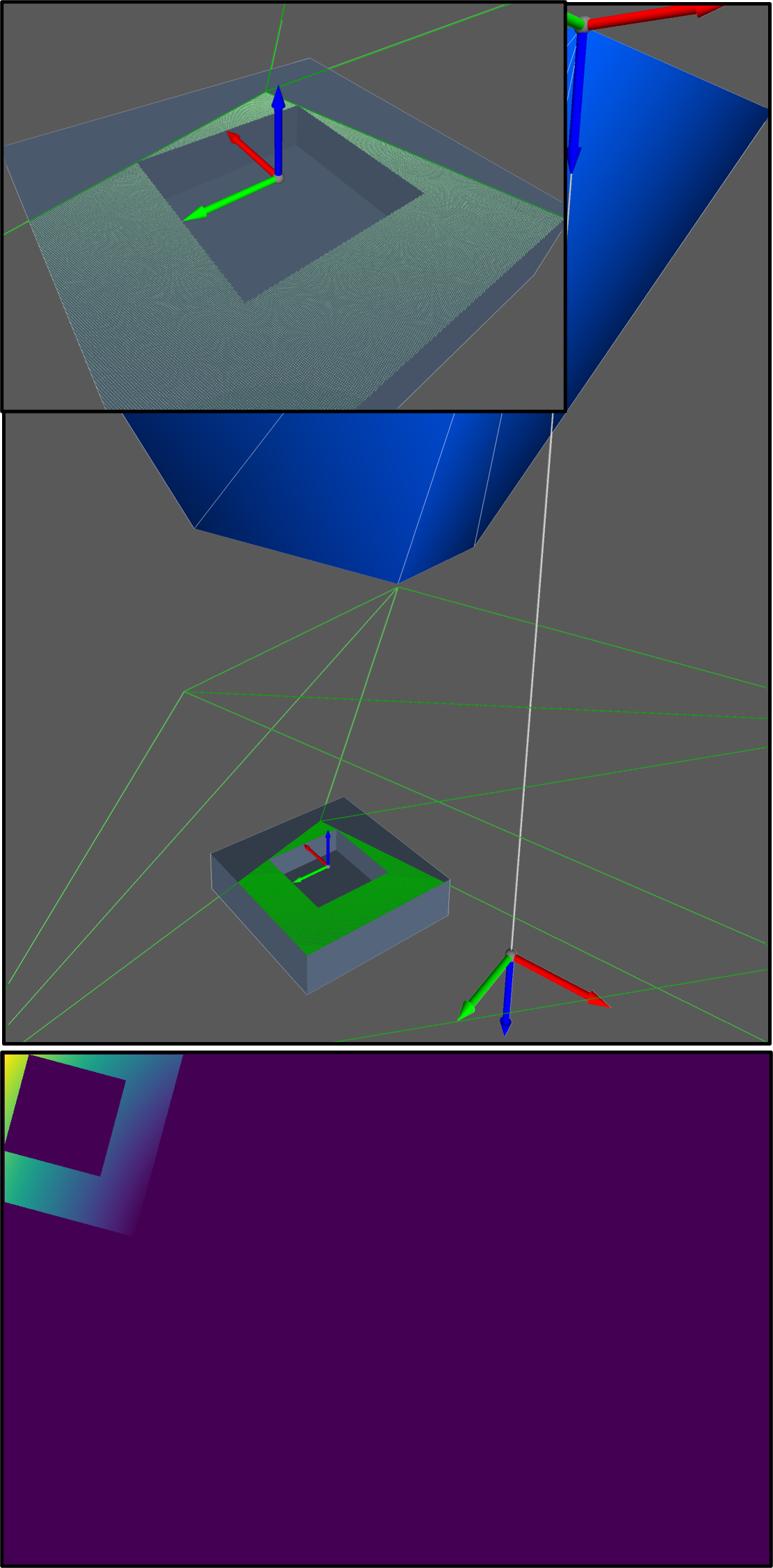}}
\par\end{center}%
\end{minipage}~%
\begin{minipage}[t]{0.32\textwidth}%
\begin{center}
\subfloat[Rendered scene at extreme viewpoint: $\protect\VECJ p{s,3}{}{f_{1}^{*}}{}\in\protect\CC{3,3}{}{}(\protect\VECJ r{s,3}0{f_{1}^{*}}{})$
\label{fig:CV3-v3}]{\centering{}\includegraphics{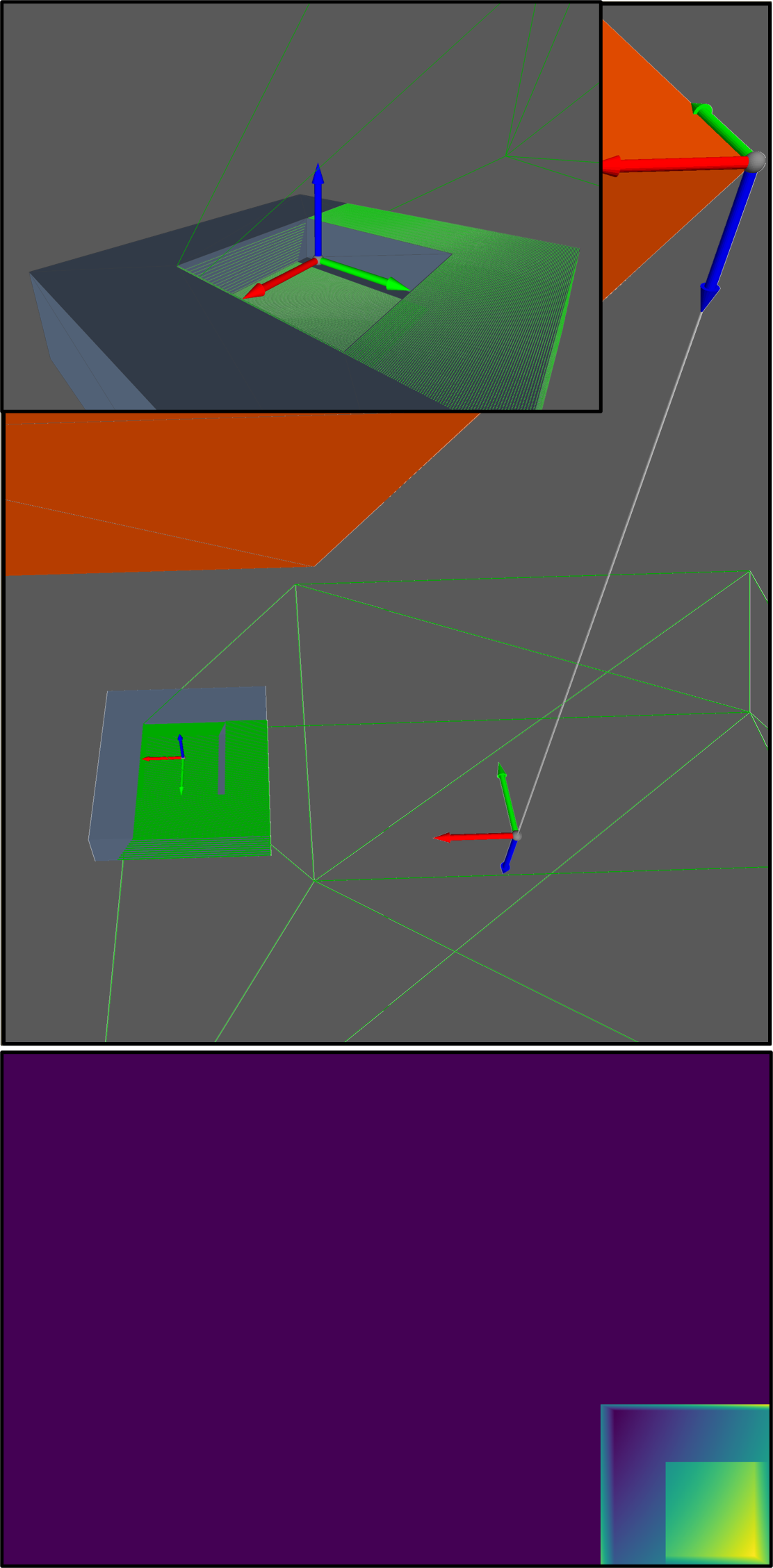}}
\par\end{center}%
\end{minipage}
\par\end{centering}
\caption{Rendered scenes at the extreme viewpoints $\protect\VECJ p{s,1}{}{f_{1}}{}$,
$\protect\VECJ p{s,2}{}{f_{1}}{}$, and $\protect\VECJ p{s,3}{}{f_{1}^{*}}{}$
for verifying that the whole feature geometry lies entirely within
the corresponding frustum spaces. Each figure displays the resulting
frustum space (manifold with green edges), corresponding rendering
point cloud (green points), and depth image (2D image in color map)
at each extreme viewpoint. \label{fig:CV3-vs}}

\end{figure*}

To verify the geometrical relationships introduced within this subsection,
a virtual camera using the trimesh Library \parencite{DawsonHaggertyetal..2022}
and the imaging parameters of $s_{1}$ was created. Then, the depth
images and their corresponding point clouds at eight extreme viewpoints,
i.e., the manifold vertices, were rendered to verify that the features
could be acquired from each viewpoint. The images and point clouds
of all extreme viewpoints confirm that the features lie at the border
of the frustum space and can be entirely captured. Figures \ref{fig:CV3-v1}–\ref{fig:CV3-v3}
demonstrate this empirically and show the depth images and point clouds
at the selected extreme viewpoints ($\VECJ p{s,1}{}{f_{1}}{},\VECJ p{s,2}{}{f_{1}}{},\VECJ p{s,3}{}{f_{1}^{*}}{}$)
from Figure \ref{fig:CV3-Veri-Scene}. 

Our approach provides an analytical and straightforward solution for
efficiently characterizing the $\ACC$ limited by the frustum space
and feature geometry. Since the delta factors can be applied directly
to the vertices of the frustum space, the computational cost is similarly
efficient to the computation time of $\CC 1{}{}$ with an average
computation time of $5.8\,ms$ and $\sigma=3.2\,ms$.


\subsubsection{Summary and Discussion}

This subsection extended the formulation of the core $\ACC$ $\CC 1{}{}$
(see Section \ref{sec:Volumetric-Viewpoint-Space}) to a $\ACC$ $\CC 3{}{}$
taking into account the feature's geometry dimensions. Using an exhaustive
trigonometric analysis, our study introduced the exact relationships
to characterize the vertices of the required multi-dimensional manifold
considering a generalized model of the feature geometry, the sensor's
orientation, and its FOV angles. Our findings show that the characterization
of the $\ACC$ constrained by the feature geometry can be computed
with high efficiency using an analytical approach.

The trigonometric relationships introduced in this section are sufficient
to characterize the $\ACC$ manifold taking into account the rotation
of the sensor in one axis. Characterizing the explicit relationships
for a simultaneous rotation of the sensor in all axes is beyond the
scope of this paper. However, the trigonometric relationships and
general approach presented in this subsection can be used as the basis
for their derivation.

\subsection{Constrained Spaces using Scaling Vectors \label{subsec:CV4-ScalingVectors}}

If a viewpoint constraint can be formulated using scaling vectors,
as suggested for the feature geometry in Subsection \ref{subsec:CV-Feature-Geometry},
then the same approach can be equally applied to characterize the
constrained space of different viewpoint constraints. This subsection
introduces a generic formulation for integrating such viewpoint constraints
and proposes characterizing kinematic errors and the sensor accuracy
following this approach.

\subsubsection{Generic Formulation }

If the influence of an $i$ viewpoint constraint $c_{i}\in\widetilde{C}$
can be characterized by a scaling vector $\boldsymbol{\Delta}(c_{i})$
to span its corresponding constrained space, $\CC i{}{}$, then its
vertices $\VECJ V{}{\CC i{}{}}{}{}$ can be scaled using a generalized
formulation of Equation $\ref{eq:V-CV3}$:
\begin{equation}
\VECJ Vk{\CC i{}{}}{}{}=\VECJ Vk{\CC 1{}{}}{}{}-\boldsymbol{\Delta}(c_{i}).\label{eq:V-Scaling}
\end{equation}

\paragraph*{Integrating Multiple Constraints}

The characterization of a jointed constrained space, which integrates
several viewpoint constraints, can be computed using different approaches.
On the one hand, the constrained spaces can be first computed and
intersected iteratively using $\ACSG$ operations, as originally proposed
in Equation \ref{eq:C-C}. However, if the space spanned by such viewpoint
constraints can be formulated according to Equation \ref{eq:V-Scaling},
the characterization of the constrained space $\CC{}{}{}$ can be
more efficiently calculated by simply adding all scaling vectors:
\begin{equation}
\VECJ Vk{\CC{}{}{}}{}{}(\widetilde{C})=\VECJ Vk{\CC 1{}{}}{}{}-\sum_{\begin{aligned}c_{i}\in\widetilde{C},i\neq\end{aligned}
1}\boldsymbol{\Delta}(c_{i}).\label{eq:V-SUM}
\end{equation}

While the computational cost of $\ACSG$ operations is at least proportional
to the number of vertices between two surface models, note that the
complexity of the sum of Equation \ref{eq:V-SUM} is just proportional
to the number of viewpoint constraints. 

\subsubsection{Compatible Constraints}

Within this subsection, we propose further possible viewpoint constraints
that can be characterized according to the scaling formulation introduced
by Equation \ref{eq:V-SUM}. 
\begin{itemize}
\item \textbf{Kinematic errors: \label{subsec:Kinematic-error}} Considering
the fourth viewpoint constraint and the assumptions addressed in Subsection
\ref{subsec:Assumptions}, the maximal kinematic error $\epsilon$
is given by the sum of the alignment error $\epsilon_{e}$, the modeling
error of the sensor imaging parameters $\epsilon_{s}$, and the absolute
position accuracy of the robot $\epsilon_{r}$:
\[
{\left|\epsilon\right|=\left|\epsilon_{e}\right|+\left|\epsilon_{s}\right|+\left|\epsilon_{r}\right|}.
\]
Assuming that the total kinematic error has the same magnitude in
all directions, all vertices can be equally scaled. The vertices of
the $\ACC$ $\CC 4{}{}(\epsilon)$ are computed using the scaling
vector $\VECJ{\Delta}{}{}{}{}(\epsilon)$:
\[
\VECJ Vk{\CC 4{}{}}{}{}(\VECJ{\Delta}{}{}{}{}(\epsilon))=\VECJ Vk{\CC 1{}{}}{}{}-\VECJ{\Delta}{}{}{}{}(\epsilon)).
\]
\item \textbf{Sensor Accuracy}: If the accuracy of the sensor $a_{s}$ can
be quantified within the sensor frustum, then similarly to the kinematic
error, the manifold of the $\ACC$ $\CC 5{}{}(a_{s})$ can be characterized
using a scaling vector $\boldsymbol{\Delta}(a_{s})$:
\[
\VECJ Vk{\CC 5{}{}}{}{}(\VECJ{\Delta}{}{}{}{}(a_{s}))=\VECJ Vk{\CC 1{}{}}{}{}-\VECJ{\Delta}{}{}{}{}(a_{s})).
\]
\end{itemize}
Figure \ref{fig: CV-Scaling} visualizes an exemplary and more complex
scenario comprising an individual scaling of each vertex. This example
shows the high flexibility and adaptability of this approach for synthesizing
$\ACC$s for different viewpoint constraints according to the particular
necessities of the application in consideration.

\subsubsection{Summary}

The use of scaling vectors is an efficient and flexible approach to
characterize the $\ACC$ spanned by any viewpoint constraint. Within
this section, we considered a few viewpoint constraints that can be
modeled aligned to this formulation However, it should be noted that
this approach requires an explicit characterization of the individual
scaling vectors and the overall characterization may be limited by
the number of vertices of the base $\ACC$ $\CC 1{}{}(\text{\ensuremath{\CF}})$.
For instance, our model of the $\ACM$ considers just eight vertices.
Thus, any viewpoint constraint that requires a more complex geometrical
representation could be limited by this.

Moreover, note that if the regarded viewpoint constraint can be explicitly
characterized as a manifold in $SE(3)$, this $\ACC$ can then be
directly intersected with the rest of the constraints as originally
suggested by Eq. \ref{eq:C-C}, such an example is given for characterizing
the robot workspace (see Subsection \ref{subsec:CV9-Characterization}).
However, note that such an approach is generally computationally more
expensive; hence, this study recommends that the characterization
of viewpoint constraints using scaling vectors be prioritized whenever
possible.

\subsection{Occlusion Space\label{subsec:CV6-Occlusion Space}}

We consider the occlusion-free view of a feature (related terms: shadow
or bistatic effect) a non-negligible requirement that must be individually
assessed for each viewpoint. In the context of our framework, this
subsection outlines the formulation of a negative topological space
—the occlusion space $\CC 6{occl}{}$— to ensure the free visibility
of each viewpoint within the $\ACC$. Although other authors \parencite{Tarabanis.1996,Pito.1999,Reed.2000}
have already suggested the characterization of such spaces, the present
study proposes a new formulation of such a space aligned to our framework.
Our approach strives for an efficient characterization of $\CC 6{occl}{}$
using simplifications about the feature's geometry and the occlusion
bodies.

\subsubsection{Formulation}

If a feature $f$ is not visible from at least one sensor pose within
the $\ACC$, it can be assumed that at least one rigid body of the
$\ARVS$ is impeding its visibility. Thus, an occlusion space for
$f$ denoted as $\CC 6{occl}{}\subset SE(3)$ exists and a valid sensor
pose cannot be an element of it $\SP{}\notin\CC 6{occl}{}$. However,
it is well known that the characterization of such occlusion spaces
can generally be computationally expensive. Therefore, it seems inefficient
to formulate $\CC 6{occl}{}$ in the special Euclidean for all possible
sensor orientations. For this reason, and by exploiting the available
$\ACC$ spanned by other viewpoint constraints, $\CC 6{occl}{}$ can
be formulated based on the previously generated $\ACC$, a given sensor
orientation $\CR{}$, and the surface models of the rigid bodies $\kappa\in K$:

\begin{equation}
\CC 6{occl}{}:=\CC 6{occl}{}(f,\CC{}{}{},\CR{},K).\label{eq:CV3-1}
\end{equation}

Contrary to all other viewpoint constraints formulations (see Equation
\ref{eq:C-C-i}), it must be assumed that the occlusion space is not
a subset of the $\ACC$ $\CC 6{occl}{}\nsubseteq\CC{}{}{}$. Hence,
let Equation \ref{eq:C-C} be reformulated as the set difference of
the $\ACC$s of other viewpoint constraints and the occlusion space:
\begin{equation}
\CC{}{}{}=\left(\stackrel[i=1,i\neq6]{j}{\bigcap}\CC i{}{}\right)\backslash\CC 6{occl}{}.\label{eq:CV6-Difference}
\end{equation}
If the resulting constrained space results being a non-empty set,
$\CC{}{}{}\neq\emptyset$, there exists at least one valid sensor
pose for the selected sensor orientation with occlusion-free visibility
to the feature $f$.

\subsubsection{Characterization}

The present work proposes a strategy to compute $\CC 6{occl}{}$,
which is broken down into the following general steps. In the first
step, the smallest possible number of view rays is computed for detecting
potential occlusions. In the second step, by means of ray-casting
techniques\footnote{For the interested reader, we refer to \textcite{Roth.1982,Glassner.1989}
for a comprehensive overview of ray-casting techniques.}, view rays are tested for occlusion against all rigid bodies of the
$\ARVS$. Then, the occlusion space is characterized using a simple
surface reconstruction method using the colliding points of the rigid
bodies and some further auxiliary points. In the last step, the occlusion
space is integrated with the $\ACC$ spanned by other viewpoint constraints
as given by Eq. \ref{eq:CV6-Difference}.

\begin{figure*}[tbh]
\hfill{}%
\begin{minipage}[t]{0.32\textwidth}%
\subfloat[Step 1: A set of view rays ($\protect\VECJ{\varsigma}{g_{f,c}}{}{}{}$)
are computed for each $c$ surface point using the limits of the $\protect\ACC$
$\protect\CC 3{}{}$. \protect \\
Step 2: The occluding points ($\protect\VECJ qf{occl,\kappa}{}{}$)
of a rigid body are found using ray casting. \label{fig:Occ-Space-Step1=0000262}]{\centering{}\includegraphics{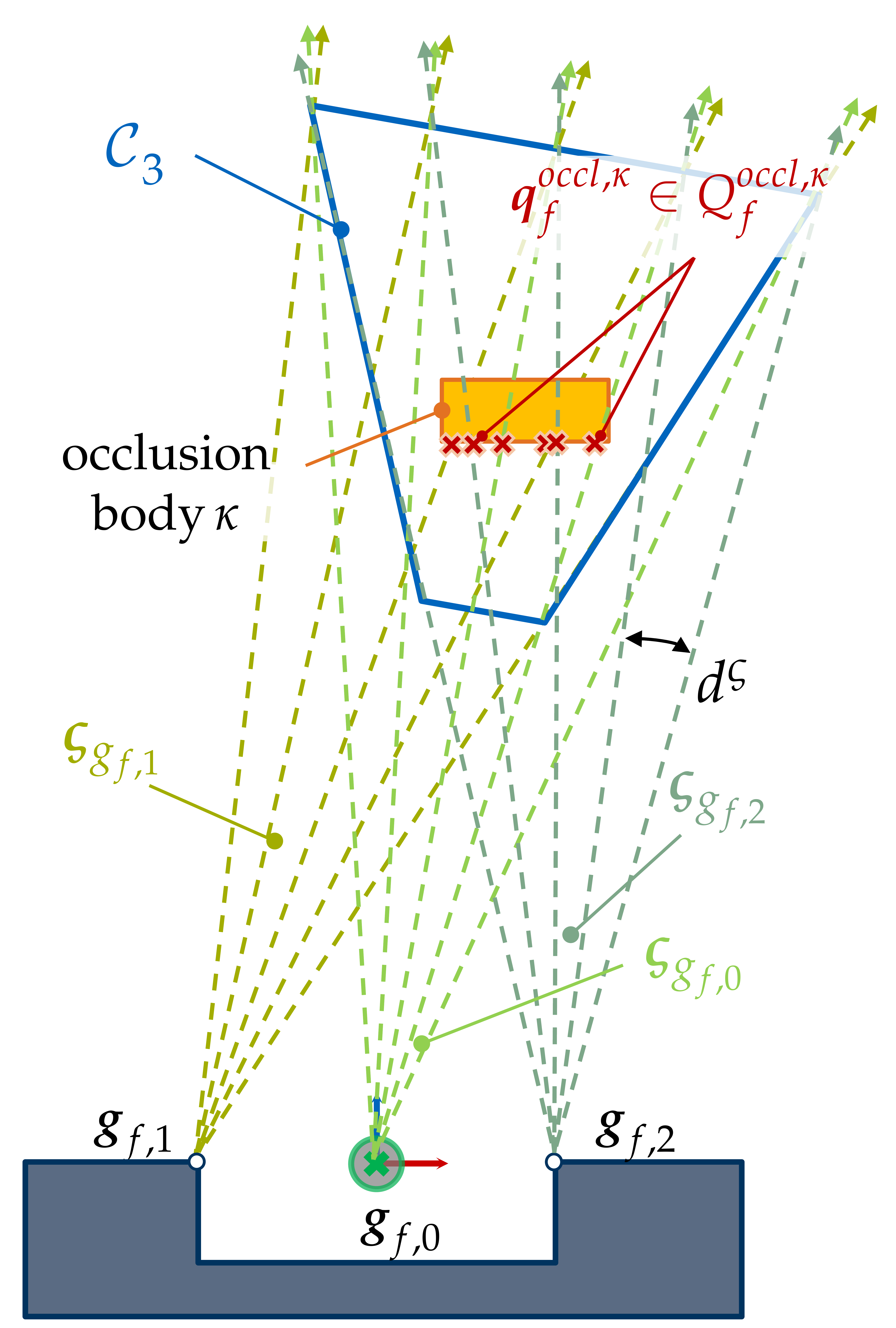}}%
\end{minipage}\hfill{}%
\begin{minipage}[t]{0.32\textwidth}%
\subfloat[Step 3: The occlusion rays ($\protect\VECJ{\varsigma}{g_{f,c}}{occl,\kappa}{}{}$)
are computed between each $c$ surface point and all occlusion points
($\protect\VECJ qf{occl,\kappa}{}{}$).\protect \\
Step 4: Select an additional point ($\protect\VECJ qf{occl,k}*{}$)
at each occlusion ray beyond the constrained space.\label{fig:Occ-Space-Step3a}]{\centering{}\includegraphics{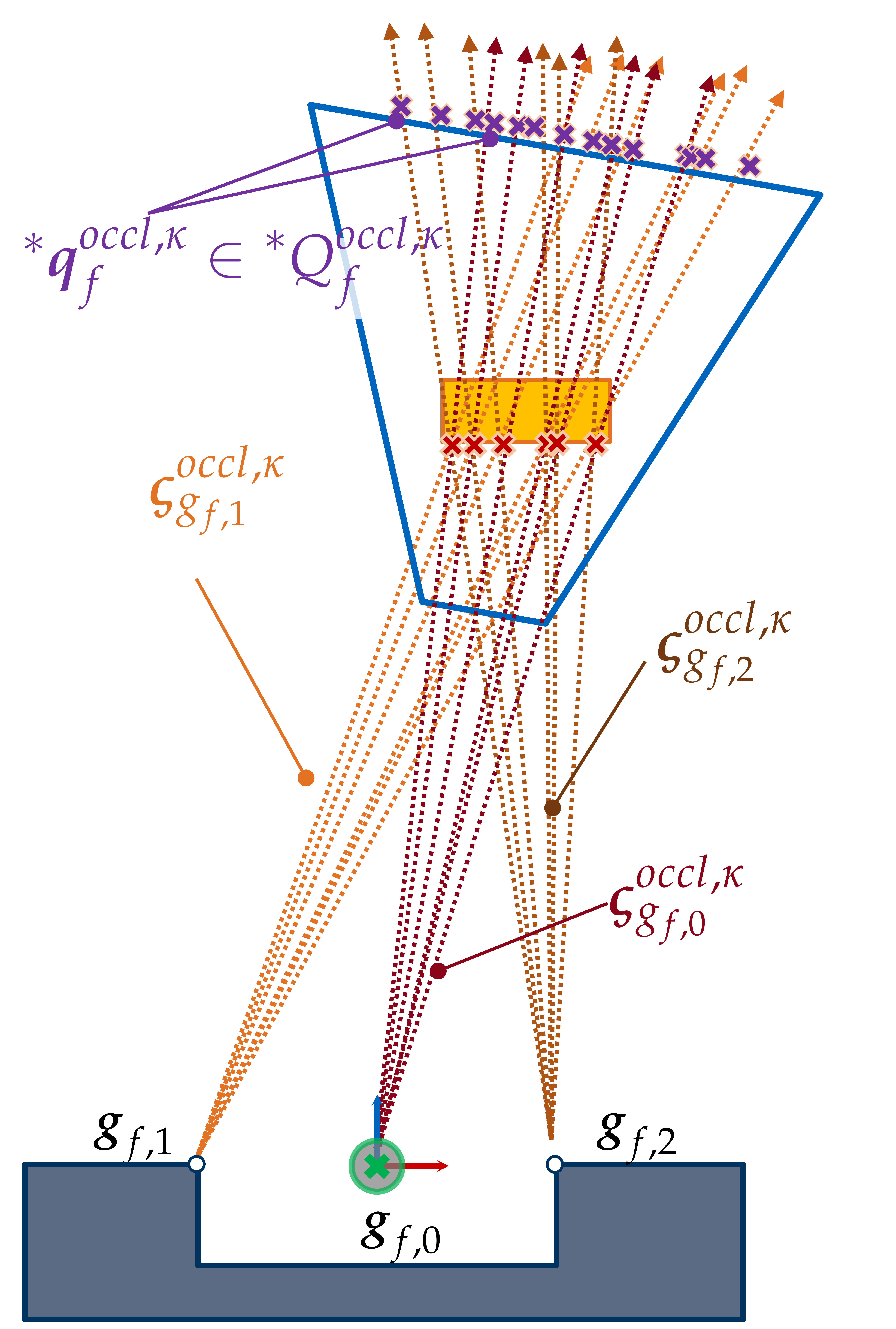}}%
\end{minipage}\hfill{}%
\begin{minipage}[t]{0.32\textwidth}%
\subfloat[Step 5: The occlusion space is characterized using the convex hull
spanned by all colliding points and the additional points lying at
the occlusion rays.\label{fig:Occl-Space-Step3b}]{\centering{}\includegraphics{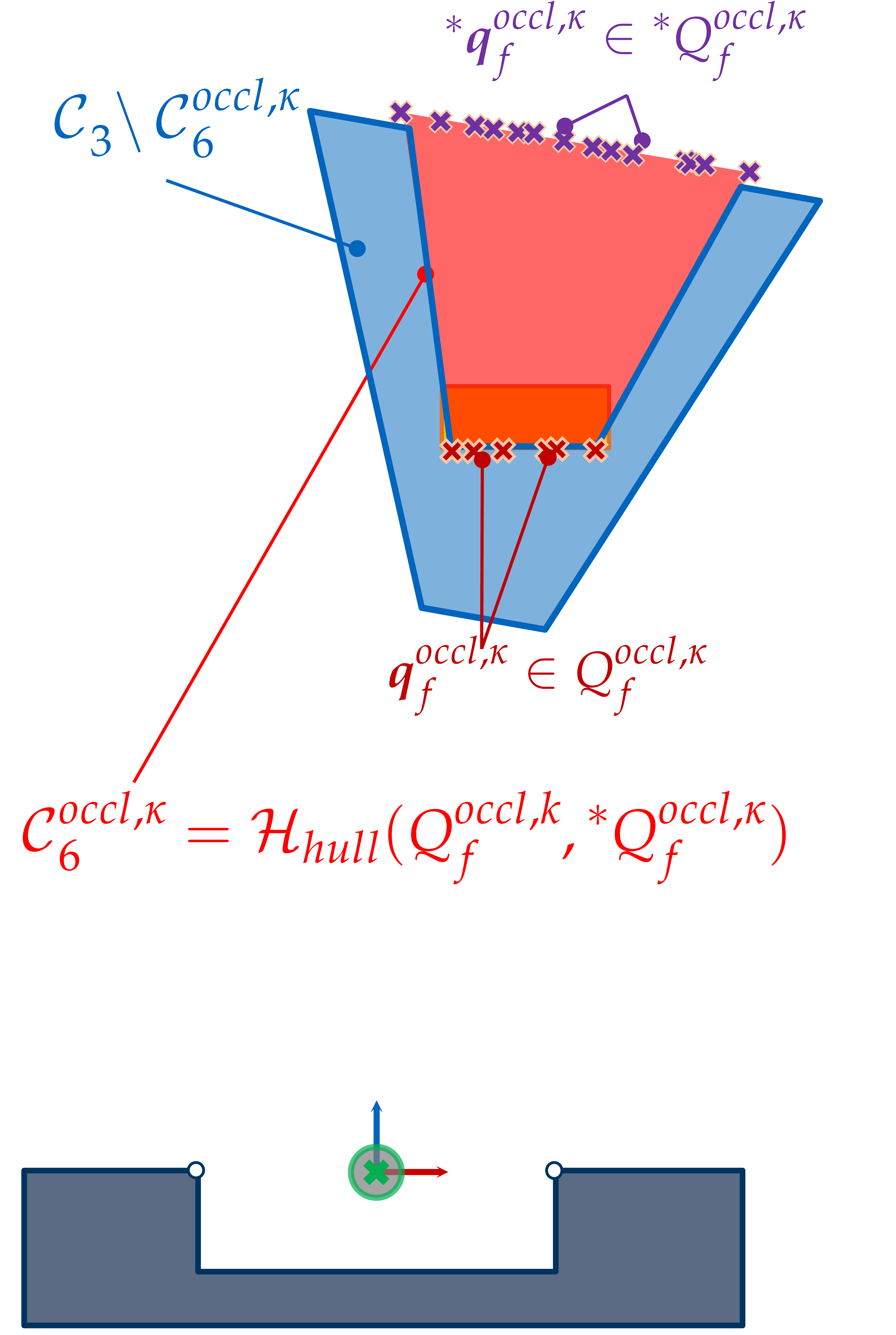}}%
\end{minipage}\hfill{}\caption{Overview of the computation steps of Algorithm \ref{alg:CV6} for
the characterization of the occlusion space $\protect\CC 6{occl,\kappa}{}$
induced by an occluding rigid body $\kappa$.\label{fig: OcclusionSpace}}
\end{figure*}

Algorithm \ref{alg:CV6} describes more comprehensively all steps
to characterize the occlusion space $\CC 6{occl}{}$ and Figure \ref{fig: OcclusionSpace}
provides an overview of the workflow and visualization of the expected
results of the most significant steps considering an exemplary occluding
object $\kappa_{1}$. A more comprehensive description for the characterization
of the view rays using the $\ACC$ characterized by other viewpoint
constraints can be found in the Appendix \ref{subsec:CV6-View-Rays}

\begin{algorithm}[tbh]
\caption{Characterization of the occlusion space $\protect\CC 6{occl}{}$ \label{alg:CV6}}

\begin{enumerate}
\item \textbf{\label{occl:1}}Compute a set of view rays $\VECJ{\varsigma}{g_{f,c}}{}{}{}(m,n)\in\Sigma$
for each surface point $\VECJ g{f,c}{}{}{}$ using a set of direction
vectors $\VECJ{\sigma}{m,n}{}{}{}$: 
\[
\VECJ{\varsigma}{g_{f,c}}{}{}{}(m,n)=\VECJ g{f,c}{}{}{}+\VECJ{\sigma}{m,n}{}{}{}(\sigma_{m}^{x},\sigma_{n}^{y},\text{\ensuremath{\CR{}}}).
\]
The direction vectors span a $m\times n$ grid of equidistant rays
with a discretization step size $d^{\varsigma}$. The aperture angles
of the view rays correspond to the maximal aperture of a previously
characterized $\ACC$ $\CC{}{}{}$.
\item Test all view rays, $\forall\VECJ{\varsigma}{g_{f,c}}{}{}{}(m,n)\in\Sigma$,
for occlusion against each rigid body $\kappa\in K$ using ray casting.
Let the collision points at the rigid bodies be denoted as:
\[
\VECJ qf{occl,\kappa}{}{}\in Q_{f}^{occl,\kappa}.
\]
\item Shoot an occlusion ray, $\VECJ{\varsigma}{g_{f,c}}{occl,\kappa}{}{}$,
from each surface point $\VECJ g{f,c}{}{}{}$ to all occluding points
of the set $\forall\VECJ qf{occl,\kappa}{}{}\in Q_{f}^{occl,\kappa}$
:
\[
\VECJ{\varsigma}{g_{f,c}}{occl,\kappa}{}{}(t)=\VECJ qf{occl,\kappa}{}{}+t\cdot(\VECJ qf{occl,\kappa}{}{}-\VECJ g{f,c}{}{}{}).
\]
\item Select one point, $\VECJ qf{occl,k}*{}$, from each occlusion ray
$\forall\VECJ{\varsigma}{g_{f,c}}{occl,\kappa}{}{}\in\Sigma^{occl,\kappa}$
considering that this must lie beyond the constrained space. Let these
points be elements of the set $\fr *{}Qf{occl,\kappa}$.
\[
\VECJ qf{occl,k}*{}\in(\fr *{}Qf{occl,\kappa},\VECJ{\varsigma}{g_{f,c}}{occl,\kappa}{}{}),\VECJ qf{occl,k}*{}\notin\CC{}{}{}
\]
\item \textbf{\label{occl:5}}Compute the convex hull spanned by all points
of $Q_{f}^{occl,k}$ and $\fr *{}Qf{occl,\kappa}$. The convex hull
corresponds to the manifold of $\CC{,6}{occl,\kappa}{}$:
\[
\CC 6{occl,\kappa}{}\leftarrow\mathcal{H}_{hull}(Q_{f}^{occl,k},\fr *{}Qf{occl,\kappa}).
\]
\item Compute the occlusion space, $\CC 6{occl,\kappa}{}$, for all rigid
bodies, $\forall\kappa\in K$ repeating Steps 2 until 5.
\item The occlusion space for all rigid bodies corresponds to the $\ACSG$
Boolean Union operation of all individual occluding spaces: 
\[
\CC 6{occl}{}=\stackrel[\kappa\in K]{}{\bigcup}\CC 6{occl,\kappa}{}.
\]
\item The occlusion space is integrated with the $\ACC$ spanned by other
viewpoint constraints using a $\ACSG$ Boolean Difference operation:
\[
\CC{}{}{}=\left(\stackrel[i=1,i\neq6]{j}{\bigcap}\CC i{}{}\right)\backslash\CC 6{occl}{}.
\]
\end{enumerate}
\end{algorithm}

\subsubsection{Verification}

For verification purposes, we consider an academic example, which
comprises an icosahedron occluding the sight of feature $f_{1}$.
The dimensions and location of the occluding object $\kappa_{1}$
are described in Table \ref{tab:Verification-Features}. Figure \ref{fig:CV6-Veri}
displays the related scene and the manifolds of the computed occlusion
space and corresponding occlusion-free space. In the first step, the
$\ACC$ for feature $f_{1}$ considering its geometry, i.e., $\CC 3{}{}$,
and the following sensor orientation: $\VECJ rs{}{f_{1}}{}(\alpha_{s}^{z}=\gamma_{s}^{x}=0,\beta_{s}^{y}=15{^{\circ}})$
was characterized. In the second step, the manifold of the occlusion
space, $\CC 6{occl}{}$, was synthesized following the steps described
in Algorithm \ref{alg:CV6}. A discretization step of $d^{\varsigma}=0.5{^\circ}$
was selected for computing the view rays $\VECJ{\varsigma}{g_{f,c}}{}{}{}$.

Figure \ref{fig:CV6-Verification} shows the rendered point cloud
and range image at one extreme viewpoint within the resulting occlusion-free
space. The rendered point cloud and image confirm that although the
occluding body lies within the frustum space of the viewpoint, the
feature and its entire geometry can still be completely captured.
As expected, the computation of the collision points using ray casting
was the most computationally expensive step with a total time of $t(\VECJ q{f,c}{occl,\kappa}{}{})=0.7$~s,
the total time required for characterizing the occlusion-free space
corresponded to $t(\CC 6{occl,\kappa}{})=1.32$~s.
\begin{center}
\begin{figure}[t]
\centering{}\includegraphics{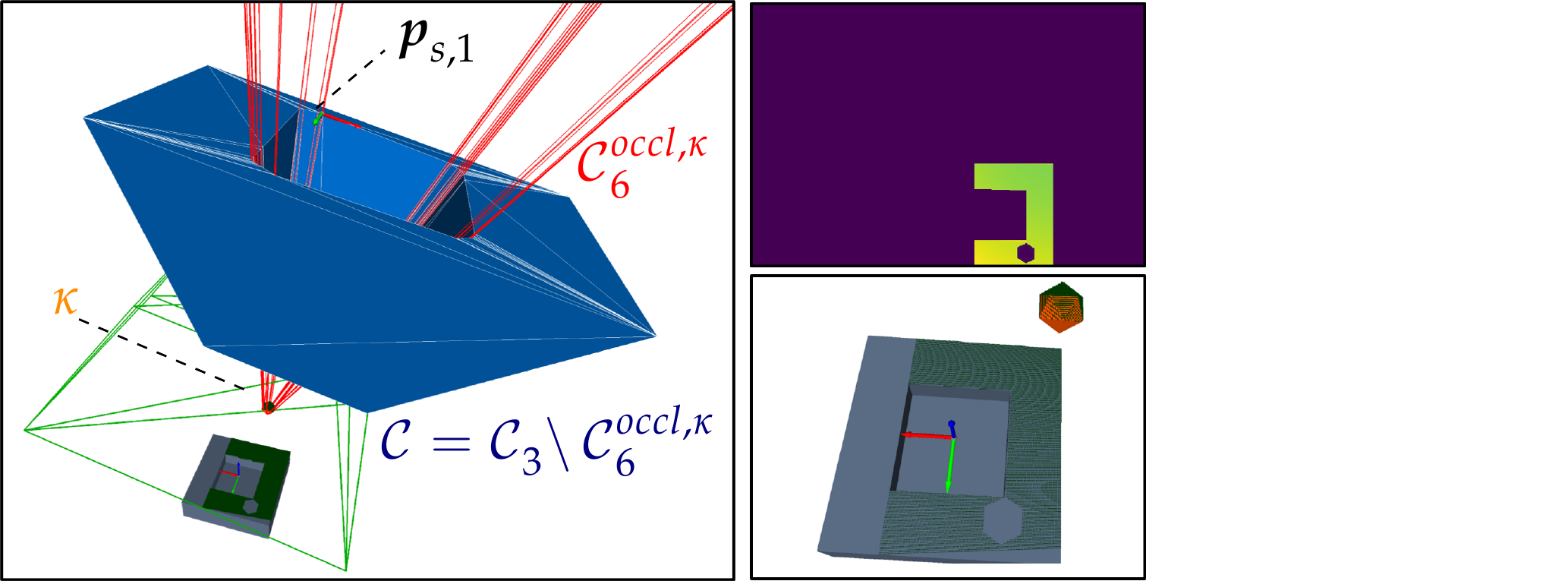}\caption{Visualization of the occlusion $\protect\ACC$ in $SE(3)$ (red manifold)
and the occlusion free space (blue manifold) to acquire a square feature
$f_{1}$ considering an occlusion body $\kappa$ (icosahedron in orange).
Verification of occlusion-free visibility: rendered scene (left image),
depth image (right image in the upper corner), and detailed view of
the rendered point cloud and object (right image in the lower corner)
at an extreme viewpoint $\protect\VECJ p{s,1}{}{}{}\text{\ensuremath{\protect\notin\protect\CC 6{occl,\kappa}{}}}$.\label{fig:CV6-Verification}}
\end{figure}
\par\end{center}

\subsubsection{Summary and Discussion}

Within this subsection, a strategy that combines ray-casting and $\ACSG$
Boolean techniques to compute an occlusion space $\CC 6{occl}{}$
was introduced. The present study showed that the $\CC 6{occl}{}$
manifold can be thoroughly integrated with the $\ACC$ spanned by
other viewpoint constraints complying with the framework proposed
within this publication. Moreover, to enhance the efficiency of the
proposed strategy, we considered a simplification of the feature geometry,
discretization of the viewpoint space, and the use of a previous computed
$\ACC$.

Due to the mentioned simplifications, it should be kept in mind that
contrary to the other $\ACC$s, the $\CC 6{occl}{}$ manifold does
not represent the explicit and real occlusion space and should be
treated as an approximation of it. For example, a significant source
of error regarding the accurate identification of all occluding points
is the chosen discretizing step size for the computations of the view
rays. This effect is known within ray-casting applications and can
also be observed in Figure \ref{fig:Occ-Space-Step1=0000262}, where
the right corner point of the colliding body is missed. Within the
context of the present study, comprising robot systems, we assume
that such simplifications can be safely taken into consideration if
the absolute position accuracy of the robot is considered for the
selection of the step size $d^{\varsigma}$. For example, assuming
a conservative absolute accuracy of a robot of $1\,\text{mm}$ and
a minimum working distance of $200\,\text{mm}$, it is reasonable
to choose a step size of $d^{\varsigma}\approx0.3{^\circ}$, using
the arc length formula $(1\,\text{mm}/200\,\text{mm})=0.005rad$.
Alternatively, more robust solutions can be achieved by scaling the
occlusion space with a safety factor.

Moreover, special attention must be given when computing the occlusion
space as suggested in Step 5 of Algorithm \ref{alg:CV6} for rigid
bodies with hollow cavities, e.g., a torus. It should be expected
that the result of the occlusion space will be more conservative.
A more precise characterization of the occlusion space falls outside
the scope of this paper and could be achieved using more sophisticated
surface reconstruction algorithms \parencite{Edelsbrunner.1983,Kazhdan.2006}.

Finally, it is important to mention that the characterization of the
occlusion space may lead to a non-watertight manifold, which may complicate
the further processing of the jointed $\ACC$. Thus, we recommend
computing the occlusion space as the last viewpoint constraint.

\subsection{Multisensor}

Considering the intrinsic nature of a range sensor, in its minimal
configuration, two imaging devices (two cameras or one camera and
one active projector) are necessary to acquire a range image. Therefore,
a range sensor can be regarded as a multisensor system. Up to this
point, it had been assumed that the $\ACC$ of a range sensor could
be characterized using just one frustum space $\CF$ and that the
resulting $\ACC$ integrates the imaging parameters and any other
viewpoint constraints of all imaging devices of a range sensor. On
the one hand, some of the previous introduced formulations and most
of the related work demonstrated that this simplification is in many
cases sufficient for computing valid viewpoints. On the other hand,
this assumption could also be regarded as restrictive and invalid
for some viewpoint constraints. For example, the characterization
of the occlusion-free space as described in Subsection \ref{subsec:CV6-Occlusion Space},
will not guarantee a free-sight to both imaging devices of a range
sensor. 

For this reason, our study assumes that each imaging device can have
individual and independent viewpoint constraints. As a result, we
consider that an individual $\ACC$ can be spanned for each imaging
device. Furthermore, this section outlines a generic strategy to merge
the individual $\ACC$s of multiple imaging devices to span a common
$\ACC$ that satisfies all viewpoint constraints simultaneously. 

To the best of our knowledge, none of the related work regarded viewpoint
constraints for the individual imaging devices of a range sensor or
for multisensor systems. In Section \ref{sec:Evaluation}, the scalability
and generality of our approach considering a multisensor system is
demonstrated with two different range sensors. 

\subsubsection{Formulation}

Our formulation is based on the idea that each imaging device can
be modeled independently and that all devices must simultaneously
fulfill all viewpoint constraints. First, considering the most straightforward
configuration of a sensor with a set of two imaging devices $\{s_{1},s_{2}\}\in\widetilde{S}$,
we can assume to have two different sensors with two different frustum
spaces $\CFn 1{}=\mathcal{I}_{0}(\VECJ ps{}{}{s_{1}},I_{s_{1}})$
and $\CFn 2{}=\mathcal{I}_{0}(\VECJ ps{}{}{s_{2}},I_{s_{2}})$ (cf.
Subsection \ref{subsec:Frustum-Space}). Thus, aligned to the formulation
from Subsection \ref{subsec:Frustum-Constraint-R3}, the core $\ACC$
for each sensor can be expressed as follows:
\[
\CC 1{s_{1}}{}(\VECJ rs{fix}{}{s_{1}},\CFn 1{},B_{f})\text{ and }\CC 1{s_{2}}{}(\VECJ rs{fix}{}{s_{2}},\CFn 2{},B_{f}).
\]

In a more generic definition that considers all viewpoint constraints
of $s_{1}$ or $s_{2}$ the $\ACC$ is denoted as $\CC{}{s_{1}}{}:=\CC 1{s_{1}}{}(f,\widetilde{C}^{s_{1}})$
or $\CC{}{s_{2}}{}:=\CC 1{s_{2}}{}(f,\widetilde{C}^{s_{2}})$.

Considering that the rigid transformation $\VECJ Ts{}{s_{1}}{s_{2}}$
between the two imaging devices of a sensor is known, it can be assumed
that a valid sensor pose for the first imaging device exists, if and
only if the second imaging device also lies within in its corresponding
$\ACC$ simultaneously. The formulation of such condition follows
\begin{equation}
\VECJ ps{}{}{s_{1}}\in\CC{}{s_{1}}{}\iff\VECJ ps{}{}{s_{2}}(\VECJ Ts{}{s_{2}}{s_{1}},\VECJ p{s_{1}}{}{}{})\in\CC{}{s_{2}}{},\label{eq:Lenses-Cond}
\end{equation}

supposing that the reference positioning frame for the sensor is the
first imaging device $s_{1}$.

If the condition from Eq. \ref{eq:Lenses-Cond} is valid, there must
exist a $\ACC$ for $s_{1}$ that integrates the viewpoint constraints
of both lenses being denoted as $\CC 7{s_{1.2}}{}$, which formulation
follows:
\[
\begin{aligned}\CC 7{s_{1,2}}{}=\CC 7{s_{1}}{}(\CC{}{s_{2}}{})= & \{\VECJ ps{}{}{s_{1}}\in\CC 7{s_{1,2}}{}\\
 & \mid\VECJ rs{fix}{}{s_{1}}\in\set Rsf{}{},\VECJ ps{}{}{s_{2}}(\VECJ Ts{}{s_{2}}{s_{1}},\VECJ ps{}{}{s_{2}})\in\CC{}{s_{2}}{}\}.
\end{aligned}
\]

The joint space can then be integrated with the $\ACC$ for $s_{1}$
employing a Boolean intersection:
\[
\CC{}{s_{1,2}}{}=\CC{}{s_{1}}{}\stackrel{}{\bigcap}\CC 7{s_{1,2}}{}.
\]

A more generic formulation of the $\ACC$ of $s_{1}$ being constrained
by all imaging devices $s_{t}\in\widetilde{S}$ is given:
\[
\CC{}{\widetilde{S}_{1}}{}=\CC{}{s_{1}}{}\stackrel[s_{t}\in\widetilde{S}]{}{\bigcap}\CC 7{s_{1},s_{t}}{}.
\]

Analogously, $\CC{}{\widetilde{S}_{2}}{}$ denotes the space for positioning
imaging device $s_{2}$ including the constraints of $s_{1}$. 

\begin{algorithm}[tbh]
\caption{Characterization of $\protect\ACC$ $\protect\CC{}{\widetilde{S}_{1}}{}$
to integrate viewpoint constraints of a second imaging device $s_{2}$.\label{alg:CV7}}

\begin{enumerate}
\item Compute the $\ACC$ of the first device considering a fixed orientation
$\VECJ rs{fix}{}{s_{1}}$and any further viewpoint constraints $\widetilde{C}^{s_{1}}$.
\[
\CC{}{s_{1}}{}(\VECJ rs{fix}{}{s_{1}},\widetilde{C}^{s_{1}}).
\]
\item Compute the $\ACC$ for the second imaging device taking into account
any viewpoint constraints and the previously defined orientation of
the first imaging device using the rigid orientation between both
devices ${\VECJ rs{fix}{}{s_{2}}(\VECJ rs{fix}{}{s_{1}})=\VECJ Rs{}{s_{2}}{s_{1}}\cdot\VECJ r{}{fix}{}{s_{1}}}$:
\[
\CC{}{s_{2}}{}(\VECJ rs{fix}{}{s_{2}}(\VECJ rs{fix}{}{s_{1}}),\widetilde{C}^{s_{2}}).
\]
\item Compute the sensor pose that the first device assumes when computing
$\CC{}{s_{2}}{}$ using the rigid transformation between both devices:
\[
\VECJ ps{\CC{}{s_{2}}{}}{}{s_{1}}=\VECJ Ts{}{s_{1}}{s_{2}}\cdot\VECJ ps{}{}{s_{2}}(\VECJ ts{}{}{s_{2}}=B_{f}).
\]
\item Duplicate the manifold of $\CC{}{s_{2}}{}$ and translate it to the
position vector of $\VECJ ps{\CC{}{s_{2}}{}}{}{s_{1}}$. The $\ACC$
$\CC 7{s_{1,2}}{}$ corresponds to this translated manifold: 
\[
\CC 7{s_{1,2}}{}=\text{\emph{translation}}(\CC{}{s_{2}}{},\VECJ ts{\CC{}{s_{2}}{}}{}{s_{1}}).
\]
\item Integrate $\CC 7{s_{1,2}}{}$ with the $\ACC$ of the first imaging
device using a Boolean Intersection operation:
\[
\CC{}{\widetilde{S}_{1}}{}=\CC{}{s_{1}}{}\stackrel{}{\bigcap}\CC 7{s_{1,2}}{}.
\]
\end{enumerate}
\end{algorithm}

\begin{figure*}[t]
\begin{centering}
\noindent\begin{minipage}[t]{1\textwidth}%
\hfill{}%
\begin{minipage}[t]{0.3\textwidth}%
\subfloat[Step 1: Compute $\protect\ACC$ $\protect\CC{}{s_{1}}{}(\protect\VECJ rs{fix}{}{s_{1}},\widetilde{C}^{s_{1}})$
for imaging device $s_{1}$. \label{fig:VC7-Step1}]{\centering{}\includegraphics{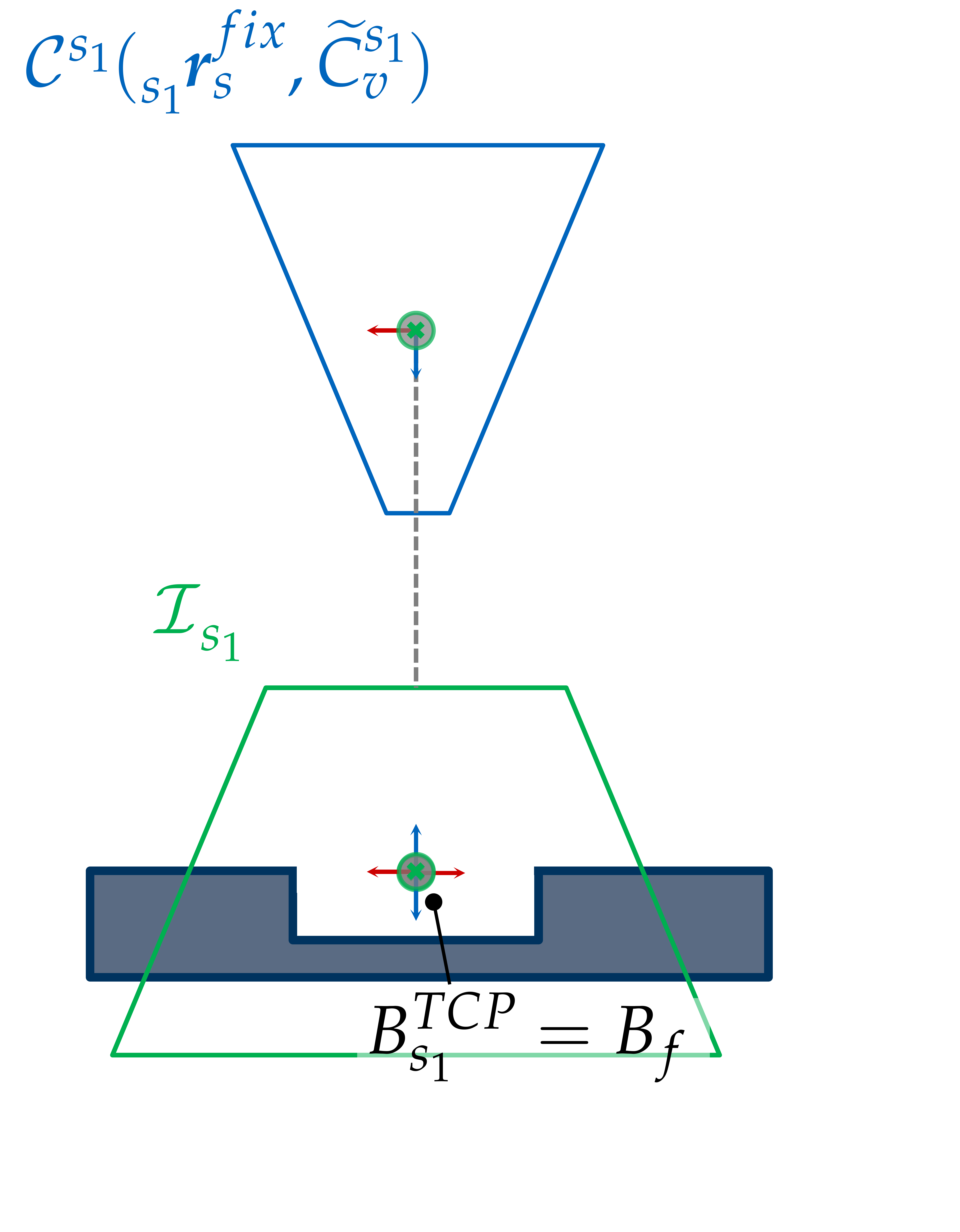}}%
\end{minipage}\hfill{}%
\begin{minipage}[t]{0.3\textwidth}%
\subfloat[Step 2: Compute $\protect\ACC$$\protect\CC{}{s_{2}}{}(\protect\VECJ rs{fix}{}{s_{2}}(\protect\VECJ rs{fix}{}{s_{1}}),\widetilde{C}^{s_{2}})$
for imaging device $s_{2}$.\label{fig:VC7-Step2}]{\centering{}\includegraphics{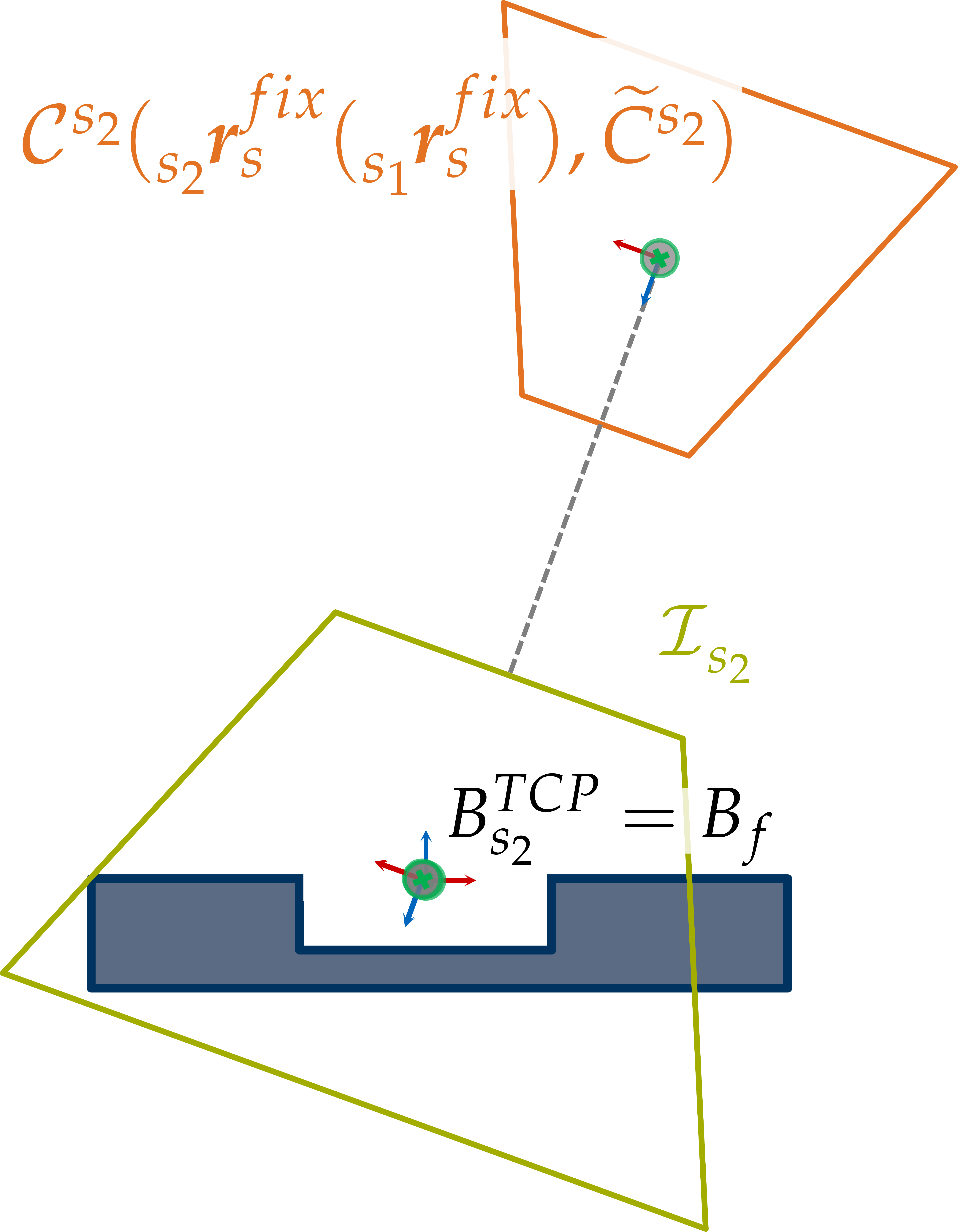}}%
\end{minipage}\hfill{}%
\begin{minipage}[t]{0.3\textwidth}%
\subfloat[Step 3: Compute the pose, $\protect\VECJ p{s_{1}}{f,s_{2}}{}{}$,
that the first sensor takes when computing $\protect\CC{}{s_{2}}{}$.\label{fig:VC7-Step3}]{\centering{}\includegraphics{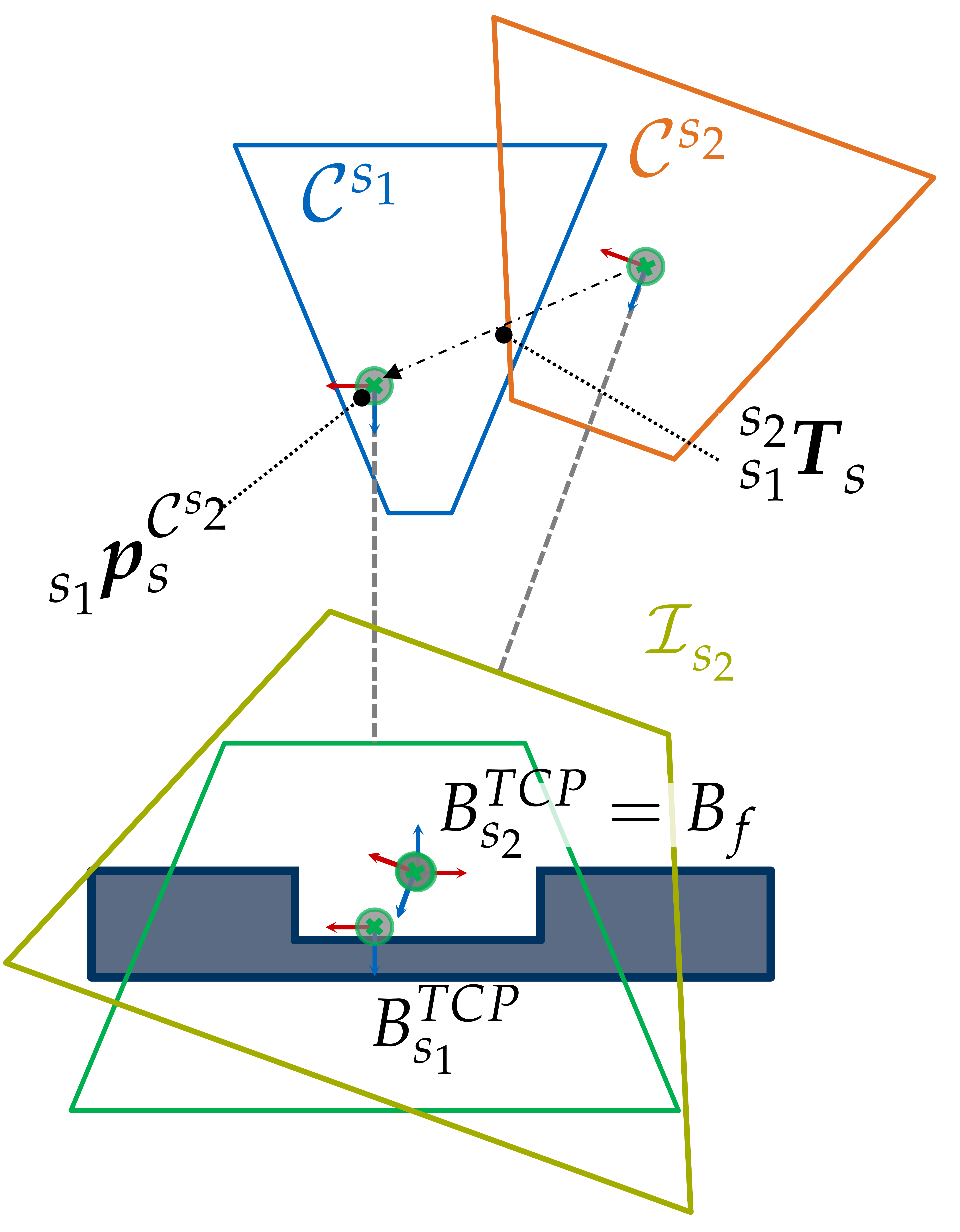}}%
\end{minipage}\hfill{}%
\end{minipage}\bigskip{}
\begin{minipage}[t]{0.9\textwidth}%
\begin{center}
\hfill{}%
\begin{minipage}[t]{0.3\textwidth}%
\subfloat[Step 4: Duplicate the manifold $\protect\CC{}{s_{2}}{}$ and translate
it to $\protect\VECJ ps{f,s_{2}}{}{s_{1}}$. $\protect\CC 7{s_{1,2}}{}$
denotes the resulting manifold at this position.\label{fig:VC7-Step4}]{\includegraphics{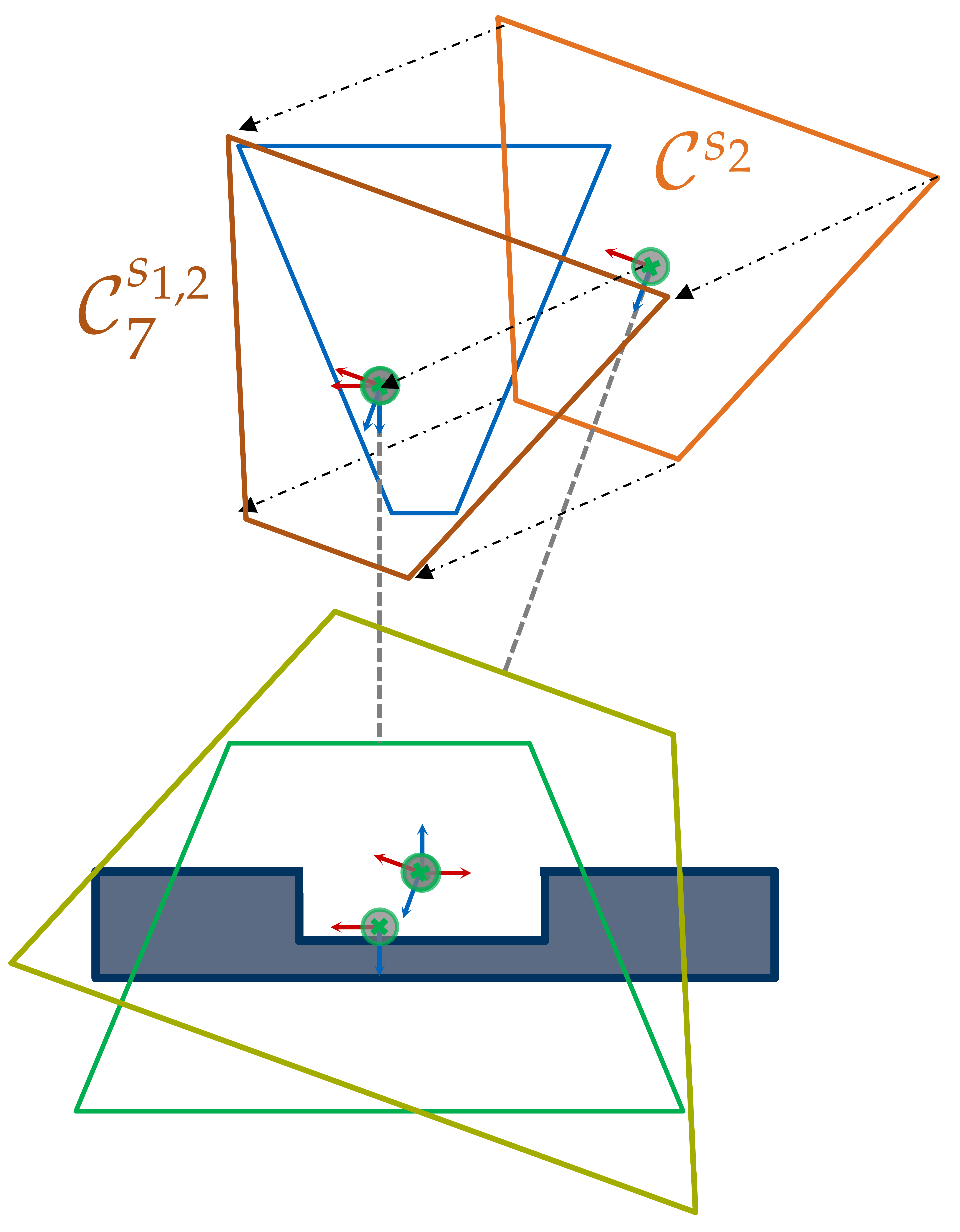}}%
\end{minipage}\hfill{}%
\begin{minipage}[t]{0.3\textwidth}%
\subfloat[Step 5: The intersection between $\protect\CC 7{s_{1,2}}{}$ and $\protect\CC{}{s_{2}}{}$
yields the space $\protect\CC{}{\widetilde{S}_{1}}{}$ for sensor
$s_{1}$, which integrates all viewpoint constraints of $s_{2}$.
\label{fig:VC7-Step5}]{\includegraphics{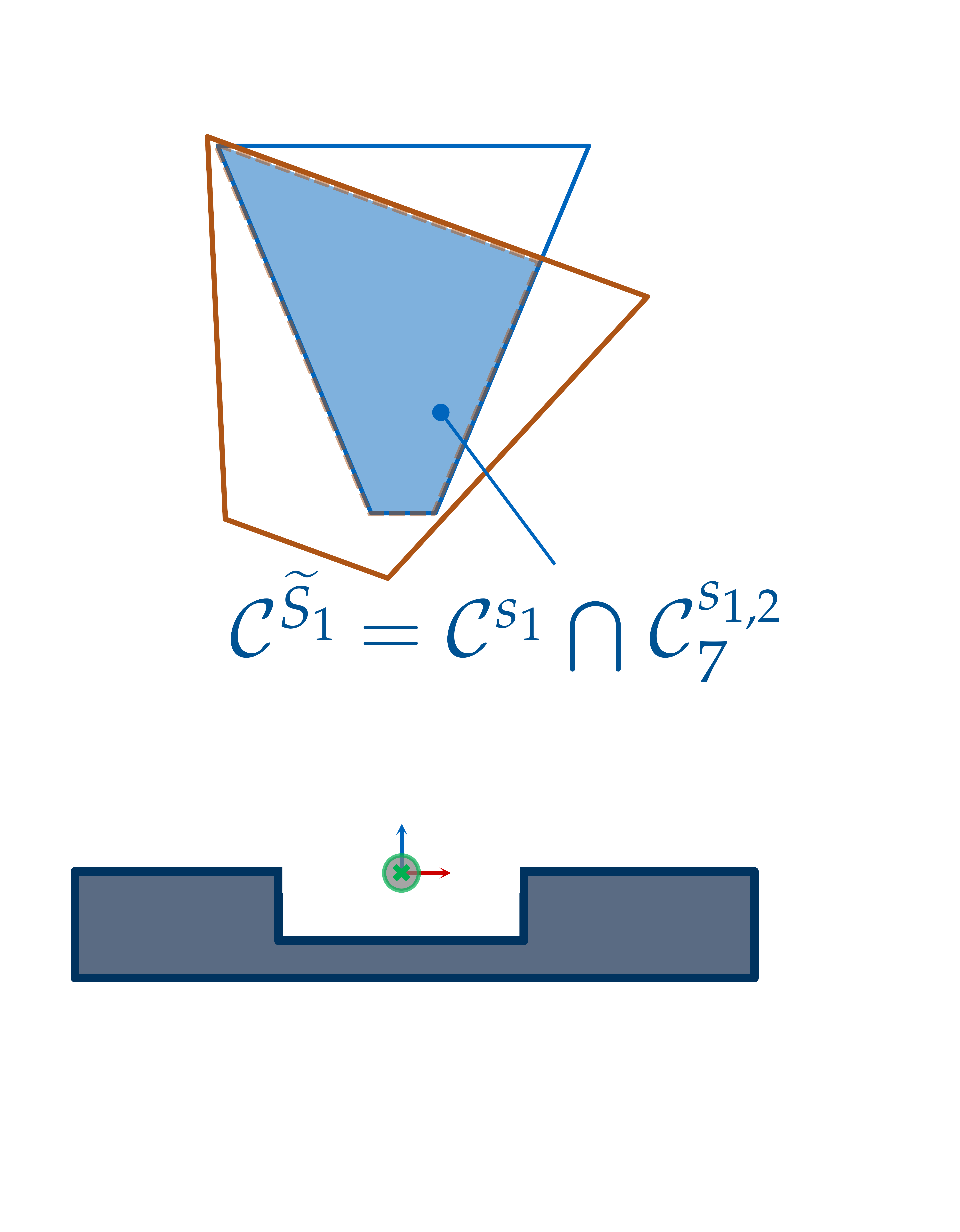}}%
\end{minipage}\hfill{}%
\begin{minipage}[t]{0.3\textwidth}%
\subfloat[Repeat Steps 1–5 for the second imaging device or duplicate and translate
manifold $\protect\CC{}{\widetilde{S}_{1}}{}$ for computing $\protect\CC{}{\widetilde{S}_{2}}{}$.
\label{fig:VC7-Step6}]{\includegraphics{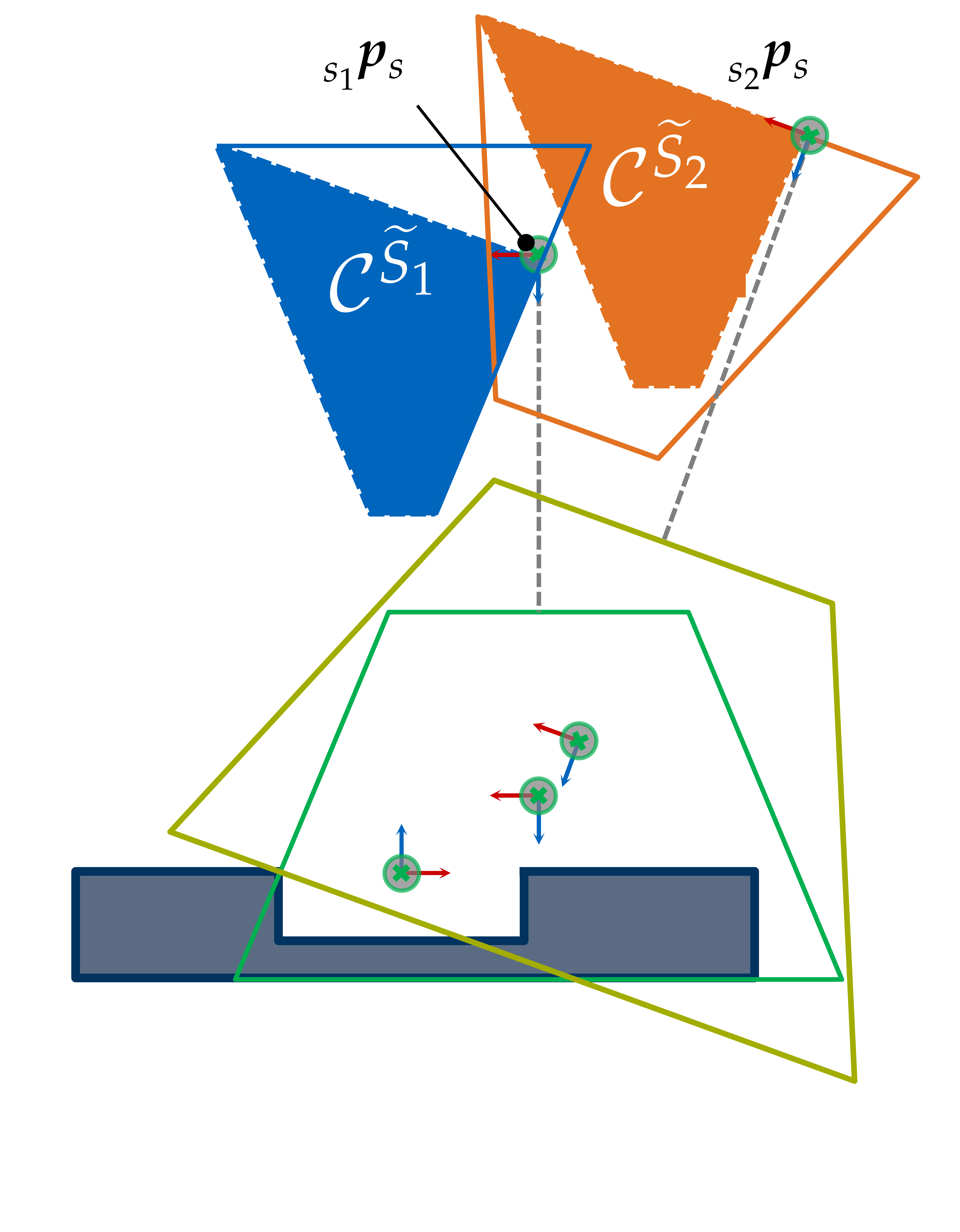}}%
\end{minipage}\hfill{}
\par\end{center}%
\end{minipage}
\par\end{centering}
\caption{Overview of the computation steps of Algorithm \ref{alg:CV7} to characterize
the $\protect\ACC$ $\protect\CC{}{\widetilde{S}_{1}}{}$ that integrates
viewpoint constraints of the two imaging devices ($s_{1}$, $s_{2}$)
from a range sensor.\label{fig:CV7-Char-Steps}}
\end{figure*}

\subsubsection{Characterization\label{subsec:CV7-Characterization}}

The $\ACC$ $\CC{}{\widetilde{S}_{1}}{}$, which comprises all constraints
of all imaging devices $s_{t}\in\widetilde{S}$, can be straightforwardly
characterized following the five simple steps given in Algorithm \ref{alg:CV7}.
Figure \ref{fig:CV7-Char-Steps} visualizes the interim manifolds
at each step to ultimately characterize the manifold $\CC{}{\widetilde{S}_{1}}{}$.
Finally, Figure \ref{fig:VC7-Step6} shows an exemplary extreme viewpoint,
where both imaging devices frames are within their respective $\ACC$s
$\VECJ ps{}{}{s_{1}}\in\CC{}{\widetilde{S}_{1}}{}$ and $\VECJ ps{}{}{s_{2}}\in\CC{}{\widetilde{S}_{2}}{}$
and the feature geometry lies within both frustum spaces.

The space $\CC{}{\widetilde{S}_{2}}{}$ for the second imaging device
can be computed analogously following the same steps. However, the
topology of the resulting $\CC{}{\widetilde{S}_{2}}{}$ will be identical
to $\CC{}{\widetilde{S}_{1}}{}$. Hence, instead of repeating the
steps described in Algorithm \ref{alg:CV7} for a second imaging device,
a more efficient alternative is to translate the manifold of $\CC{}{\widetilde{S}_{1}}{}$
to the position of $s_{2}$ at $\VECJ ps{}{}{s_{2}}(\VECJ ts{}{}{s_{2}}=B_{f})$
using the rigid translation
\[
\CC{}{\widetilde{S}_{2}}{}=\text{\emph{translation}}(\CC{}{\widetilde{S}_{1}}{},\VECJ Ts{}{s_{2}}{s_{1}}).
\]

Note that if two imaging devices are the same orientation, the resulting
$\ACC$s are identical, hence $\CC{}{\widetilde{S}_{1}}{}=\CC{}{\widetilde{S}_{2}}{}$.

\begin{figure}[tbh]
\begin{centering}
\includegraphics{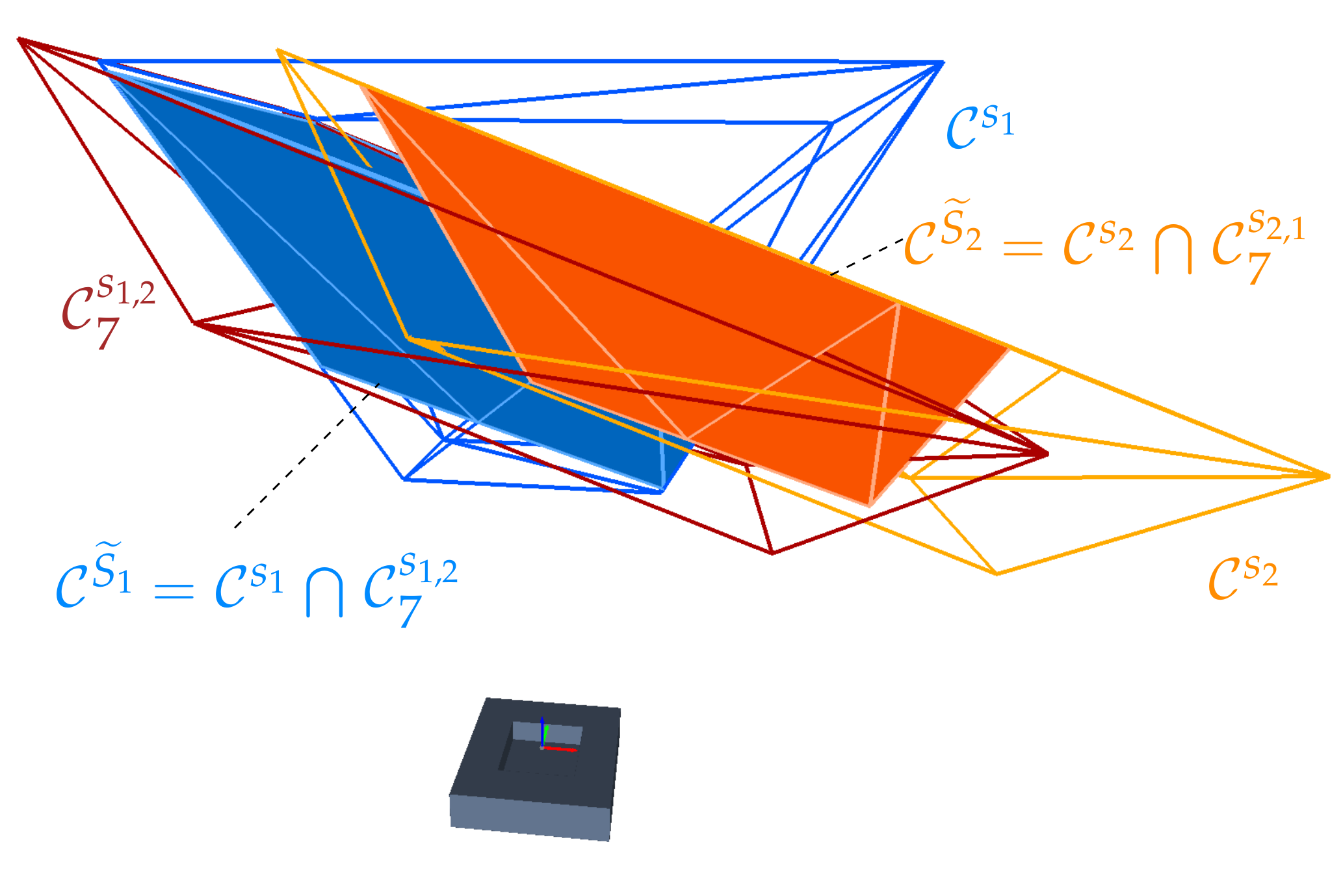}\caption{Characterization of the $\protect\ACC$ for the first sensor in $SE(3)$,
$\protect\CC{}{\widetilde{S}_{1}}{}$ (blue manifold), being delimited
by the $\protect\ACC$ of the second sensor, $\protect\CC{}{s_{2}}{}$(orange
manifold without fill), to acquire a square feature $f_{1}$. The
$\protect\CC{}{\widetilde{S}_{2}}{}$ (orange manifold) analogously
characterizes the $\protect\ACC$ for sensor $s_{2}$ considering
the constraints of $s_{1}$. \label{fig:VC7-Verification-Scene}}
\par\end{centering}
\end{figure}

\begin{figure}[tbh]
\centering{}\includegraphics{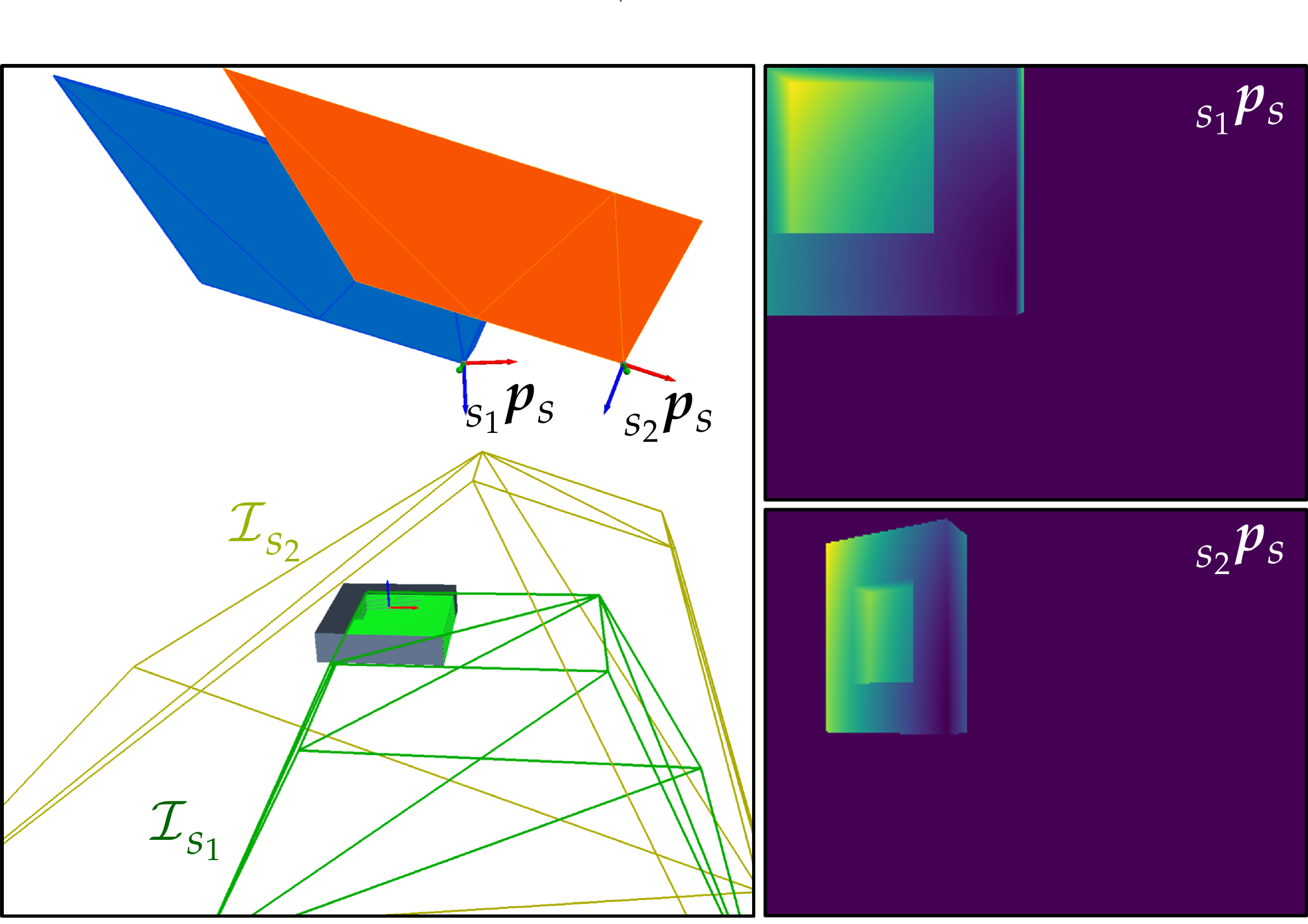}\caption{Verification of occlusion-free visibility at an extreme viewpoint
${\protect\VECJ ps{}{}{s_{1}}\in\protect\CC{}{\widetilde{S}_{1}}{}}$
and ${\protect\VECJ ps{}{}{s_{2}}\in\text{\ensuremath{\protect\CC{}{\widetilde{S}_{2}}{}}}}$:
rendered scene (left image), depth images of $s_{1}$ (right image
in the upper corner) and $s_{2}$ (right image in the lower corner).\label{fig:VC7-v1}}
\end{figure}

\subsubsection{Verification}

The joint constrained spaces $\CC{}{\widetilde{S}_{1}}{}$ and $\CC{}{\widetilde{S}_{2}}{}$
for $s_{1}$ and $s_{2}$ were computed according to the steps provided
in Algorithm \ref{alg:homeomorphism P-Space} for acquiring feature
$f_{1}$. The imaging parameters of both sensors and rigid transformation
between them are given in Table \ref{tab:Image Paremeters s1 and s2}. 

First, the $\ACC$s of both sensors were spanned considering its imaging
parameters and a null orientation of the first sensor, i.e., ${\VECJ r{s,1}{}f{}(\alpha_{s}^{z}=\beta_{s}^{y}=\gamma_{s}^{x}=0{^\circ})}$.
The individual constrained spaces were computed for each sensor considering
the constrained space affected by the feature geometry, i.e., $\CC 3{s_{1}}{}$
and $\CC 3{s_{2}}{}$. Since the frustum space of the second sensor
always lies within the first one, we additionally considered a fictitious
accuracy constraint for the depth of the second sensor, $\CC 5{s_{2}}{}(a_{s_{2}}(z_{min}=500\,\text{mm},z_{max}=700\,\text{mm}))$,
to limit its working distance.

Figure \ref{fig:VC7-Verification-Scene} visualizes the described
scene and resulting manifolds of the constrained spaces. Figure \ref{fig:VC7-v1}
displays the frustum spaces, rendered depth images of both sensors,
and the resulting point cloud at an extreme viewpoint. The rendered
images provide a visual verification of our approach demonstrating
that $f_{1}$ is visible from both sensors. Note that the second device
represents an active structured light projector (see in Table \ref{tab:Image Paremeters s1 and s2}.) 

The total computation time of the constrained space was estimated
at $200\,ms$. The computation time depended mainly on the intersection
from Step 5, which corresponded to $100\,ms$. The computation results
just apply to this case. Such analyses are difficult to generalize
since the complexity of Boolean operations depends decisively on the
number of vertices of the manifolds of the occluding objects.

\subsubsection{Summary }

This subsection introduced the formulation and characterization of
a $\ACC$, which considers the intrinsic configuration of range sensors
comprising at least two imaging devices. Our approach enables the
combination of individual viewpoint constraints for each device and
the characterization of a $\ACC$, which fulfills simultaneously all
viewpoint constraints from all imaging devices. 

The strategy proposed in Algorithm \ref{alg:CV7} demonstrated to
be valid and efficient to characterize the manifolds of such $\ACC$.
Nevertheless, we do not dispose alternative approaches for its characterization.
For instance, if the frustum spaces are intersected initially, the
resulting frustum space can be used as the base for spanning the rest
of the constraints. However, we consider the steps proposed within
this subsection more traceable, modular, and extendable to consider
further constraints, multisensor systems, or even transferable to
similar problems. For example, a variation of the algorithm could
be applied for maximizing or guaranteeing the registration space between
two different viewpoints, which represents a fundamental challenge
within many vision applications \parencite{Bauer.2021}. 

\subsection{Robot Workspace \label{subsec:CV8}}

This section outlines the formulation of the robot workspace as a
further viewpoint constraint to be seamlessly and consistently integrated
with the other $\ACC$s.

\subsubsection{Formulation}

A viewpoint can be considered valid if a sensor pose is reachable
by the robot: hence, it lies within the robot workspace $\SP{}\in\mathcal{W}_{r}$.
The formulation of a constraint can then be straightforwardly formulated
as follows:
\begin{equation}
\CC 8{}{}=\mathcal{W}_{r}=\{\SP{}\in\CC{,8}{}{}\mid\SP{}\in\mathcal{W}_{r}\}.\label{eq:CV8}
\end{equation}

\subsubsection{Characterization and Verification}

\begin{figure}[t]
\begin{centering}
\includegraphics{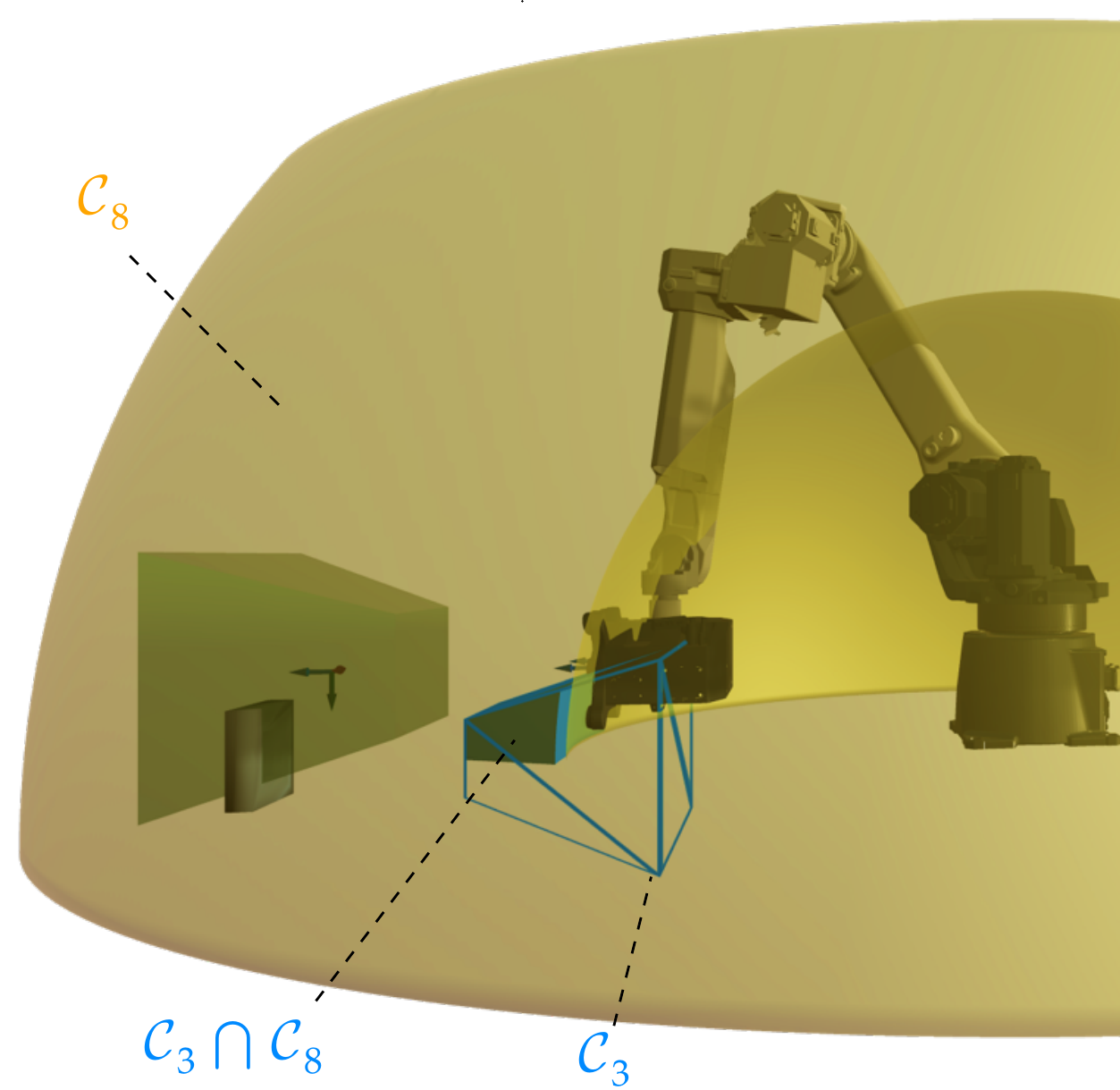}
\par\end{centering}
\caption{Characterization of the robot workspace as a further $\protect\ACC$
$\protect\CC 8{}{}$ and integration with other $\protect\ACC$s,
e.g., here $\protect\CC 3{}{}$, using a $\protect\ACSG$ Intersection
Operation. \label{fig:CV8}}
\end{figure}

In our work, we assume that the robot workspace is known and can be
characterized by a manifold in the special Euclidean. Considering
this assumption, the constrained space $\CC 8{}{}$ can be seamlessly
intersected with the rest of the viewpoint constraints. Figure \ref{fig:CV8}
shows an exemplary scene for acquiring feature $f_{1}$ and the resulting
constrained manifold $\CC 3{}{}\stackrel{}{\bigcap}\CC 8{}{}$, which
considers a robot with a workspace of a half-sphere and a working
distance of $1000\,\text{mm}$–$1800\,\text{mm}$ and the $\ACC$
manifold $\CC 3{}{}$ spanned by the feature geometry.

\subsubsection{Discussion}

A more comprehensive formulation and characterization of the robot
workspace to consider singularities requires more detailed modeling
of the robot kinematics. Furthermore, our study has not considered
the explicit characterization of the collision-free space of a robot,
which had been the focus of exhaustive research in the last three
decades. We assume that an explicit proof for collision must be performed
in the last step for a selected sensor pose within the $\ACC$. Nevertheless,
we consider that our approach contributes substantially to a significant
problem simplification by delimiting the search space to compute collision-free
robot joint configurations more efficiently. 

\subsection{Multi-Feature Spaces\label{subsec:CV9}}

Up to this point, our work has outlined the formulation and characterization
of $\ACC$s to acquire just one feature. Within this subsection we
briefly outline the characterization of a $\text{\ensuremath{\ACC}}$,$\CC{}{}F$,
which allows capturing a set of features $F$ and the simultaneous
fulfillment of all viewpoint constraints from all features $f_{m}\in F$
with $m=1,\dots,n$.

\subsubsection{Characterization\label{subsec:CV9-Characterization}}

\begin{figure}[t]
\begin{centering}
\begin{minipage}[t]{0.8\columnwidth}%
\begin{center}
\includegraphics{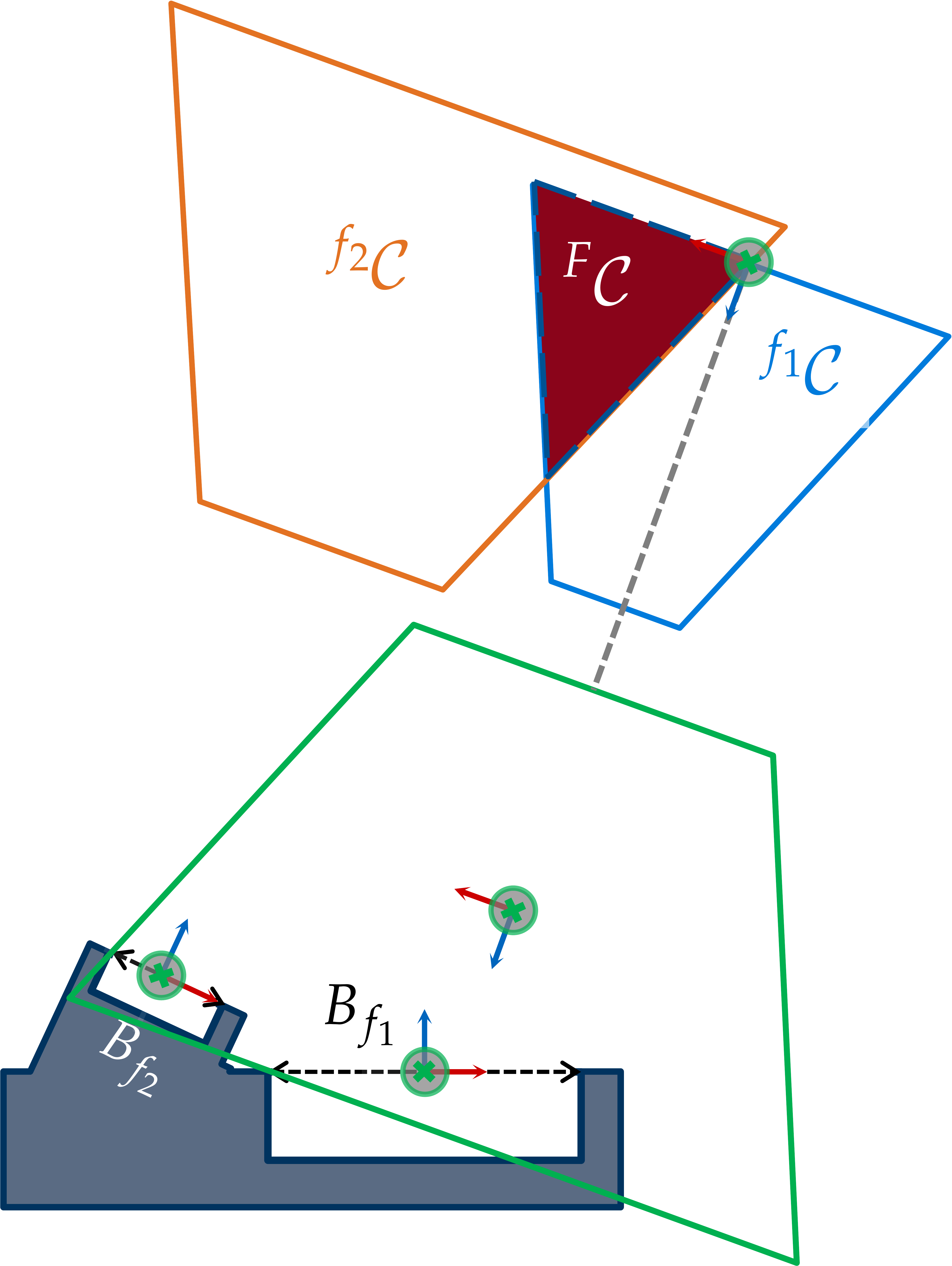}
\par\end{center}%
\end{minipage}
\par\end{centering}
\caption{The characterization of the $\protect\ACC$ spanned by two features
is computed by  intersecting its constrained spaces using the same
sensor orientation.\label{fig:CV9}}
\end{figure}

The characterization of $\CC{}{}F$ can be seamless achieved according
to the two steps described in Algorithm \ref{alg:CV9}. In the first
step, the $\ACC$ for all $n$ features, $\CC{}{}{f_{m}}$, are characterized
considering a fixed sensor orientation $\CR{}$ and the individual
constraints $\widetilde{C}(f_{m})$. Then the constrained space $\CC{}{}F$
is synthesized by intersecting all individual constrained spaces.
Figure \ref{fig:CV9} shows the characterization of such a space for
the acquisition of two features. 

\begin{algorithm}[tbh]
\caption{Integration of $\protect\ACC$ for multiple features \label{alg:CV9}}

\begin{enumerate}
\item Compute all $n$ $\ACC$s for each feature $\forall f_{m}\in F$ with
$m=1,\dots,n$. 
\[
\CC{}{}{f_{m}}:=\CC{}{}{}(\CR{},\widetilde{C}(f_{m}))
\]
\item Compute the joint $\ACC$ by intersecting all $n$ $\ACC$s: 
\[
\CC{}{}F=\CC 9{}F=\stackrel[\begin{aligned}f_{m}\in F\end{aligned}
]{n}{\bigcap}\CC{}{}{f_{m}}
\]
\end{enumerate}
\end{algorithm}

\subsubsection{Verification}

\begin{figure}[t]
\centering{}\includegraphics{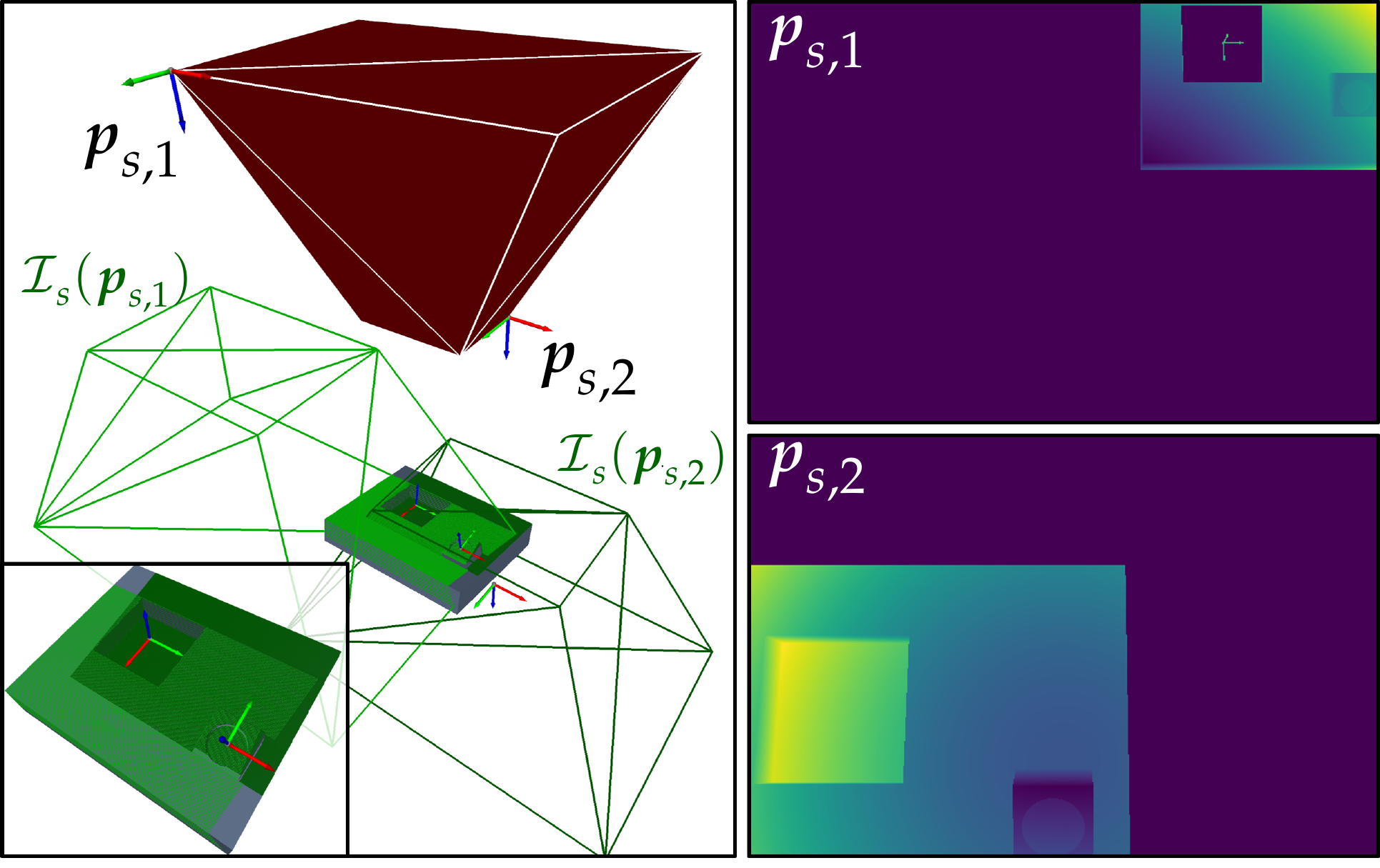}\caption{Verification of the $\protect\CC{}{}F$ at two extreme viewpoints
$\{\protect\VECJ p{s,1}{}{}{},\protect\VECJ p{s,2}{}{}{}\}\text{\ensuremath{\in\protect\CC{}{}F}}$:
rendered scene (left image) and depth images of $\protect\VECJ p{s,1}{}{}{}$(right
image in the upper corner) and $\protect\VECJ p{s,2}{}{}{}$ (right
image in the lower corner). \label{fig:CV9-v1}}
\end{figure}

To verify the proposed characterization of a constrained space for
acquiring multiple features, we outlined an exemplary use case comprising
two features $\{f_{1},f_{2,}\}\in F$ (see Table \ref{tab:Verification-Features}).
We computed the space, $\CC{}{}F$, according to the steps of Algorithm
\ref{alg:CV9}, considering an orientation of $\VECJ rs{}{f_{1}}{}(\alpha_{s}^{z}=\beta_{s}^{y}=0{^\circ},\gamma_{s}^{x}=-10{^\circ},)$
relative to the feature $f_{1}$. Figure \ref{fig:CV9-Verification-Scene}
shows the described scene and visualizes the resulting joint space.
The rendered scene and range images of Figure \ref{fig:CV9-v1} confirm
the validity of the $\CC{}{}F$, demonstrating that both features
can be simultaneously acquired at two extreme viewpoints within this
space.

\subsubsection{Summary and Discussion}

Within this subsection, we demonstrated that $\ACC$s from different
features can be seamlessly combined to span a topological space that
guarantees the acquisition of these features satisfying simultaneously
the individual feature viewpoint constraints. 

The current study assumes that the sensor orientation can be arbitrarily
chosen and that the features can be acquired jointly by the sensor.
In most applications, such assumptions cannot always be met and the
following fundamental question arises: which features can be acquired
simultaneously and which is an adequate sensor orientation? This questions
fall outside the scope of this paper and yields the motivation of
our ongoing research, which addresses the efficient combination of
$\ACC$s to tackle the superordinated $\AVPP$.

\subsection{Constraints Integration Strategy\label{subsec:Constraints-Integration-Strategy} }

The integration of viewpoint constraints can be considered to be commutative,
i.e., the order of computation and integration of the constraints
do not affect the characterization of the final constrained space.
However, due to the diverse computation techniques that our our framework
considers, a well-thought-out strategy may contribute to increase
the computational efficiency of the overall process. In this publication,
we outline one possible and simple strategy described in Algorithm
\ref{alg:Viewpoint-Constraints-Integration} to integrate all viewpoint
constraints into a single $\ACC$. 

The optimal integration of constraints falls outside the scope of
this publication. Moreover, we consider that an optimal and efficient
strategy can be tailored just considering the individual application
and its specific constraints.

\begin{algorithm}[tbh]
\caption{Strategy for the integration of viewpoint constraints. \label{alg:Viewpoint-Constraints-Integration}}

\begin{enumerate}
\begin{onehalfspace}
\item Consider a fixed sensor orientation $\VECJ rs{fix}{}{s_{1}}$ for
the reference imaging device $s_{1}$.
\end{onehalfspace}
\item Compute the $\ACC$s manifolds 
\[
\CC{1-6}{s_{t}}{f_{m}}(\VECJ rs{}{}{s_{t}}(\VECJ rs{fix}{}{s_{1}}))
\]
of imaging device $s_{t}$ for each feature $\forall f_{m}\in F$
considering the sensor orientation of the first device $\VECJ rs{fix}{}{s_{1}}$
and the viewpoint constraints $1-6$.
\begin{onehalfspace}
\item Compute the $\ACC$ of all features for sensor $s_{t}$:
\[
\CC{}{s_{t}}F=\stackrel{n}{\bigcap}\CC{1-6}{s_{t}}{f_{m}}.
\]
\item Repeat Steps 1–3 for all imaging devices $\forall s_{t}\in\widetilde{S}$.
\item Compute the $\ACC$ for all $u$ imaging devices, e.g., for $s_{1}$:
\[
\CC{}{\widetilde{S}_{1}}F=\CC{}{s_{1}}F\stackrel[s_{t}\in\widetilde{S}]{}{\bigcap}\CC 7{s_{1},s_{t}}F.
\]
\item Intersect the robot workspace to obtain the final $\ACC$ b, e.g.,
for $s_{1}$: 
\[
\CC{}{\widetilde{S}_{1}}F=\CC{}{\widetilde{S}_{1}}F\bigcap\CC 8{}{}.
\]
\end{onehalfspace}
\end{enumerate}
\end{algorithm}

\section{Evaluation \label{sec:Evaluation}}

Within this section, a comprehensive evaluation of the constraint
formulations and their integration is undertaken. First, Subsection
\ref{subsec:Simulation-Analysis} verifies the formulations of all
regarded viewpoint constraints of our work based on an academic example.
In Subsection \ref{subsec:Experimental-Analysis}, the framework's
broad generality and applicability is evaluated within an industrial
$\ARVS$ comprising two different sensors.

\begin{table*}[tbh]
\caption{Overview and description of the viewpoint constraints considered for
the simulated-based analysis.\label{tab:VC-Verification}}

\centering{}\begin{center}
\noindent\resizebox{1.0\textwidth}{!}{
\begin{tabular}{>{\centering}p{0.15\textwidth}>{\raggedright}p{0.5\textwidth}>{\raggedright}p{0.25\textwidth}}
\toprule \textbf{Viewpoint Constraint} & \multicolumn{1}{>{\centering}p{0.5\textwidth}}{\textbf{Description}} & \multicolumn{1}{>{\centering}p{0.2\textwidth}}{\textbf{Approach}}\tabularnewline
\midrule 1 & Two sensors($s^{1}$, $s^{2}$) with two imaging devices each: $\{s_{1}^{1},s_{2}^{1},s_{3}^{2},s_{4}^{2}\}\in\widetilde{S}$.
The imaging parameters of all devices are specified in Table \ref{tab:Image Paremeters s1 and s2}. & Linear algebra and geometry\tabularnewline
\midrule 2 & Relative orientation to the object's frame: $\VECJ rs{}o{}(\alpha_{s}^{z}=\gamma_{s}^{x}=0,\beta_{s}^{y}=6.64{^\circ})$. & Linear algebra and geometry\tabularnewline
\midrule 3 & A planar rectangular object with three different features $\{f_{1},f_{2},f_{3}\}\in F$
(see Table \ref{tab:Verification-Features}) & Linear algebra, geometry, and trigonometry\tabularnewline
\midrule 4–5 & The workspace of the second imaging device is restricted in the $z$-axis
to the following working distance $z_{s_{2}}>450\,\text{mm}$. & Linear algebra and geometry\tabularnewline
\midrule 6 & Two objects with the form of an icosahedron ($\kappa_{1}$) and a
octahedron ($\kappa_{2}$) occlude the visibility of the features. & Linear algebra, ray-casting, and $\ACSG$ Boolean Operations\tabularnewline
\midrule 7 & All constraints must be satisfied by all four imaging devices simultaneously. & Linear algebra and $\ACSG$ Boolean Operations\tabularnewline
\midrule 8 & Both sensors are attached to a 6-axis industrial robot. The robot
has a workspace of a half-sphere with a working distance of $1000\,\text{mm}$–$1800\,\text{mm}$. & $\ACSG$ Boolean Operation\tabularnewline
\midrule 9 & All features from the set $G$ must be captured simultaneously. & $\ACSG$ Boolean Operation \tabularnewline\bottomrule
\end{tabular}}\end{center}
\end{table*}

\subsection{Academic Simulation-Based Analysis\label{subsec:Simulation-Analysis}}

This subsection presents a simple but thorough academic use case that
considers all introduced viewpoint constraints to perform an exhaustive
evaluation of the presented formulations. As stated in the introduction,
with this present scenario we aim to provide a first draft of a much-needed
benchmark for other researchers that can be used as basis for further
development, reproducibility and comparison. The surface models, the
resulting manifolds of the computed $\ACC$s, and the frustum spaces,
can be found attached in the additional material of our publication.

\subsubsection{Use-Case Description}

The exemplary case regards an $\ARVS$ with two sensors and an object
of interest containing three different features with different sizes
and geometries. Table \ref{tab:VC-Verification} gives a detailed
overview of the considered constraints. Besides the imaging parameters
of both range sensors, all other parameters can be assumed to be fictitious
though realistic. The kinematic and imaging models corresponds to
the real $\ARVS$ depicted in Figure \ref{fig:AiBox}.

\begin{figure*}[tbh]
\centering{}\includegraphics{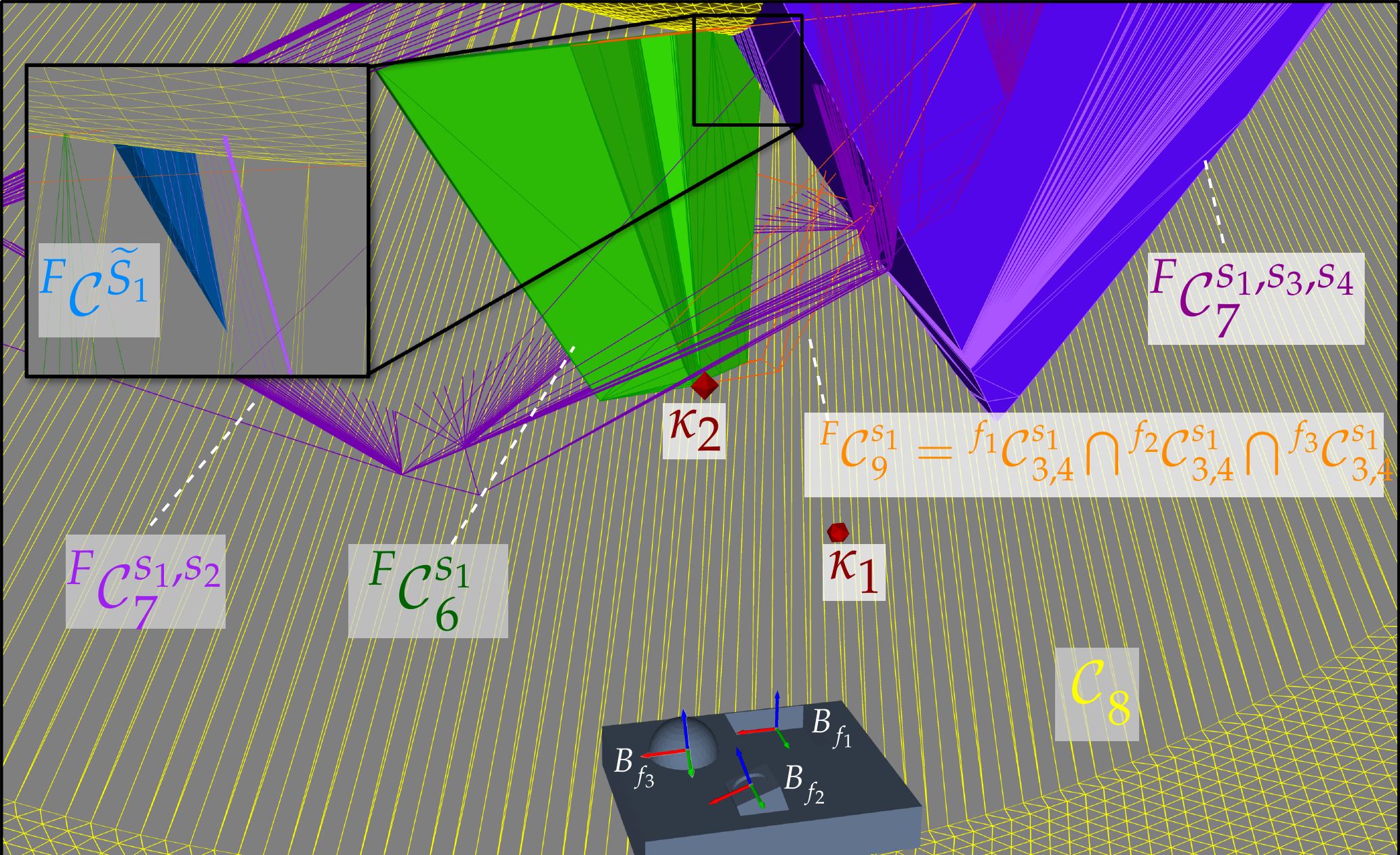}\caption{Characterization of the $\protect\ACC$ spanned by a set of viewpoint
constraints (see Table \ref{tab:VC-Verification}) for a multisensor
scenario to capture a set of features $F$. \label{fig:Eval-Sim-Characterization}}
\end{figure*}
\begin{figure*}[t]
\centering{}\includegraphics{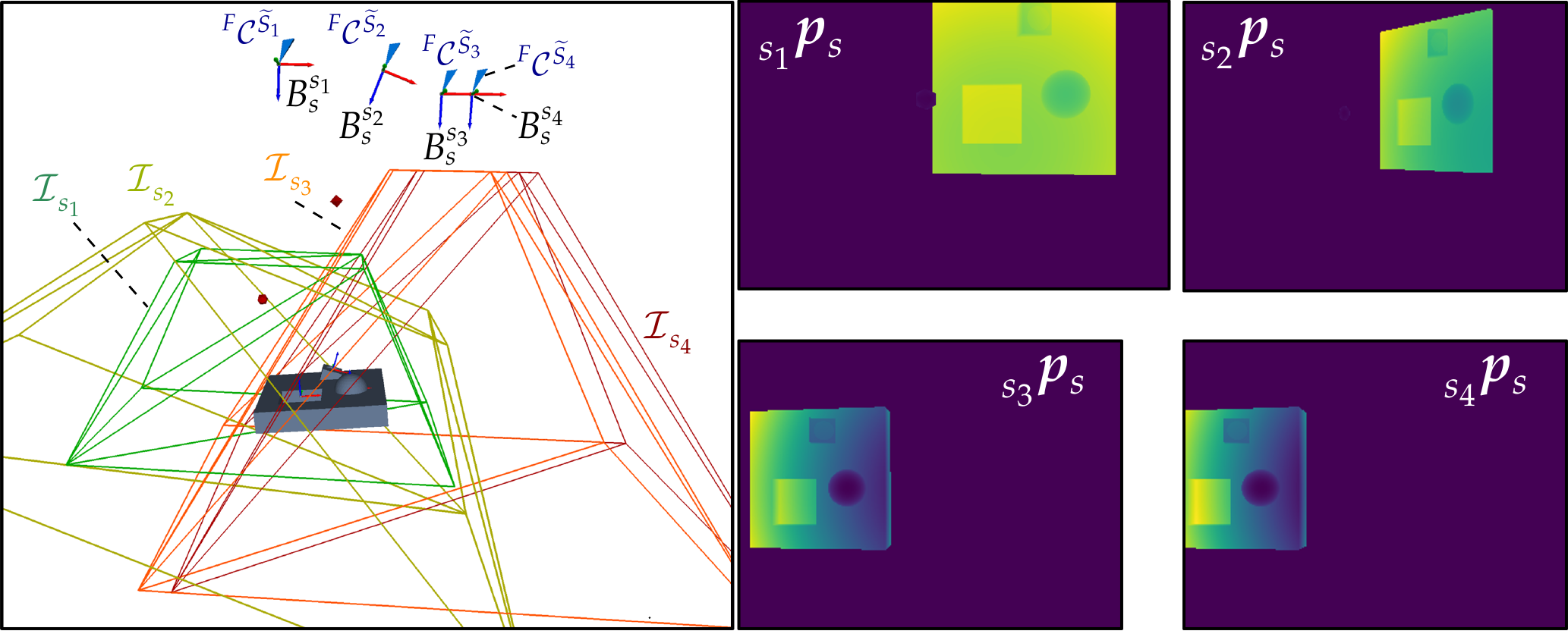}\caption{Left: Verification scene visualizing the frames and $\protect\ACM$s
of all imaging devices at the extreme sensor pose $\protect\VECJ ps{}{}{s_{1}}\in\protect\CC{}{\widetilde{S}_{1}}F$
($\protect\VECJ ps{}{}{s_{2}}\in\protect\CC{}{\widetilde{S}_{2}}F$,
$\protect\VECJ ps{}{}{s_{3}}\in\protect\CC{}{\widetilde{S}_{3}}F$,
and $\protect\VECJ ps{}{}{s_{4}}\in\protect\CC{}{\widetilde{S}_{4}}F$)
that fulfills all viewpoint constraints. Right: Depth images of all
imaging devices at the corresponding sensor pose. \label{fig:Eval-Sim-Rendered}}
\end{figure*}

\subsubsection{Results}

Following the strategy described in Algorithm \ref{alg:Viewpoint-Constraints-Integration},
the joint $\ACC$s of the four imaging devices, i.e., $\CC{}{\widetilde{S}_{1}}F$,
$\CC{}{\widetilde{S}_{2}}F$, $\CC{}{\widetilde{S}_{3}}F$, and $\CC{}{\widetilde{S}_{4}}{F,}$
were computed for acquiring all features considering all viewpoint
constraints from Table \ref{tab:VC-Verification}. Figure \ref{fig:Eval-Sim-Characterization}
shows the complexity of the described case comprising three features
and some of the resulting $\ACC$s. The blue manifold $\CC{}{\widetilde{S}_{1}}F$
represents the final constrained space of the first imaging device
$s_{1}$. It can be appreciated that $\CC{}{\widetilde{S}_{1}}F$
is characterized by the intersection of all other $\ACC$s. Moreover,
Fig. \ref{fig:Eval-Sim-Characterization} shows that the $\CC{}{\widetilde{S}_{1}}F$
manifold is mainly constrained by the $\ACC$ corresponding to the
second range sensor, i.e., $\CC 7{s_{1},s_{3},s_{4}}F$. 

To verify the validity of the computed $\ACC$s, the depth images
and point clouds for all imaging devices at eight extreme viewpoints
were rendered. Figure \ref{fig:Eval-Sim-Rendered} shows the corresponding
rendered scene and resulting depth images for each imaging device
at one extreme viewpoint $\VECJ ps{}{}{s_{1}}\in\CC{}{\widetilde{S}_{1}}F$.
The depth images demonstrate that all imaging device can successfully
acquire all features without occlusion simultaneously.

The total computation time for characterizing all $\ACC$s corresponded
to $t(\CC{}{\widetilde{S}_{1}}F,\CC{}{\widetilde{S}_{2}}F,\CC{}{\widetilde{S}_{3}}F,\CC{}{\widetilde{S}_{4}}F)\approx50\,s$.
However, this time comprises other computation steps (e.g., frames
transformation and inverse kinematic operations using ROS-Services),
which distort the effective computation time of the $\ACC$s. A proper
analysis of the computational efficiency of the whole strategy remains
to be further investigated. 

\subsubsection{Summary}

Despite the complexity of the use case, the framework (models, methods,
and integration strategy) presented within this paper demonstrated
its effectiveness in characterize a continuous topological space in
the special Euclidean, where all defined viewpoint constraints could
be fulfilled. The simulated depth-images and point clouds confirmed
that all selected viewpoints within the characterized $\ACC$ satisfied
all regarded constraints. Moreover, the proposed academic example
effectively outlines a simple but sufficient complex scenario to benchmark
our and future viewpoint planning strategies.

\subsection{Real Experimental Analysis \label{subsec:Experimental-Analysis}}

To assess the usability and validity of our framework within real
applications, the framework presented in this study was utilized to
generate automatically valid viewpoints for capturing different features
of a car door using real $\ARVS$, i.e., the \emph{AIBox }from \emph{ZEISS}.
The AIBox is an industrial measurement cell used to automate different
vision-based quality inspection tasks such as dimensional metrology
and digitization, among others.

\subsubsection{System Description}

\begin{figure}[t]
\includegraphics{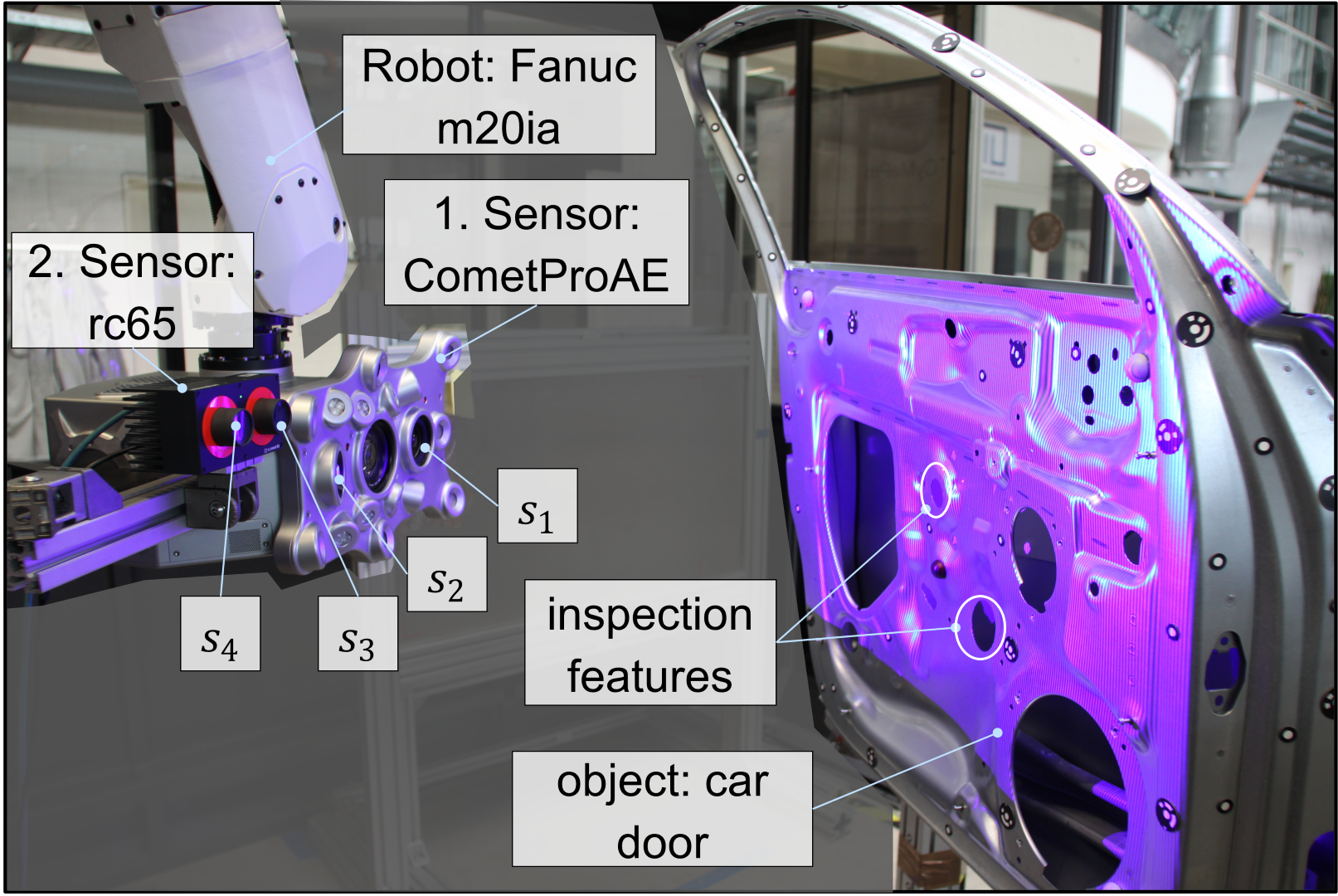}

\caption{Overview of the core components of the reconfigured inspection $\protect\ARVS$
AIBox.\label{fig:AiBox}}
\end{figure}

The AIBox is an integrated industrial $\ARVS$, equipped with a structured
light sensor (\emph{ZEISS COMET PRO AE}), a six-axis industrial robot
(\emph{Fanuc M-20ia}), and a rotary table for mounting an inspection
object. Moreover, to evaluate the use of our approach considering
a multisensor system, we additionally attached a stereo sensor (\emph{rc\_visard
65, Roboception}) to the structured light sensor. The imaging parameters
of both sensors are given in Table \ref{tab:Image Paremeters s1 and s2}.
Figure \ref{fig:AiBox} provides an overview of the reconfigured AIBox
with the stereo sensor. We assume that the inspection object is roughly
aligned, e.g., in \textcite{Magana.2020b} we presented a CNN fully
trained on synthetic data to automate this task using the $\ARVS$
sensor.

\subsubsection{Vision Task Definition}

The validation of our framework was performed on the basis of two
vision tasks considering diverse viewpoint constraints. For the first
task, we considered just the ZEISS sensor to acquire the features
$\{f_{1},f_{2}\}\in F_{1}$, which lie on the outside of the door
and can be potentially occluded by the door fixture. For the second
task, we considered both sensors and the acquisition of two features
$\{f_{3},f_{4},f_{5}\}\in F_{2}$ on the inside of the door. The incidence
angle for the first case corresponded to a sensor orientation of $\VECJ r{s_{2}}{}o{}(\alpha_{s}^{z}=\gamma_{s}^{x}=0{^\circ},\beta_{s}^{y}=-15{^\circ})$
and for the second of $\VECJ r{s_{1}}{}o{}(\alpha_{s}^{z}=\beta_{s}^{y}=\gamma_{s}^{x}=0{^\circ})$.
To compensate any kinematic modeling uncertainties, we consider an
overall kinematic error of $\epsilon_{s_{1}}^{x,y,z}=(70.0,70.0,50.0)\,\text{mm}$
for $s_{1}$ and of $\epsilon_{s_{3,4}}^{x,y,z}=(30.0,30.0,30.0)\,\text{mm}$
for $s_{3}$ and $s_{4}$. 

\subsubsection{Results}

\begin{figure}[tbh]
\centering{}\includegraphics{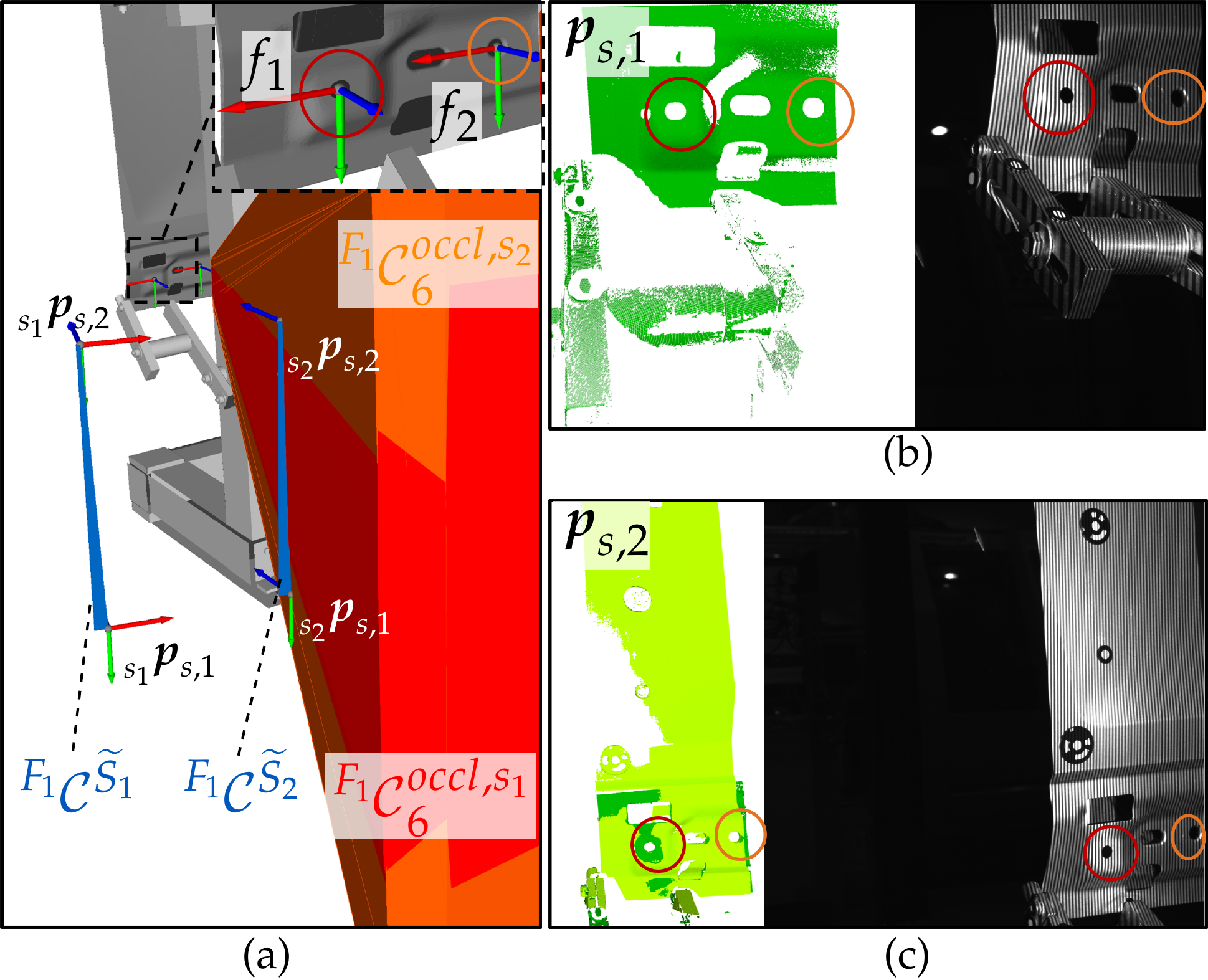}\caption{a) Visualization of the characterized $\protect\ACC$s($\protect\CC{}{\widetilde{S}_{1}}{F_{1}}$,$\protect\CC{}{\widetilde{S}_{2}}{F_{1}}$)
to capture features $f_{1}$ and $f_{2}$ by the \emph{COMET ProAE.}
Right figures: 2D images and corresponding point clouds at two extreme
viewpoints: b) for $\protect\VECJ p{s,1}{}{}{s_{1}}\in\protect\CC{}{\widetilde{S}_{1}}{F_{1}}$
and c) for $\protect\VECJ p{s,2}{}{}{s_{1}}\in\protect\CC{}{\widetilde{S}_{1}}{F_{1}}$.
\label{fig:Validation-CyMePro-Plan1}}
\end{figure}

For both vision tasks we computed the necessary $\ACC$s aligned to
the strategy presented by Algorithm \ref{alg:Viewpoint-Constraints-Integration}.
The $\ACC$s of the first inspection scenario for the camera $\CC{}{\widetilde{S}_{1}}{F_{1}}$
and projector $\CC{}{\widetilde{S}_{2}}{F_{1}}$ of the \emph{Comet
Pro AE }and its corresponding occlusion spaces are displayed on the
left image of Figure \ref{fig:Validation-CyMePro-Plan1}. To assess
the validity of the characterized $\ACC$s, we chose diverse extreme
viewpoints at the vertices of the $\CC{}{\widetilde{S}_{1}}{F_{1}}$
manifold and performed real measurements. On the right side of Figure
\ref{fig:Validation-CyMePro-Plan1} the real monochrome images of
the camera and the resulting point clouds at two validating viewpoints
are displayed. The 2D images and point clouds prove that both features
can be successfully acquired from these both viewpoints, which confirms
the free sight for the sensor and the illumination of both features
without shadows.

\begin{figure*}[tbh]
\centering{}\includegraphics{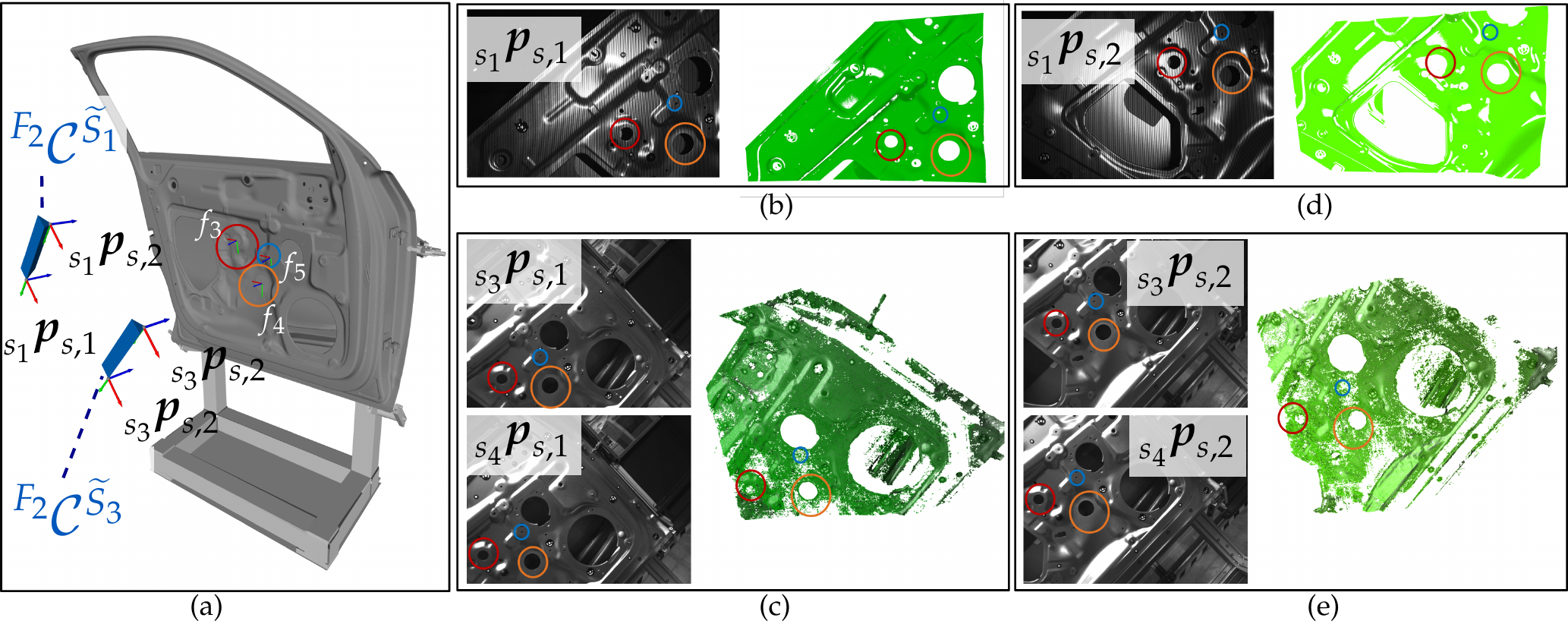}\caption{a): Visualization of the characterized $\protect\ACC$s ($\protect\CC{}{\widetilde{S}_{1}}{F_{2}}$,$\protect\CC{}{\widetilde{S}_{3}}{F_{2}}$)
to capture the feature set $\{f_{3},f_{4},f_{5}\}\in F_{2}$ by the
\emph{COMET ProAE} and \emph{rc\_visard 65}. Right figures: 2D images
and corresponding point clouds for both sensors at two extreme viewpoints
$\{\protect\VECJ p{s,1}{}{}{s_{1}},\protect\VECJ p{s,2}{}{}{s_{1}}\}\in\protect\CC{}{\widetilde{S}_{1}}{F_{2}}$
(upper figures b) and d): \emph{Comet Pro AE}, lower figures c) and
e): \emph{rc\_visard 65}). \label{fig:Validation-CyMePro-Plan2}}
\end{figure*}

Moreover, on the left of Figure \ref{fig:Validation-CyMePro-Plan2}
the constrained spaces of the first imaging device of each sensor,
i.e., $\CC{}{\widetilde{S}_{1}}{F_{2}}$ and $\CC{}{\widetilde{S}_{3}}{F_{2}}$,
are visualized for the second inspection scenario. Analogously to
the first scenario, two extreme viewpoints at the vertices of the
manifolds were selected to assess the validity of the computed $\ACC$s.
As expected, the real 2D images of all imaging devices and the resulting
point clouds of both sensors at two exemplary extreme viewpoints (shown
on the right of Figure \ref{fig:Validation-CyMePro-Plan1}) demonstrate
that all features can be successfully acquired by the four imaging
devices of both sensors. 

\subsubsection{Summary}

Using an industrial RVS, and regarding real viewpoint constraints,
we were able to validate the formulations, characterization, and application
of $\ACC$s for inspection tasks in an industrial context. These experiments
show the suitability of our framework for an industrial application
on a real $\ARVS$ with multiple range sensors. 

Furthermore, our strategy for merging individual $\ACC$s to capture
more than one feature proved to be effective for the regarded vision
tasks. However, a more complex task such as the inspection of all
door features requires a more complex strategy, which considers the
search of features that can be acquired together. This question recalls
the overall $\AVPP$, which falls outside the scope of this publication
and we intend to address in our future work.


\section{Summary and Outlook}

\subsection{Summary }

The computation of valid viewpoints considering different system constraints,
named $\AVGP$ in this publication, is considered a complex and unsolved
challenge that lacks a generic and holistic framework for its proper
formulation and resolution. In this paper, we outline the $\AVGP$
as a geometric problem that can be solved explicitly in the special
Euclidean $SE(3)$ using suitable and explicit models of all related
domains of an $\ARVS$ and viewpoint constraints. Within this context,
much of our effort was devoted to the comprehensive and systematic
formulation of the $\AVGP$ and the exhaustive characterization of
domains and viewpoint constraints aligned to the formulation of geometric
problems. 

The core result of this study is the characterization of $\ACC$s,
which can be understood as topological manifolds that span a space
with infinite viewpoint solutions to acquire one feature or a group
of features considering various viewpoint constraints and modeling
uncertainties. Our approach focuses on providing rather infinite valid
solutions instead of optimal ones. If the entire a priori knowledge
of the RVS can be formalized and integrated into the $\ACC$, then
it we can assume that any viewpoint within it is a local optimum.
Our work shows that a handful of viewpoint constraints can be efficiently
and simply modeled geometrically and integrated in a common framework
to span such constrained spaces. Finally, based on a comprehensive
academic example and a real application, we demonstrate the usability
of such a framework. 

\subsection{Limitations and Chances}

We are aware that the framework proposed in the present study may
have some limitations that may prevent its straightforward application
for other $\ARVS$ or use-cases. First, it must be regarded that our
framework can be classified under the category of model-based approaches.
Therefore, in a first step a priori information to model the components
of the considered $\ARVS$ must be regarded. We consider that an exhaustive
and explicit modeling of the necessary domains is necessary for delivering
solutions that offer a higher generalization for other applications
and systems. For the benefit of generalization, complexity reduction,
and computational efficiency we regarded various simplifications,
which could affect the accuracy of some models and might yield more
conservative, however, more robust solutions. 

We firmly believe that the $\AVGP$ can be efficiently solved geometrically.
We demonstrated that many constraints can be explicitly and efficiently
characterized by combining several techniques, including linear algebra,
trigonometry, and geometrical analysis. In the scope of our experiments,
we confirmed that the computation of the $\ACC$s manifolds based
on these approaches ran efficiently in linear times. However, we also
noted that algorithms comprising $\ACSG$ Boolean operations are more
computationally expensive, especially on calculations considering
multiple Boolean operations on the same manifold. Although this limitation
can be minimized by filtering and smoothing algorithms for decimating
manifolds, this characteristic could still be considered insufficient
for some users and applications. Although the shortcomings of $\ACSG$
Boolean techniques regarding their computational efficiency have been
mentioned in some prior works, we also believe that the present available
computational performance and paralleling capabilities of CPUs and
GPUs require a new reevaluation of their overall performance. Additionally,
our work also suggests that combined with efficient imaging processing
libraries, approaches requiring heavy use of $\ACSG$ operations can
be efficiently used within many applications. Nevertheless, a comprehensive
computational efficiency analysis to find a break-even point between
our approach and others remains to be further investigated.

Moreover, we also see room for improvement to increase the efficiency
of some of the algorithms presented. For instance, the computation
of the occlusion space and integration of constraints could also be
improved using an alternative strategy and more efficient algorithms
implemented in low-level programming languages. Additionally, we also
see potential for improving the efficiency of some algorithms. For
instance, the performance of many algorithms could enormously benefit
of computational optimization techniques such as parallelization and
GPU computation. For replication purposes of our work, we encourage
the reader to make a thorough evaluation of the performance of the
state-of-the-art libraries available at the present time, according
to their application needs and system requirements.

\subsection{Outlook}

We consider the use of $\ACC$s appropriate, but not limited to vision
tasks that rely on features. For example, we showed how our concept
could be extended to applications that generally would not consider
features and demonstrate its application for an object detection problem
with a certain level of spatial uncertainty. Our ongoing work concentrates
on assessing further applications or systems that may benefit from
our approach, e.g., feature-based robot calibration or adaption to
laser sensors. Further studies should still be undertaken in this
direction to verify the usability and explore the limitations of our
framework within other applications and $\ARVS$s.

Recalling that we neglected any sensor parameters that may directly
constrain the $\ACC$, e.g., exposure time, gain, and others, we consider
some other lines of research that integrate such a parameter space
in the $\ACV$. For instance, our ongoing study investigates the combination
of a data-based approach to optimize the exposure times and the use
of $\ACC$s for finding optimized sensor poses. 

The most promising future research should be devoted to the overall
problem of the $\AVPP$, which could be reformulated based on the
present study and its findings. Further research that will exploit
this and comprises a holistic strategy for its resolution is already
in progress.

We believe that our work will serve as a solid base and guideline
for further studies to adapt and extend our framework according to
the individual requirements of their concrete applications and $\ARVS$.

\section*{Acknowledgments}

The framework presented in this paper was developed and thoroughly
evaluated within the scope of the ``CyMePro'' (Cyber-physical measurement
technology for 3D digitization in the networked production) project,
which was funded by the Bavarian Ministry of Economic Affairs, Regional
Development and Energy (funding code ESB036 / 001). 

We thank our research partners AUDI AG and Carl Zeiss Optotechnik
GmbH for the fruitful discussions and their cooperation. Moreover,
we are deeply grateful to our colleagues Daria Leiber for the fruitful
discussions regarding the initial drafting and Benedikt Schmucker
for the critical revision of the present publication.

\printbibliography[heading=bibintoc]

\appendix
\begin{table}[tbh]
\caption{Description of General Requirements\label{tab:General-Requirements}}
\begin{center}
\resizebox{0.50\textwidth}{!}{%

\begin{tabular}{>{\raggedright}p{0.15\textwidth}>{\raggedright}p{0.3\textwidth}}
\toprule General Requirement  & Description\tabularnewline
\midrule 1. Generalization \label{GR:GR1}  & The models and approaches used should be abstracted and generalized
in the best possible level so that they can be used for different
components of an RVS and can be applied to solve a broad range of
vision tasks.\tabularnewline
\midrule 2. Computational Efficiency\label{GR:GR2} & The methods and techniques used should strive a low level of computational
complexity. Whenever possible, analytical and linear models should
be preferred over complex techniques, such as stochastic and heuristic
algorithms. Nevertheless, considering offline scenarios the trade-off
between computing a good enough solution within an acceptable amount
of time should be individually assessed.\tabularnewline
\midrule 3. Determinism \label{GR:GR3} & Due to traceability and safety issues within industrial applications
deterministic approaches should be prioritized. \tabularnewline
\midrule 4. Modularity and Scalability \label{GR:GR4} & The approaches and models should consider in general a modular structure
and promote their scalability. \tabularnewline
\midrule 5. Limited A-priori Knowledge \label{GR:GR5} & The parameters required to implement the models and approaches should
be easily accessible for the end-users. Neither in-depth optics nor
robotic knowledge should be required. \tabularnewline\bottomrule
\end{tabular}}
\end{center}
\end{table}

\begin{table}[tbh]
\caption{Overview of index notations for variables \label{tab:Notations}}

\noindent \begin{center}
\resizebox{0.48\textwidth}{!}{
\begin{tabular}[t]{>{\raggedright}p{0.08\textwidth}>{\raggedright}p{0.36\textwidth}}
\toprule Notation  & \multicolumn{1}{>{\centering}p{0.36\textwidth}}{Index Description}\tabularnewline
\midrule & $x\coloneqq$variable, parameter, vector, frame or transformation\tabularnewline
\noalign{\vskip\doublerulesep}
 & $d\coloneqq$$\ARVS$ domain, i.e., \uline{s}ensor, \uline{r}obot,
\uline{f}eature, \uline{o}bject, \uline{e}nvironment or \emph{d}
element of a list or set\tabularnewline
\noalign{\vskip\doublerulesep}
$\fr rbxdn$ & $n\coloneqq$related domain, additional notation or depending variable\tabularnewline
\noalign{\vskip\doublerulesep}
 & $r\coloneqq$base frame of the coordinate system $B_{r}$ or space
of feature $f$\tabularnewline
\noalign{\vskip\doublerulesep}
 & $b\coloneqq$origin frame of the coordinate system $B_{b}$\tabularnewline
\noalign{\vskip\doublerulesep}
\midrule Notes & The indexes $r$ and $b$ just apply for pose vectors, frames and
transformations.\tabularnewline
\midrule Example & The index notation can be better understood consider following examples:
\begin{itemize}
\item $d$: Let the geometry of a feature be described by a surface point
\textbf{$\VECJ gf{}{}{}\in\mathbb{R}^{3}$}
\item $d$: If the feature comprises more surface points, then let the point
with the index $2$ be denoted by $\VECJ g{f,2}{}{}{}\in\mathbb{R}^{3}$.
\item $r$: Assuming that the position of a surface point \textbf{$\VECJ gf{}{}{}$}
is described in the coordinate system of the feature, $B_{f}$, then
it follows: \textbf{$\VECJ gf{}f{}$}. In case that the base coordinate
frame has the same notation as the domain itself, i.e. $r=d$, then
just the index for the domain is given: $\VECJ gf{}{}{}=\VECJ gf{}f{}$.
\item $b$: In case that the frame of the surface point is given in the
coordinate reference system of the object $B_{o}$, then following
notation applies: $\VECJ gf{}o{}=\VECJ gf{}of$.
\end{itemize}
\tabularnewline
 & \tabularnewline\bottomrule
\end{tabular}}
\end{center}
\end{table}

\begin{table}[tbh]
\centering{}\caption{List of most common symbols \label{tab:List-Symbols}}
\begin{center}
\resizebox{0.48\textwidth}{!}{
\begin{tabular}{>{\raggedright}m{0.15\textwidth}>{\raggedright}m{0.3\textwidth}}
\toprule\textbf{ Symbol} & \multicolumn{1}{>{\raggedright}p{0.3\textwidth}}{\textbf{Description}}\tabularnewline
\midrule  & General\tabularnewline
\midrule $c$ & Viewpoint constraint\tabularnewline
$\widetilde{C}$  & Set of viewpoint constraints\tabularnewline
$f$ & Feature\tabularnewline
$F$ & Set of features\tabularnewline
$s_{t}$ & $t$ imaging device of sensor $s$\tabularnewline
\textbf{$v$} & Viewpoint\tabularnewline
$\SP{}$ & Sensor pose in $SE(3)$\tabularnewline
\midrule  & Spatial Dimensions\tabularnewline
\midrule $B$ & Frame\tabularnewline
$\VECJ t{}{}{}{}$ & Translation vector in $\mathbb{R}^{3}$\tabularnewline
$\VECJ r{}{}{}{}$ & Orientation matrix in $\mathbb{R}^{3x3}$\tabularnewline
$\boldsymbol{V}$ & Manifold vertex in $\mathbb{R}^{3}$\tabularnewline
\midrule  & Topological Spaces\tabularnewline
\midrule $\CC{}{}{}$ & $\ACC$ for a set of viewpoint constraints $\widetilde{C}$ \tabularnewline
$\CC i{}{}$ & i $\ACC$ of the viewpoint constraint $c_{i}$\tabularnewline
$\CF$ & Frustum space\tabularnewline\bottomrule
\end{tabular}}
\end{center}
\end{table}

\begin{table*}[tbh]
\caption{Overview of features and occlusion objects used for verification steps
and simulation-based analysis \label{tab:Verification-Features}}

\centering{}\begin{center}
\resizebox{1.0\textwidth}{!}{
\begin{tabular}{>{\raggedright}m{0.12\textwidth}|>{\raggedright}m{0.1\textwidth}>{\raggedright}m{0.1\textwidth}>{\raggedright}m{0.12\textwidth}>{\raggedright}m{0.12\textwidth}>{\raggedright}m{0.12\textwidth}>{\raggedright}m{0.12\textwidth}}
\toprule Feature & \multicolumn{1}{>{\raggedright}p{0.1\textwidth}}{$f_{0}$} & \multicolumn{1}{>{\raggedright}p{0.1\textwidth}}{$f_{1}$,$f_{1}^{*}$} & \multicolumn{1}{>{\raggedright}p{0.12\textwidth}}{$f_{2}$} & \multicolumn{1}{l}{$f_{3}$} & \multicolumn{1}{l}{$\kappa_{1}$} & \multicolumn{1}{l}{$\kappa_{2}$}\tabularnewline
\midrule Topology & Point & Slot & Circle & Half-Sphere & Icosahedron & Octaeder\tabularnewline
\midrule Generalized Topology & - & Square & Square & Cube & - & -\tabularnewline
\midrule Dimensions in $\text{mm}$ & $l_{f_{o}}=0$ & $l_{f_{1}}=50$, $h_{f_{1}^{*}}=30$ & $l_{f_{2}}=20$ & ${l_{f_{2}}=40}$, $h_{f_{2}}=40$ & edge length: $\approx14.0$ & edge length: $\approx20.0$\tabularnewline
\midrule Translation vector in object's frame $\fr{}{}to{}=(x_{o},y_{o},z_{o})^{T}$
in $\text{mm}$ & $\left(\begin{array}{c}
0.0\\
0.0\\
0.0
\end{array}\right)$ & $\left(\begin{array}{c}
0.0\\
0.0\\
0.0
\end{array}\right)$ & $\left(\begin{array}{c}
75.0\\
150.0\\
20.0
\end{array}\right)$ & $\left(\begin{array}{c}
120.0\\
30.0\\
0.0
\end{array}\right)$ & $\left(\begin{array}{c}
-67.5\\
0.0\\
240.0
\end{array}\right)$ & $\left(\begin{array}{c}
117.5\\
100.0\\
445.0
\end{array}\right)$\tabularnewline
\midrule Rotation in Euler Angles in object's frame $\fr{}{}ro{}(\gamma_{s}^{x},\beta_{s}^{y},\alpha_{s}^{z})$
in ${^\circ}$ & $(0,0,0)$ & $(0,0,0)$ & $(0,20,0)$ & $(0,0,0)$ & $(0,0,0)$ & $(0,0,0)$\tabularnewline\bottomrule
\end{tabular}}
\end{center}
\end{table*}

\begin{figure}[tbh]
\begin{centering}
\includegraphics{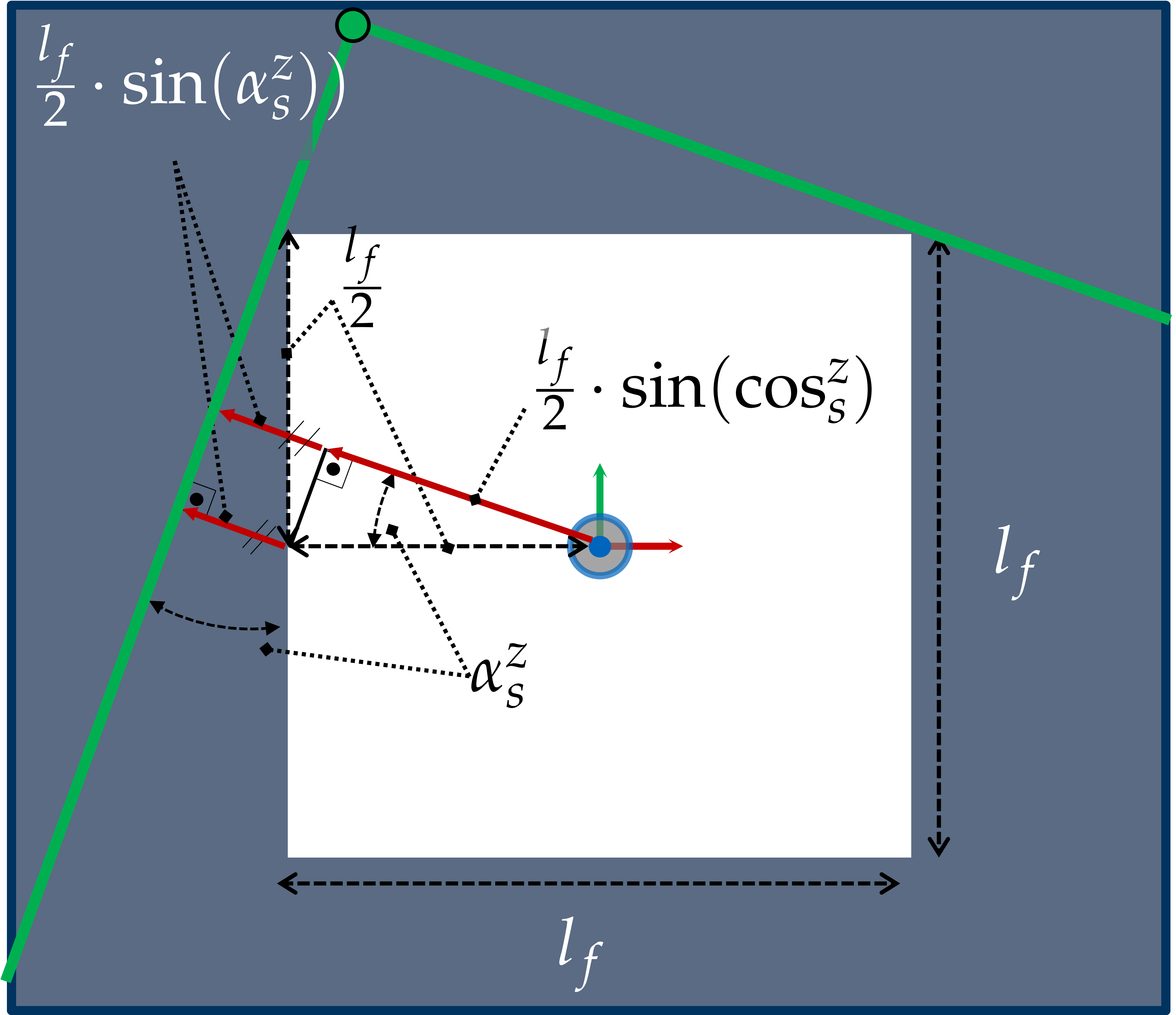}
\par\end{centering}
\caption{Geometrical analysis for rotation around z axis.\label{fig:CV3-SE1-zRot}}
\end{figure}

\begin{table*}[tbh]
\caption{Scaling factors for the vertices of the constrained space $\protect\VECJ Vk{\protect\CC 3{}{}}{}{}$
considering a sensor rotation around the $x$-axis or $y$-axis relative
to the feature frame \label{tab:Vertices-Deltas-xyRot}}

\begin{center}
\resizebox{0.95\textwidth}{!}{%

\setlength{\extrarowheight}{5pt}%
\begin{tabular}{>{\centering}p{0.1\textwidth}|>{\centering}p{0.08\textwidth}|>{\centering}p{0.08\textwidth}|>{\centering}p{0.1\textwidth}|>{\centering}p{0.08\textwidth}|>{\centering}p{0.1\textwidth}|>{\centering}p{0.08\textwidth}|>{\centering}p{0.08\textwidth}|>{\centering}p{0.08\textwidth}}
 & \multicolumn{4}{>{\centering}p{0.35\textwidth}|}{\textbf{Rotation around y-Axis}} & \multicolumn{4}{>{\centering}p{0.35\textwidth}}{\textbf{Rotation around x-Axis}}\tabularnewline
\multirow{3}{0.1\textwidth}{$k$ Vertex of $\VECJ Vk{\CC 3{}{}}{}{}$} & \multicolumn{4}{c|}{$\VECJ rs{}f{}(\alpha_{s}^{z}=\gamma_{s}^{x}=0,\beta_{s}^{y}\neq0)$} & \multicolumn{4}{c}{$\VECJ rs{}f{}(\alpha_{s}^{z}=\beta_{s}^{y}=0,\gamma_{s}^{x}\neq0)$}\tabularnewline
\cline{2-9} \cline{3-9} \cline{4-9} \cline{5-9} \cline{6-9} \cline{7-9} \cline{8-9} \cline{9-9} 
 & \multicolumn{2}{c|}{$\Delta_{k}^{x}$} & $\Delta_{k}^{y}$ & $\Delta_{k}^{z}$ & $\Delta_{k}^{x}$ & \multicolumn{2}{c|}{$\Delta_{k}^{y}$} & $\Delta_{k}^{z}$\tabularnewline
 & \centering{}$\beta_{s}^{y}<0$ & \centering{}$\beta_{s}^{y}>0$ &  &  &  & \centering{}$\gamma_{s}^{x}<0$ & \centering{}$\gamma_{s}^{x}>0$ & \tabularnewline
\hline 
\centering{}1 & \centering{}$\lambda^{x}$ & \centering{}$\rho^{x}$ & \centering{}$\frac{l_{f}}{2}$ & \centering{} & \centering{}$\frac{l_{f}}{2}$ & \centering{}$\lambda^{y}$ & \centering{}$\rho^{y}$ & \centering{}\tabularnewline
\centering{}2 & \centering{}$\rho^{x}$ & \centering{}$\lambda^{x}$ & \centering{}$\frac{l_{f}}{2}$ & \centering{} & \centering{}$\frac{l_{f}}{2}$ & \centering{}$\lambda^{y}$ & \centering{}$\rho^{y}$ & \centering{}\tabularnewline
\centering{}3 & \centering{}$\sigma^{x}$ & \centering{}$\rho^{x}$ & \centering{}$\frac{l_{f}}{2}+\varsigma^{x,y}$ & \centering{} & \centering{}$\frac{l_{f}}{2}+\varsigma^{y,x}$ & \centering{}$\rho^{y}$ & \centering{}$\sigma^{y}$ & \centering{}\tabularnewline
\centering{}4 & \centering{}$\rho^{x}$ & \centering{}$\sigma^{x}$ & \centering{}$\frac{l_{f}}{2}+\varsigma^{x,y}$ & \multirow{2}{0.08\textwidth}{\centering{}$\rho^{z,y}$} & \centering{}$\frac{l_{f}}{2}+\varsigma^{y,x}$ & \centering{}$\rho^{y}$ & \centering{}$\sigma^{y}$ & \multirow{2}{0.08\textwidth}{\centering{}$\rho^{z,x}$}\tabularnewline
\centering{}5 & \centering{}$\lambda^{x}$ & \centering{}$\rho^{x}$ & \centering{}$\frac{l_{f}}{2}$ &  & \centering{}$\frac{l_{f}}{2}$ & \centering{}$\rho^{y}$ & \centering{}$\lambda^{y}$ & \tabularnewline
\centering{}6 & \centering{}$\rho^{x}$ & \centering{}$\lambda^{x}$ & \centering{}$\frac{l_{f}}{2}$ & \centering{} & \centering{}$\frac{l_{f}}{2}$ & \centering{}$\rho^{y}$ & \centering{}$\lambda^{y}$ & \centering{}\tabularnewline
\centering{}7 & \centering{}$\sigma^{x}$ & \centering{}$\rho^{x}$ & \centering{}$\frac{l_{f}}{2}+\varsigma^{x,y}$ & \centering{} & \centering{}$\frac{l_{f}}{2}+\varsigma^{y,x}$ & \centering{}$\sigma^{y}$ & \centering{}$\rho^{y}$ & \centering{}\tabularnewline
\centering{}8 & \centering{}$\rho^{x}$ & \centering{}$\sigma^{x}$ & \centering{}$\frac{l_{f}}{2}+\varsigma^{x,y}$ & \centering{} & \centering{}$\frac{l_{f}}{2}+\varsigma^{y,x}$ & \centering{}$\sigma^{y}$ & \centering{}$\rho^{y}$ & \centering{}\tabularnewline
\end{tabular}}
\end{center}
\end{table*}

\begin{figure*}[t]
\centering{}%
\begin{minipage}[t]{0.48\textwidth}%
\begin{center}
\subfloat[1. Vertex of $\protect\VECJ V1{\protect\CC 3{}{}}{}{}$]{\centering{}\includegraphics{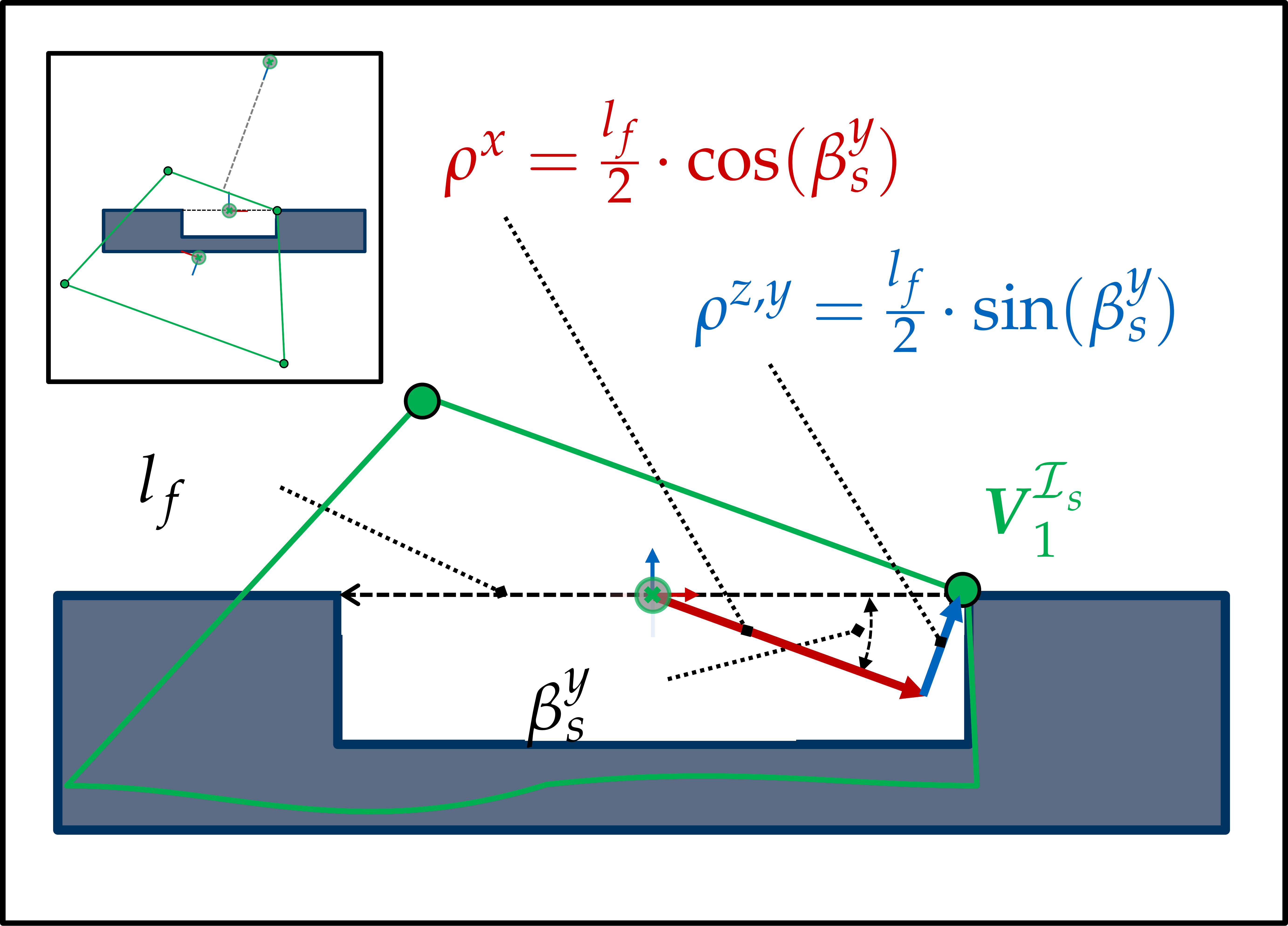}}
\par\end{center}%
\end{minipage}\hfill{}%
\begin{minipage}[t]{0.48\textwidth}%
\begin{center}
\subfloat[2. Vertex of $\protect\VECJ V2{\protect\CC 3{}{}}{}{}$]{\centering{}\includegraphics{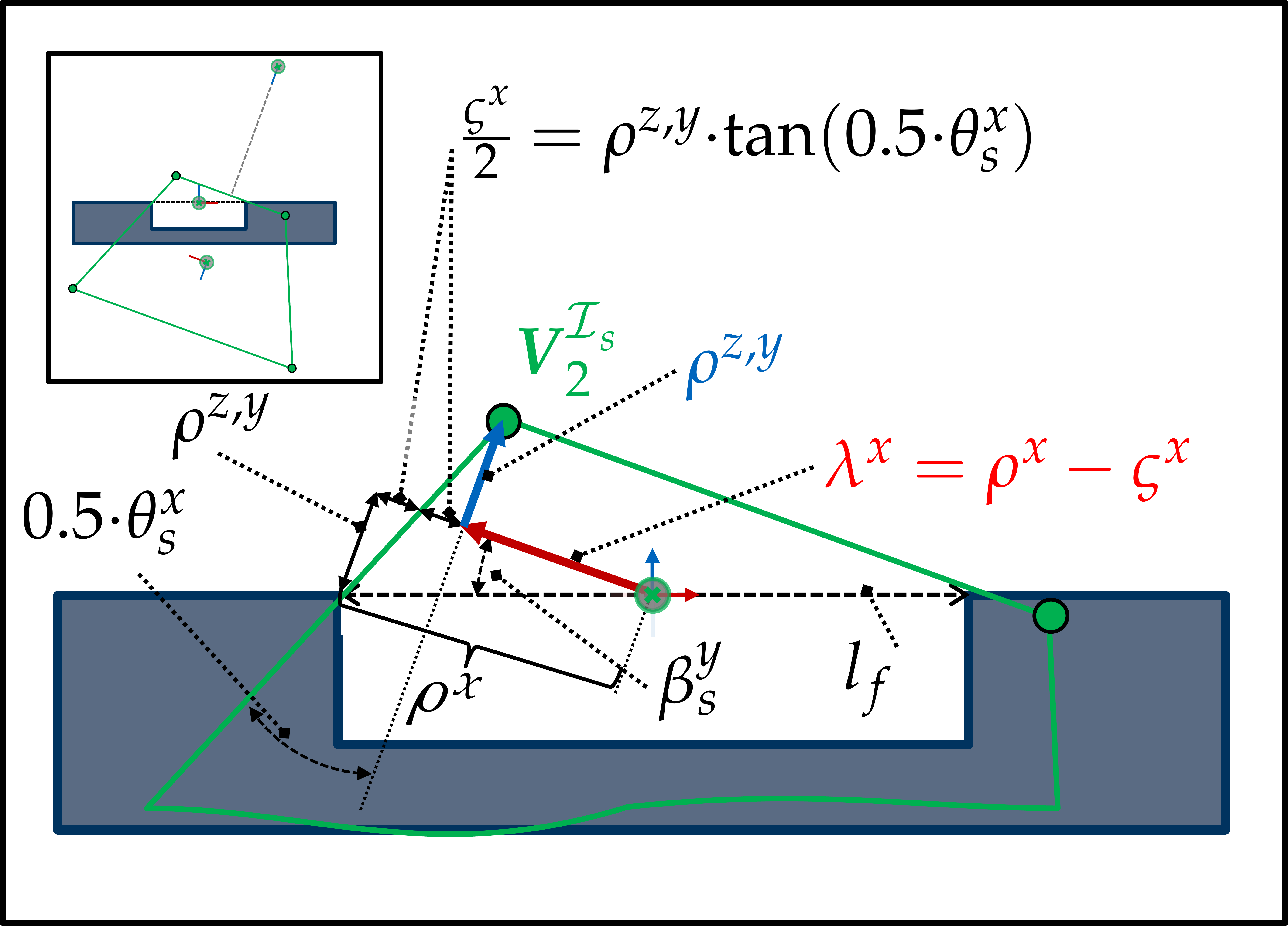}}
\par\end{center}%
\end{minipage}\hfill{}%
\begin{minipage}[t]{0.48\textwidth}%
\begin{center}
\subfloat[3. Vertex of $\protect\VECJ V3{\protect\CC 3{}{}}{}{}$]{\centering{}\includegraphics{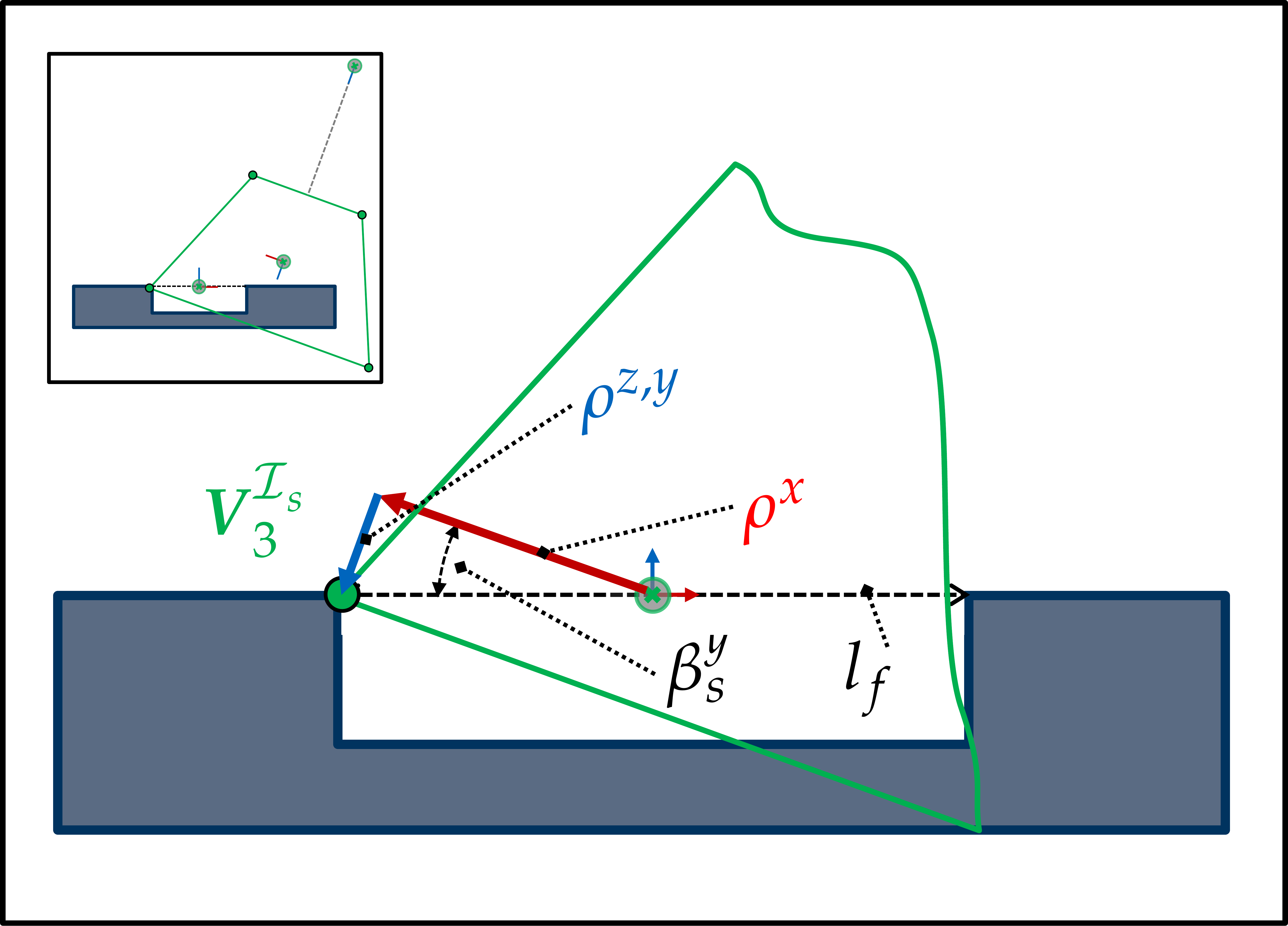}}
\par\end{center}%
\end{minipage}\hfill{}%
\begin{minipage}[t]{0.48\textwidth}%
\begin{center}
\subfloat[4. Vertex of $\protect\VECJ V4{\protect\CC 4{}{}}{}{}$]{\centering{}\includegraphics{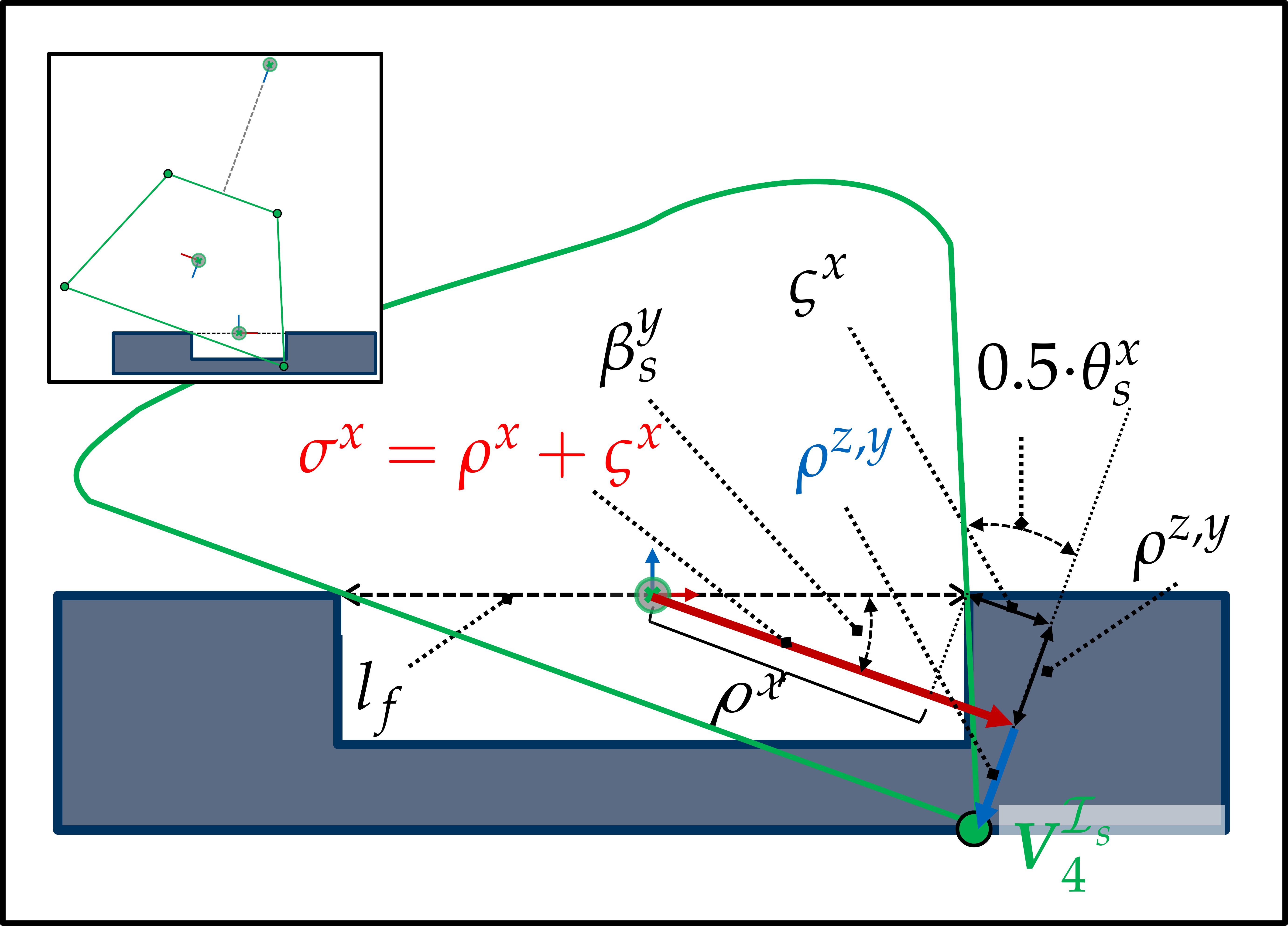}}
\par\end{center}%
\end{minipage}\caption{Derivation of the geometrical relationships for each vertex of the
$\protect\ACC$ $\protect\CC 3{}{}$ considering considering a sensor
rotation of $\protect\VECJ rs{}f{}(\beta_{s}^{y}>0,\alpha_{s}^{z}=\gamma_{s}^{x}=0)$
using the\emph{ Extreme Viewpoint Interpretation}. \label{fig:CV3-SE1-yRot} }
\end{figure*}

\begin{figure}[tbh]
\centering{}\includegraphics{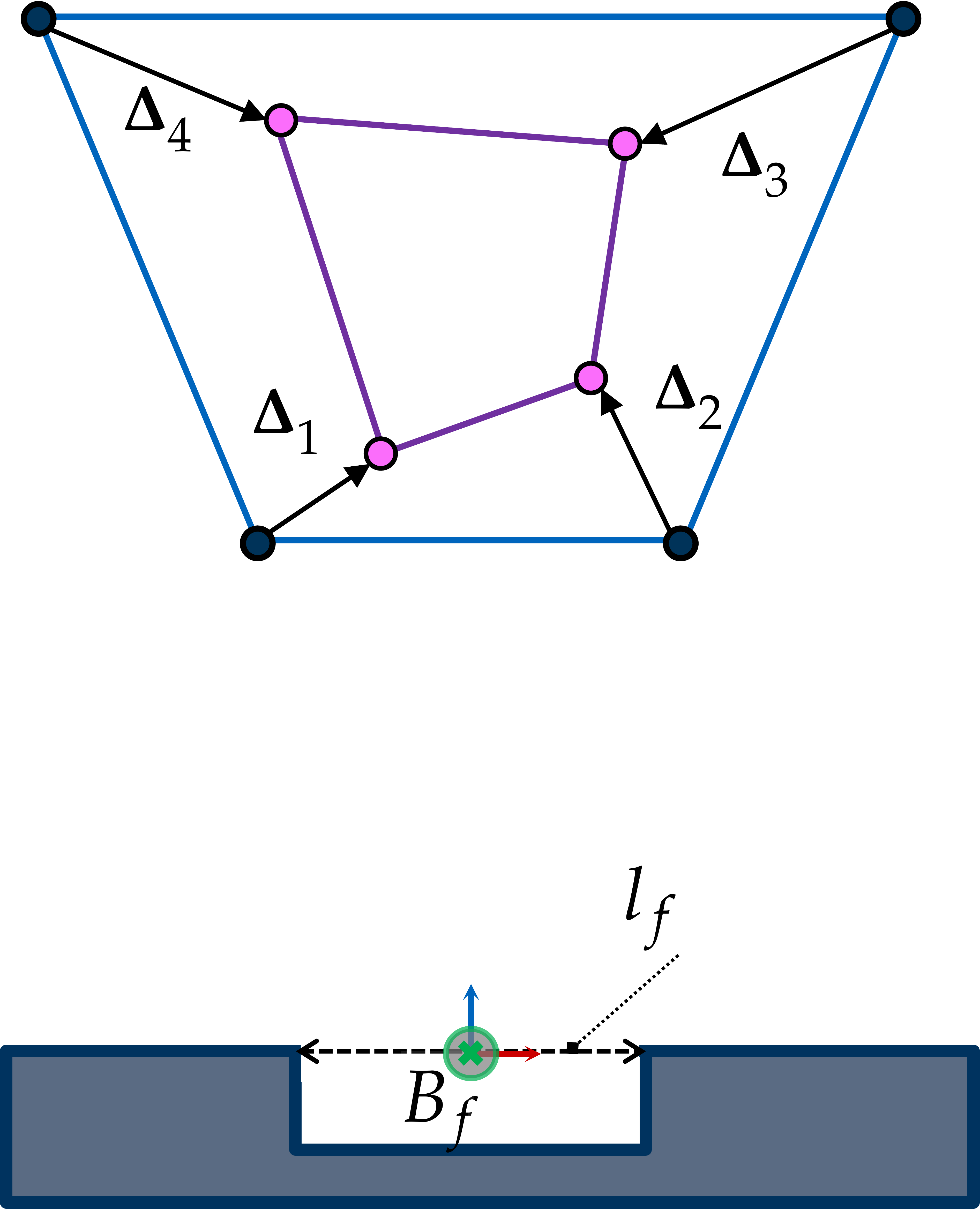}\caption{Exemplary flexible characterization of the viewpoint constraints (e.g.,
kinematic errors and sensor accuracy) using different scaling vectors
for each vertex.\label{fig: CV-Scaling}}
\end{figure}

\begin{algorithm}[tbh]
\caption{Computation of View Rays for Occlusion Space\label{subsec:CV6-View-Rays}}

\begin{enumerate}
\item Considering the simplification of the feature topology (cf. \ref{subsec:Simplification-Feature})
let a set of view rays denoted by $\VECJ{\tau}{f,c,l,m}{}{}{}\in\Sigma_{c},c=\{0,1,2,3,4\}$
be shot at each feature corner point $\VECJ g{f,c}{}{}{}\in\mathbb{R}^{3}$
with following direction vectors $\VECJ{\sigma}{c,l,m}{}{}{}(\sigma_{c,l}^{x},\sigma_{c,m}^{y})\in\mathbb{R}^{3}$:
\[
\VECJ{\tau}{f,c,m,n}{}{}{}=\VECJ g{f,c}{}{}{}+\VECJ{\sigma}{c,m,n}{}{}{}(\sigma_{m}^{x},\sigma_{n}^{y})
\]
\item The direction vectors are characterized by a grid of equidistant rays,
which can be expressed by means of the aperture angles $\sigma_{m}^{x}$
and $\sigma_{n}^{y}$:
\[
-\frac{\sigma_{max}^{x}}{2}\leq\sigma_{m}^{x}\leq\frac{\sigma_{max}^{x}}{2}\text{ and }-\frac{\sigma_{max}^{y}}{2}<\sigma_{n}^{y}<\frac{\sigma_{max}^{y}}{2}.
\]
The maximal aperture angles $\sigma_{max}^{x}$ and $\sigma_{max}^{y}$
can simply correspond to the FOV angles of the sensor. An efficient
alternative is to consider the aperture angles $\CC{}{}{}$, which
already comprises the FOV angles and other constraints. The total
number of rays depends on the chosen step size $d^{\sigma}$ for computing
the equidistant view rays:
\[
\begin{aligned}l\in[1,\dotsc,\frac{\sigma_{max}^{x}-\sigma_{min}^{x}+1}{d^{\sigma}}] & ,\\
m\in[1,\dotsc,\frac{\sigma_{max}^{y}-\sigma_{min}^{y}+1}{d^{\sigma}}] & .
\end{aligned}
\]
\end{enumerate}
\end{algorithm}

\begin{figure}[tbh]
\begin{centering}
\includegraphics{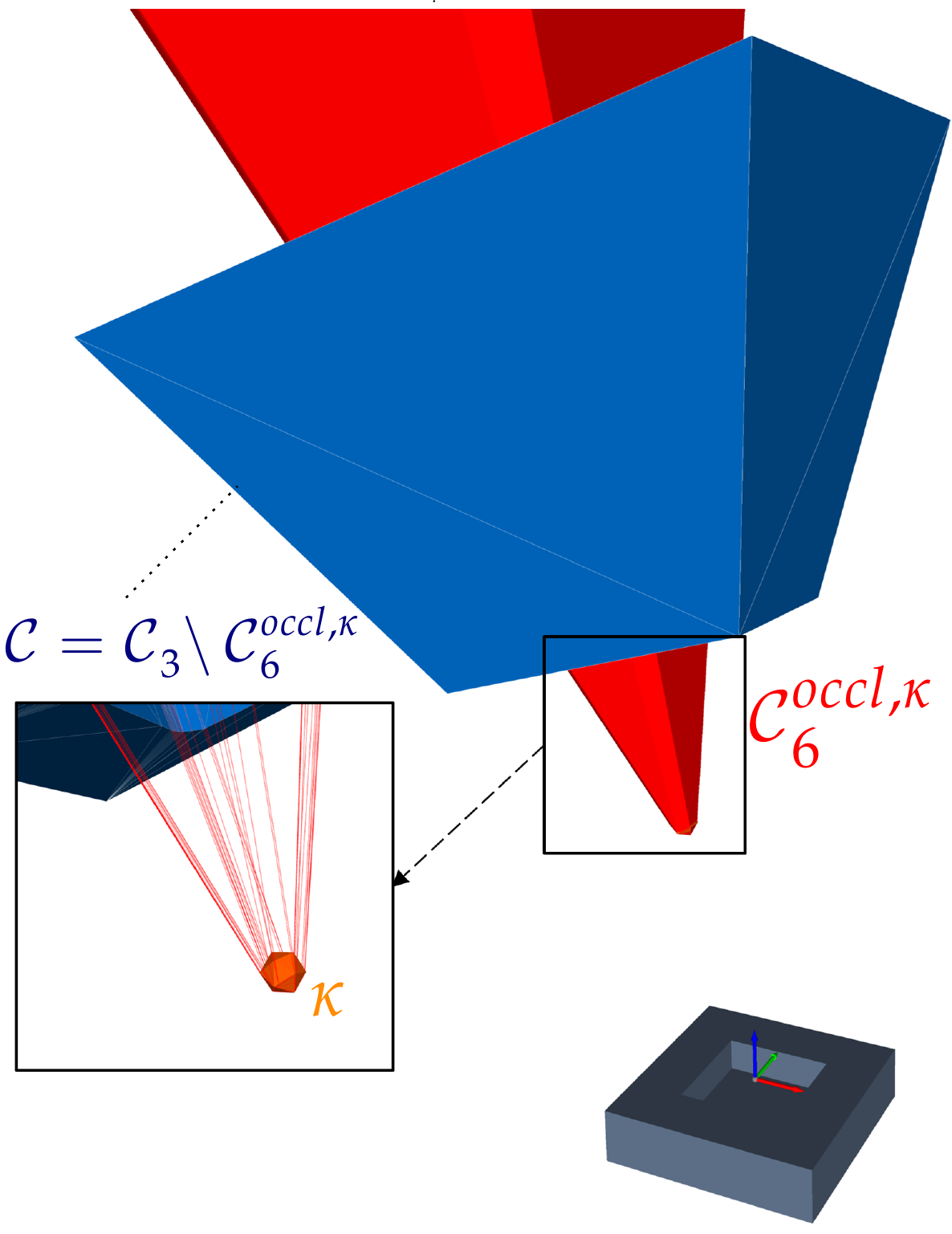}
\par\end{centering}
\caption{Occlusion $\protect\ACC$ in SE(3) (red manifold) and the occlusion
free $\protect\ACC$ (blue manifold) to acquire a square feature $f_{1}$
considering an occlusion body $\kappa$ (icosahedron in orange). \label{fig:CV6-Veri}}

\end{figure}

\begin{figure}[t]
\centering{}\includegraphics{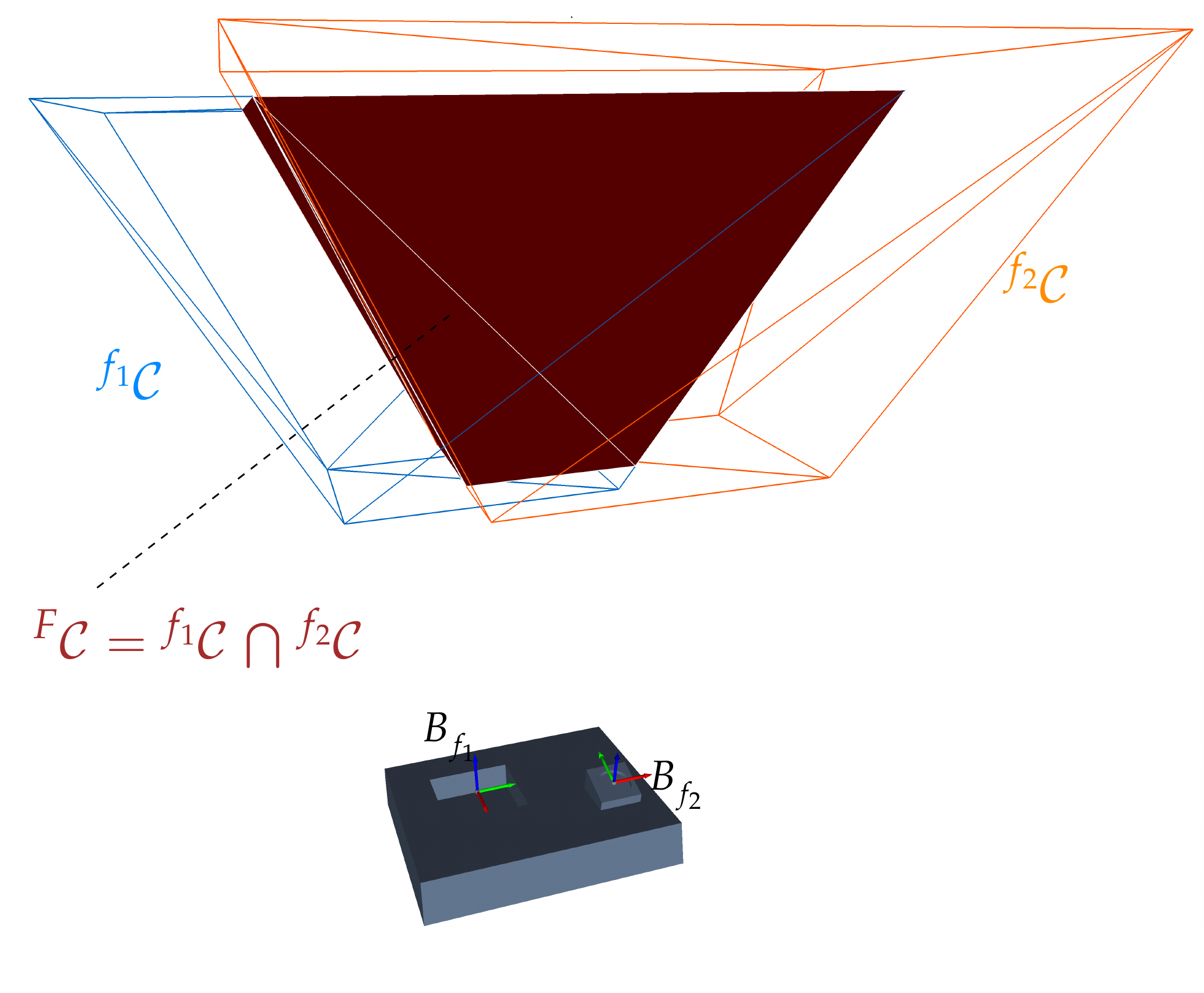}\caption{Characterization of the $\protect\ACC$, $\protect\CC{}{}F$, in $SE(3)$
to acquire a set of features $\{f_{1},f_{2,}\}\in F$ being characterized
by the intersection of the individual $\protect\ACC$s $\protect\CC{}{}{f_{1}}$
and $\protect\CC{}{}{f_{2}}$. \label{fig:CV9-Verification-Scene}}
\end{figure}

\end{document}